\documentclass[twoside,openright]{scrreprt}

\usepackage[msc]{tugrazthesis}

\usepackage{natbib}
\usepackage{hyperref}
\usepackage{url}
\usepackage{eso-pic}
\usepackage{graphicx}
\usepackage{booktabs}
\usepackage{multirow}
\usepackage{float}
\usepackage{makecell}

\usepackage{bbm}
\usepackage{geometry}
\usepackage{subcaption}
\usepackage{amsmath}
\usepackage{caption}
\usepackage{array}
\usepackage{float}
\usepackage{amssymb}
\usepackage{enumitem}
\begin{document}


\thesisauthor[Firstname Lastname]{Christina Thrainer, BSc}

\thesistitle[Short Thesis Title]{AI-Based Culvert-Sewer Inspection}

\thesisdate[ ]{October 2025}

\supervisortitle{\germanenglish{Betreuerin/Betreuer}{Supervisor}}

\supervisor{%
  Friedrich Fraundorfer, Univ.-Prof. Dipl.-Ing. Dr.techn. \\
  Institute of Visual Computing

}

\cosupervisortitle{\germanenglish{Co-Betreuerin/Betreuer}{Co-Supervisors}}

\cosupervisor{
  Christian Guetl, Assoc.Prof. Dipl.-Ing. Dr.techn. \\
  Institute of Human-Centred Computing\\
  \vspace{0.3cm}
  Mahdi Abdelguerfi, Dr. \\
  Md Meftahul Ferdaus, Dr.  \\
  Canizaro Livingston Gulf States Center for Environmental Informatics
}

\academicdegree{Diplom-Ingenieur/Diplom-Ingenieurin}

\curriculum{Computer Science}



\printthesistitle


\chapter*{Abstract}

Culverts and sewer pipes are critical components of drainage systems, and their failure can lead to serious risks to public safety and the environment. Regular inspections are necessary. However, manual inspections are time-consuming and prone to error. In this thesis, we explore methods to improve automated defect segmentation in culverts and sewer pipes. Collecting and annotating data in this field is cumbersome and requires domain knowledge. Having a large dataset for structural defect detection is therefore not feasible. Our proposed methods are tested under conditions with limited annotated data to demonstrate applicability to real-world scenarios.\\
\\
First, we evaluate preprocessing strategies, including traditional data augmentation and dynamic label injection. These techniques significantly improve segmentation performance, increasing both Intersection over Union (IoU) and F1 score. Second, we introduce FORTRESS, a novel architecture that combines depthwise separable convolutions, adaptive Kolmogorov–Arnold Networks (KAN), and multi-scale attention mechanisms. FORTRESS achieves state-of-the-art performance on the culvert sewer pipe defect dataset, while significantly reducing the number of trainable parameters, as well as its computational cost. It generates accurate segmentation masks and handles diverse defect appearances effectively, even without data augmentation. Finally, we investigate few-shot semantic segmentation and its applicability to defect detection. Few-shot learning aims to train models with only limited data available. By employing a bidirectional prototypical network with attention mechanisms, the model achieves richer feature representations and achieves satisfactory results across evaluation metrics.\\
\\
Overall, this thesis proposes three methods to significantly enhance defect segmentation and handle data scarcity. This can be addressed either by enhancing the training data or by adjusting a model's architecture. For the latter, we propose two complementary approaches. FORTRESS allows accurate real-time inference, while few-shot learning can quickly adapt to new classes and reduce problems caused by class imbalance.

\chapter*{Kurzfassung}

Durchlässe und Abwasserrohre sind wichtige Bestandteile des Abwassersystems. Deren Ausfall kann Risiken für die öffentliche Sicherheit und die Umwelt darstellen. Regelmäßige Inspektionen sind daher notwendig. Manuelle Inspektionen sind jedoch zeitaufwendig und fehleranfällig. In dieser Masterarbeit untersuchen wir verschiedene Methoden zur Verbesserung der automatisierten Defekterkennung in Durchlässen und Abwasserrohren. Die Datenerfassung und Annotation in diesem Bereich ist aufwendig und erfordert Fachwissen. Ein großer Datensatz ist daher nicht möglich. Unsere vorgestellten Methoden wurden mit limitierten Daten getestet, um ihre Anwendbarkeit in der realen Welt zu demonstrieren.\\
\\
Zunächst evaluieren wir verschiedene Ansätze zur Datenaufbereitung, einschließlich klassischer Datenerweiterungsstrategien und dynamischer Label-Injektion. Diese Techniken verbessern die Segmentierungsleistung deutlich und erhöhen sowohl die Intersection over Union (IoU, Schnittmenge über Vereinigung) als auch das F1-Maß. Darauf folgend stellen wir unsere neu entwickelte Architektur, FORTRESS, vor. Sie kombiniert kanalweise separierbare Convolution, adaptive Kolmogorov-Arnold-Netzwerke (KAN) und verschiedene Attention-Mechanismen und übertrifft bei unserem verwendeten Datensatz mit Defekten im Abwassersystem alle bisherigen Modelle und reduziert gleichzeitig die Anzahl der trainierbaren Parameter erheblich. Es erzeugt präzise Segmentierungsmasken und verarbeitet unterschiedliche Defektarten effektiv, selbst ohne Datenerweiterung. Zuletzt wird die Anwendbarkeit einer semantischen Segmentierung der Defekte im Few-Shot-Szenario untersucht. Few-Shot-Lernen umfasst das effiziente Training eines Modells bei begrenzter Datenverfügbarkeit. Durch den Einsatz eines bidirektionalen prototypischen Netzwerks mit verschiedenen Attention-Mechanismen werden reichhaltige Repräsentationen der Daten erzeugt und führt zu zufriedenstellenden Ergebnissen über alle Evaluierungsmetriken. \\
\\
Zusammenfassend stellt diese Arbeit drei Methoden vor, die die Defektsegmentierung deutlich verbessern und mit begrenzter Datenverfügbarkeit umgehen können. Dies kann entweder durch Erweiterung der Trainingsdaten oder durch Anpassung der Modellarchitektur erfolgen. Für Letzteres stellen wir zwei komplementäre Ansätze vor. FORTRESS ermöglicht eine präzise Echtzeit-Inferenz, während Few-Shot-Lernen schnell auf neue Klassen adaptieren kann und Probleme zusammenhängend mit einer ungleichen Anzahl von Bildern pro Klasse reduziert.

\cleardoublepage

\chapter*{\germanenglish{Danksagung}{Acknowledgements}}

First, I would like to sincerely thank my main supervisor, Professor Friedrich Fraundorfer, for his guidance and feedback throughout the writing process of this thesis. I would also like to thank Professor Christian Gütl for making my research stay in New Orleans possible and guiding me before, during and after my stay. Also, I really appreciate the financial support I received from the Austrian Marshall Plan Foundation. \\
\\
I am very grateful to the Canizaro Livingston Gulf States Center for Environmental Informatics (GulfSCEI) in New Orleans allowing me to stay with them for incredible five months. I would especially like to thank my supervisors there, Professor Mahdi Abdelguerfi and Professor Md Meftahul Ferdaus, for their guidance throughout my research and their never-ending patience. Their support made my time at GulfSCEI memorable.\\
\\
Finally, I want to thank my family. Their support and encouragement throughout my studies, during my stay abroad, and while writing this thesis helped me a lot.

\cleardoublepage

\tableofcontents

\listoffigures

\listoftables


% ...


\chapter{\germanenglish{Einleitung}{Introduction}}

Culverts and sewer pipes are essential components of our drainage systems. Their failure poses a significant threat to public safety and the environment. Regular inspections and improvements of flood control infrastructure are therefore crucial~\cite{kuchi21}. Incorporating autonomous defect detection systems instead of manual inspections offers substantial benefits and enhances the efficiency and accuracy of the inspection process. The United States has more than 100,000 miles of levees and associated culverts, and currently, all checks are performed manually~\cite{acsi2023}. Traditional approaches inspect pipes for holes, deformations and cracks using video inspection. The videos are manually reviewed, which is time-consuming and error-prone. By incorporating autonomous defect detection, a more reliable inspection compared to manual methods is possible. This will enable early detection of defects leading to better planning of repairs and an increased lifetime of pipes and culverts~\cite{alshawi24b}. \\
\\
Computer vision has gained great importance in the industry.  The combination of image information and deep learning methods achieves a better understanding of the image content, leading to a reliable decision-making~\cite{alshawi24b}. Semantic segmentation is a popular method in computer vision for identifying objects in images and separating them from their background. The segmentation process assigns a representative class label to each pixel in an input image based on certain characteristics such as texture and color intensities. All pixels belonging to the same class share certain features~\cite{bali15}. The introduction of fully convolutional networks (FCNs)~\cite{long15} was an important milestone for semantic segmentation approaches because it allowed a more generalized segmentation by assigning classes pixel by pixel and accepting input images of arbitrary size. U-Net~\cite{ronneberger2015u} and Feature Pyramid Network (FPN)~\cite{kim18} are both fully convolutional networks and popular deep learning models in the area of semantic segmentation. While U-Net utilizes an encoder-decoder architecture, FPN has a pyramid-like structure where it stores feature maps at different scales in a hierarchical pyramid. The recent introduction of Vision Transformers (ViT)~\cite{dosovitskiy21} opened up new possibilities. ViTs include self-attention mechanisms to gather global context and long-range relationships to focus on important features and neglect irrelevant ones. U-Net, FPN and ViT perform well on standard segmentation tasks, but poorly on infrastructure inspection tasks. Culverts and sewer pipes include significant challenges due to their variance in appearance. They differ a lot in shape, size, and visual characteristics due to different environmental conditions. They are affected by vegetation, debris, changing lighting conditions, and occlusions. Additionally, defects are often small and not easy to identify~\cite{alshawi24b}. This introduces a foreground-background class imbalance, where only a few pixels are labeled as the foreground compared to the background class~\cite{terven23}. Furthermore, the model must integrate multi-scale feature extraction to effectively detect small details while capturing the entire pipe structure. Encoder-decoder networks (EDN), such as U-Net, have problems with detecting features at different scales, including small defects and large pipe characteristics. Pyramid-like architectures, such as FPN, handle the multi-scale problem more effectively but are more easily affected by class imbalance than EDNs, often making biased predictions in favor of overrepresented classes. Vision Transformers are very powerful, but simultaneously computationally expensive and rely on extensive training data~\cite{alshawi24b}. Important factors for autonomous defect detection of culverts and pipes include multi-scale feature extraction, handling class imbalance, and computational efficiency. Recent advancements in semantic segmentation have shown promising results and demonstrated improvements in this area by incorporating advanced deep learning techniques~\cite{alshawi23a, alshawi23b, alshawi24b}. However, segmenting pipes and culverts remains challenging, and further enhancements are necessary. \\
\\
Previously, the lack of datasets on defects in culverts and sewer pipes made training segmentation models for defect detection very difficult. In 2024, Alshawi et al.~\cite{alshawi24b} introduced a dataset containing images of culverts and pipes. However, the dataset is not balanced. Some defects, such as joint misalignments and cracks, appear frequently in the dataset, while other deficiencies, like erosions and deformations, are underrepresented. As a result, models tend to predict the overrepresented defects more often and perform worse on the rare ones. This lowers both accuracy and generalization. Collecting more data is difficult due to the complex and time-consuming process of collecting and annotating data. To overcome this issue, researchers introduced not only advancements in the model's architecture but also new data preprocessing strategies. Data augmentation is a common approach that creates a more diverse dataset by reusing existing image–mask pairs and applying, for example, different geometric transformations or color adjustments. This helps the model learn different variations of defect classes and adapt to new perspectives and conditions~\cite{shorten19}.\\
\\
Few-shot learning is a machine learning approach that aims to train models effectively with only limited data. This is especially valuable in domains where collecting and annotating data is costly and time-consuming. Few-shot semantic segmentation is a specialization of this approach for segmentation tasks. Instead of requiring large datasets with thousands of annotated images, the model only needs a few annotated samples to generate accurate segmentation masks. Training is organized into episodes. In each episode, the model receives a support set containing images and their ground-truth masks, as well as a query set containing only images. The model extracts useful information from the support set and applies this knowledge to predict the segmentation mask of the query image. In typical settings, the model receives one, five, or ten images per class per episode. This episodic training allows the model to adapt quickly to new classes, and because each class has the same number of images, no class imbalance causes biased predictions~\cite{catalano23}. Several recent studies highlight the potential of few-shot segmentation for critical tasks, such as medical image analysis and structural defect detection. In these areas, rare but important classes often appear only a few times in the dataset. By using a few-shot approach, models achieve reasonable accuracy for segmenting these underrepresented classes~\cite{pachetti24}. However, few-shot segmentation is a relatively new research direction, and further investigations are needed.\\
\\
In this work, we propose a combination of advanced data preprocessing techniques and architectural innovations to improve performance in structural defect detection. The preprocessing pipeline includes both traditional data augmentation methods and dynamic label injection~\cite{caruso24}, a novel copy-and-paste augmentation strategy. We also introduce a new architecture, FORTRESS, which integrates an adaptive Kolmogorov–Arnold Network (KAN), multiple attention mechanisms, and a feature fusion module to enhance feature representation and capture complex relationships and patterns. To further improve efficiency, we replace standard convolutions with depthwise separable convolutions, which significantly reduce the number of trainable parameters without sacrificing accuracy. Finally, we explore the use of few-shot semantic segmentation in the context of structural defect detection. Our results demonstrate its potential and motivate further research in this direction.

\section{Thesis Structure}

The thesis begins by providing a comprehensive summary of the current state-of-the-art in methods for segmentation in structural defect detection, data augmentation, Kolmogorov-Arnold Networks (KANs), and few-shot semantic segmentation in Section~\ref{related_works}. The chapter summarizes the advancements achieved by other researchers, which guided our implementations. Section~\ref{theory} provides the theoretical background of our research, including details about semantic segmentation, various deep learning architectures, data preprocessing steps, and few-shot semantic segmentation. This section establishes a solid foundation for our research and explains the theory behind all components used in this work. Section~\ref{culvert_sewer_defect_dataset} describes our utilized dataset, including data details and the preprocessing pipeline. Preprocessing steps include data augmentation, dynamic label injection and downsampling. We also provide implementation details, as well as a final evaluation, comparing our results with those of baseline models. In Section~\ref{fortress}, we present our new segmentation architecture, FORTRESS, which incorporates an adaptive KAN block, depthwise separable convolutions, and various attention mechanisms. This design achieves superior results on the culvert and sewer pipe defect dataset while keeping the model parameter-efficient. In the implementation details, we provide a comparison to other state-of-the-art models, emphasizing its applicability to structural defect detection. Section~\ref{few_shot_segmentation} describes our first attempt at a few-shot segmentation model and evaluates its performance on structural defect detection tasks. The chapter also includes results achieved by adding attention mechanisms to the prediction head. Finally, Chapter~\ref{conclusion} concludes the thesis by summarizing all insights and findings.

\chapter{\germanenglish{Related Work}{Related Work}}\label{related_works}
%
This section provides information on the development and history of semantic segmentation, data augmentation, Kolmogorov-Arnold Networks, and few-shot learning. It also highlights recent state-of-the-art methods, summarizing advancements achieved by other researchers that form the foundation for our work.

\section{Structural Defect Detection and Segmentation} 

Automated defect detection has advanced significantly over the last years due to new advancements in deep learning technologies and computer vision~\cite{bhattacharya22}. In the early days, they relied heavily on handcrafted features and conventional image processing methods, like edge detection or texture analysis. These techniques obtained valuable information about the characteristics of defects. However, they are unable to adapt to new circumstances and detect defects in real-world scenarios with varying shapes, sizes, backgrounds, and environmental conditions~\cite{alshawi24b}. The introduction of more advanced deep learning methods opened up new possibilities. Convolutional Neural Networks (CNNs)~\cite{lecun1998} introduced a paradigm shift, and more effective methods arose. Following that, Fully Convolutional Networks (FCNs)~\cite{long15} revolutionized the field of semantic segmentation by allowing pixel classification of arbitrarily-sized input images. They constitute the foundation for effective structural defect detection. Encoder-decoder networks like U-Net~\cite{ronneberger2015u}, bottom-up top-down architectures like Feature Pyramid Network (FPN)~\cite{lin2017feature}, and Vision Transformers (ViT)~\cite{dosovitskiy21} are prominent deep learning architectures for semantic segmentation tasks. They are continuously adapted to address different segmentation challenges. \\
\\
U-Net was developed for medical image segmentation, but adapted for various applications, including defect detection. Ronneberger et al.~\cite{ronneberger2015u} integrated skip connections in the U-Net architecture to combine fine-grained details from lower levels with semantic understanding from higher levels to enhance the segmentation process. Su et al.~\cite{su22} integrated Convolutional Block Attention Modules (CBAM) into the U-Net architecture to enhance the crack detection process on bridges. CBAM includes channel and spatial attention mechanisms to allow the model to focus on salient regions and suppress irrelevant parts. Similarly, ASCU-Net~\cite{tong21} uses a triple attention mechanism to improve the segmentation of skin lesions. It combines attention gates, spatial and channel attention, exhibiting an effective border handling while disregarding noise and artifacts in the image. The ACE module, introduced by Xie et al.~\cite{xie19}, incorporates the CBAM module, as well as Squeeze-and-Excitation (SE) blocks~\cite{hu18} and Selective Kernels (SK)~\cite{li19}. This multi-attention approach enhances the feature representation and boosts segmentation performance. Recent advancements by Alshawi et al.~\cite{alshawi23a, alshawi23b, alshawi24a} show promising results in culverts and sewer pipes defect detection by incorporating different deep learning techniques. These include the depth-wise separable U-Net with multiscale filters~\cite{alshawi23a}, which has been used to detect sinkholes. Replacing several standard convolutional layers with sparsely connected convolutional layers that include multiscale filters allows the network to detect defects of different scales, increasing its accuracy while reducing the number of trainable parameters and training time. Alshawi et al.~\cite{alshawi23b} introduced the Dual-Attention U-Net with Feature Injection (DAU-FI~Net). DAU-FI~Net addresses persistent semantic segmentation problems, such as training on multiclass imbalanced datasets with limited data. It improves object localization by detecting patterns at different scales and injecting helpful features at certain stages of the detection process. These models have shown significant improvements in defect detection tasks compared to conventional methods.\\
\\
On the other hand, the FPN architecture is a bottom-up top-down architecture handling multiscale feature extraction better than encoder-decoder networks because of its hierarchical structure. The model's high detection accuracy stems from fusing low-resolution and high-resolution information to capture multi-scale features effectively. The Bi-Directional Feature Pyramid Network (Bi-FPN)~\cite{tan2020efficientdet} follows a bidirectional approach. Unlike FPN, which only allows top-down feature fusion, Bi-FPN allows features to flow in both directions, top-down and bottom-up, with a learnable weighted fusion. NAS-FPN~\cite{ghiasi19} follows a similar approach and utilizes a neural architecture search to find the optimal feature fusion topology. Alshawi et al.~\cite{alshawi24b} introduced the Enhanced Feature Pyramid Network (E-FPN) in 2024, an architecture specifically designed for culvert and sewer pipe defect detection. It incorporates Inception blocks and depth-wise separable convolutions to extract multi-scale features and handle object variations efficiently. They further demonstrated the advantages of class decomposition and data augmentation to mitigate the problem of imbalanced class distribution in a dataset. \\
\\
Dosovitskiy et al.~\cite{dosovitskiy21} adapted the transformer framework from natural language processing tasks and made it suitable for computer vision applications. Vision Transformer (ViT) is a novel method that differs significantly from conventional approaches. It includes self-attention, which allows the model to capture global context and long-range relationships, expanding the receptive field to the whole image. This characteristic enables the model to focus on important features while neglecting irrelevant areas by dynamically weighting features regarding their importance. Zheng et al.~\cite{zheng21} adapted the idea of ViT to make it applicable to semantic segmentation tasks and proposed the model SETR in 2021. Its backbone consists of multiple CNN decoders to upsample features and create pixel-level predictions. SwinTransformer~\cite{liu21swin} is a multipurpose backbone for computer vision tasks that uses shifted windows to provide a more efficient self-attention mechanism. Although SwinTransformer can be applied to semantic segmentation tasks, it focuses more on the encoding part. SegFormer~\cite{xie2021segformer}, proposed by Xie et al., includes a lightweight All-MLP decoder to combine local and global information. First, a hierarchical transformer encoder creates a hierarchical feature representation that is similar to the multi-level features provided by a CNN framework. This new encoder structure preserves spatial information effectively and makes position embeddings unnecessary, on which previous methods heavily relied. Furthermore, SegFormer uses a lightweight decoder consisting of multiple MLP layers. This simple structure of solely using MLP blocks is sufficient because the hierarchical transformer provides a larger receptive field than traditional CNN encoders.

\section{Data Augmentation} 
Data augmentations are commonly used methods to boost the performance of a model while not adjusting the model itself, but by improving the quantity and quality of the dataset. Methods such as color adjustments~\cite{jurio10, krizhevsky12, chatfield14, wu15, taylor18}, geometric transformations~\cite{simard03, chatfield14, girshick15, taylor18, wang20b}, noise injections~\cite{moreno18, lopes19, akbiyik23} and sample mixing~\cite{inoue18, zhang18, yun19, takahashi19, summers19} have shown great potential and were able to achieve state-of-the-art results. Apart from standard augmentation techniques, copy-paste augmentation generates new samples by extracting instances from images using the provided labels and combining them with new backgrounds. In 2011, Karsch et al.~\cite{karsch11} had already proposed a promising semi-automatic approach to render virtual objects into new images while taking scene geometry and lighting into account. However, this approach required manual annotation of surfaces and light sources. In 2016, Gupta et al.~\cite{gupta16} introduced the idea of adding text to new backgrounds to create new training samples and improve the overall performance of text detection models overall. They estimated local surface normals and generated a pixel-wise depth map to align the text with the scene geometry while considering object boundaries. Dvornik et al.~\cite{dvornik18} performed context-aware copy-paste augmentation by combining real-world instances with backgrounds, while considering the presence of objects in the scene. They emphasized that the context in which an object normally appears is as important as the object itself when synthesizing new data samples.  Furthermore, multiple approaches~\cite{dwibedi17, fang19, ghiasi21} have demonstrated the effectiveness of simply copying and pasting instances to random positions, which simplifies the data synthesis process further. Data augmentation provides significant benefits for relatively small datasets, which are commonly encountered in structural defect detection, and enhances the generalizability and robustness of models~\cite{alshawi24b}.

\section{Kolmogorov-Arnold Networks and Function Composition}
The introduction of Kolmogorov-Arnold Networks (KANs) has drawn attention to a more parameter-efficient and theory-based alternative to conventional deep learning models. The fundamental theory behind KANs is the Kolmogorov-Arnold Representation Theorem~\cite{kolmogorov1961}. This theorem states that any continuous multivariate function can be represented by a finite number of continuous univariate functions. While traditional networks heavily rely on linear transformations followed by an activation function for function approximation, KANs take a completely different approach. KANs perform function composition and represent complex multivariate relationships as a combination of simpler, learnable one-dimensional functions~\cite{teymoor25}. This allows the model to learn complex patterns, enhance feature representation and increase interpretability~\cite{liu24}. Liu et al.~\cite{liu24} presented the first practical implementation of KAN using spline-based function approximation and demonstrated its superior performance compared to traditional state-of-the-art models on various tasks. In recent years, multiple variations have been introduced. Chebyshev KAN~\cite{ss2024} uses a Chebyshev polynomial to further improve the nonlinear function approximation, offering a promising extension of KAN. FastKAN~\cite{li24} demonstrates that the B-spline functions in KAN can be approximated using Gaussian radial basis functions, reducing computational cost and making the model more appealing for real-world applications. Convolutional Kolmogorov-Arnold Networks (ConvKAN)~\cite{bodner24} incorporate KAN principles into convolution layers by replacing linear weights with learnable nonlinear B-spline functions. ConvKAN achieves comparable results to the traditional implementation while requiring only half of the parameters previously needed. Furthermore, KAN has been successfully employed for time-series forecasting~\cite{genet25} and hyperspectral space detection~\cite{seydi25}, effectively handling data variations and complex nonlinear patterns. MonoKAN~\cite{polo24} replaces the B-spline functions in KAN with cubic Hermite splines to enforce monotonicity constraints. This enables the model to follow expert knowledge and common-sense rules, while improving the interpretability and transparency of its decisions. The Kolmogorov-Arnold Transformer (KAT)~\cite{yang24} substitutes MLP layers in a transformer architecture with KAN layers to address issues like value initialization and computational inefficiency. On the other hand, Kolmogorov-Arnold Networks with Interactive Convolutional Elements (KANICE)~\cite{ferdaus24} incorporates KAN principles and Interactive Convolutional Blocks (ICB) into a CNN architecture to allow complex and non-linear feature representation and dynamic feature extraction. KANICE outperforms current state-of-the-art methods on various datasets. Krzywda et al.~\cite{krzywda25} tested the application of KAN on metal surface defect detection and recorded superior accuracy compared to baseline CNN architectures while using fewer parameters. However, the implementation and application of KAN in structural defect detection and infrastructure inspection is rather unexplored, which offers new research opportunities~\cite{krzywda25}.

\section{Few-Shot Semantic Segmentation}
Few-shot learning has gained increasing interest in the last few years. Initially, most approaches focused on image classification~\cite{snell17, satorras18, sung18}. However, in recent years, attention has shifted to segmentation tasks~\cite{shaban17, wang19, lin23}. In 2017, Shaban et al.~\cite{shaban17} introduced the first conditional network performing few-shot learning for pixel-level classification. They demonstrated that their proposed model achieved superior performance compared to different baseline models and efficiently segments novel classes with access to only one annotated example. In 2018, Zhang et al.~\cite{zhang20} introduced SG-One, a one-shot segmentation model that uses masked average pooling to extract guidance features and cosine similarity to generate similarity maps between support and query features. This approach is similar to prototypical networks, which also compare support and query features to guide segmentation. Prototypical networks represent each class with a prototype feature vector extracted from the support set via masked average pooling and classify query samples based on their similarity to these prototypes. They were first introduced for image classification by Snell et al.~\cite{snell17}, and Dong et al.~\cite{dong18} were the first to adapt them for semantic segmentation. However, due to the difficulty of extracting suitable prototypes and the model's overall complexity, Wang et al.~\cite{wang19} developed PANet. PANet is a simpler design that laid the foundation for all future prototypical networks in semantic segmentation. Several approaches have been proposed to improve few-shot semantic segmentation by aligning support and query features more effectively. The Class-Agnostic Segmentation Network (CANet)~\cite{zhang19} demonstrated the effectiveness of iterative refinement in few-shot learning. This method gradually improves segmentation predictions and achieves strong generalization across classes. The Prior Guided Feature Enrichment Network (PFENet)~\cite{tian20} introduces a feature enrichment module that fuses semantic and spatial knowledge from support features and prior masks with query features. Prior masks add spatial information to query features and guide the segmentation process by indicating where an object may be located. With the rise of transformers, attention mechanisms have become increasingly popular in various application areas, including few-shot learning. Hou et al.~\cite{hou19} introduced the Cross-Attention Network in 2019, and showed that cross-attention between support and query features allows the classification model to focus on task-relevant image regions and produce better feature embeddings. Building on this idea, the Multi-Scale Cross-Attention (MSCA)~\cite{ren2025} model was introduced for few-shot segmentation. It integrates multi-scale cross-attention between support and query features, and aligns features at multiple levels. A slightly different approach is CAT-Net~\cite{lin23}. CAT-Net uses a transformer with cross-attention as its backbone, while also incorporating an iterative refinement to enhance the predicted segmentation masks. Few-shot semantic segmentation also offers advantages for defect detection. LERENet~\cite{ding24} addresses the high intra-class variation of defects on metal surfaces using a multi-prototype approach. Representing one class with multiple prototypes guides the segmentation process more effectively and achieves more precise predictions. Lastly, Shi et al.~\cite{shi23} propose comparing defect and defect-free images to improve defect localization. This strategy mitigates the impact of cluttered backgrounds and demonstrates strong generalization.

\chapter{\germanenglish{Theorie}{Theoretical Background}}\label{theory}
This section provides the theoretical background necessary for our research. We begin by introducing the fundamental concepts of semantic segmentation, followed by an overview of popular semantic segmentation models and advancements for structural defect detection. Next, we introduce different data augmentation techniques as well as dynamic label injection to handle small and imbalanced datasets. Finally, we discuss few-shot semantic segmentation and include descriptions of different network types. These sections form the foundation for understanding our proposed approaches.

\section{\germanenglish{Semantic Segmentation}{Semantic Segmentation}}
\begin{figure}[H]
    \centering   %
\setkeys{Gin}{width=\linewidth}
\begin{subfigure}{.3\textwidth}
    \centering
    \captionsetup{justification=centering,margin=0cm}
    \includegraphics[width=0.6\linewidth]{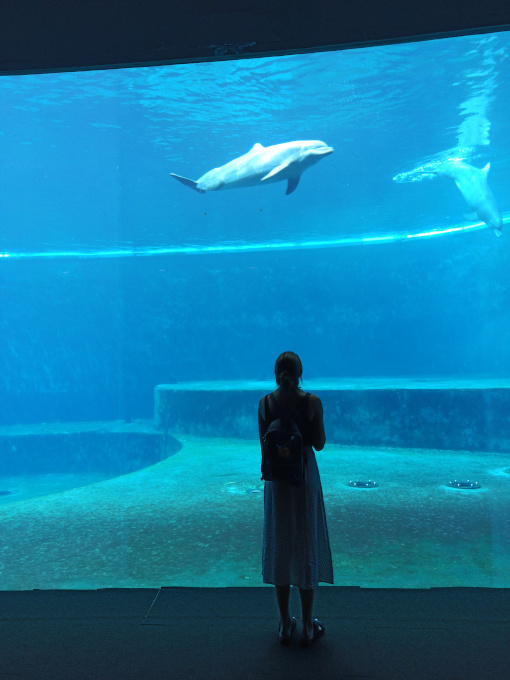}
    \captionsetup{width=0.9\linewidth}
    \caption{Original}
    \label{fig:original}
\end{subfigure}
\begin{subfigure}{.3\textwidth}
    \centering
    \captionsetup{justification=centering,margin=0cm}
    \includegraphics[width=0.6\linewidth]{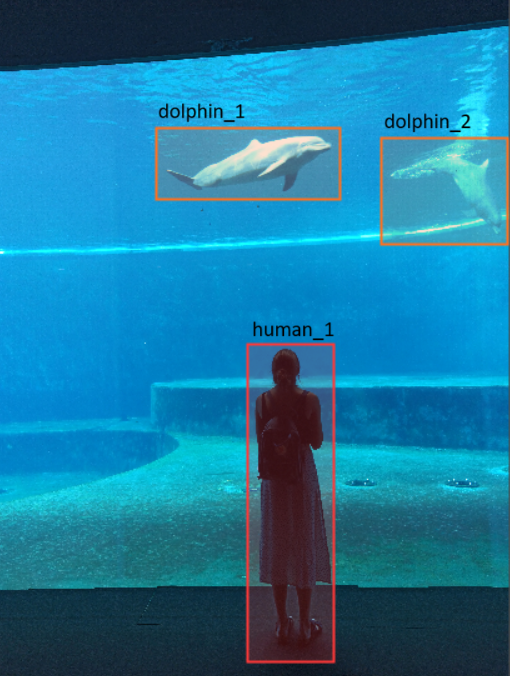}
    \captionsetup{width=0.9\textwidth}
    \caption{Object Detection}
    \label{fig:object_detection}
\end{subfigure}
\begin{subfigure}{.3\textwidth}
    \centering
    \captionsetup{justification=centering,margin=0cm}
    \includegraphics[width=0.6\linewidth]{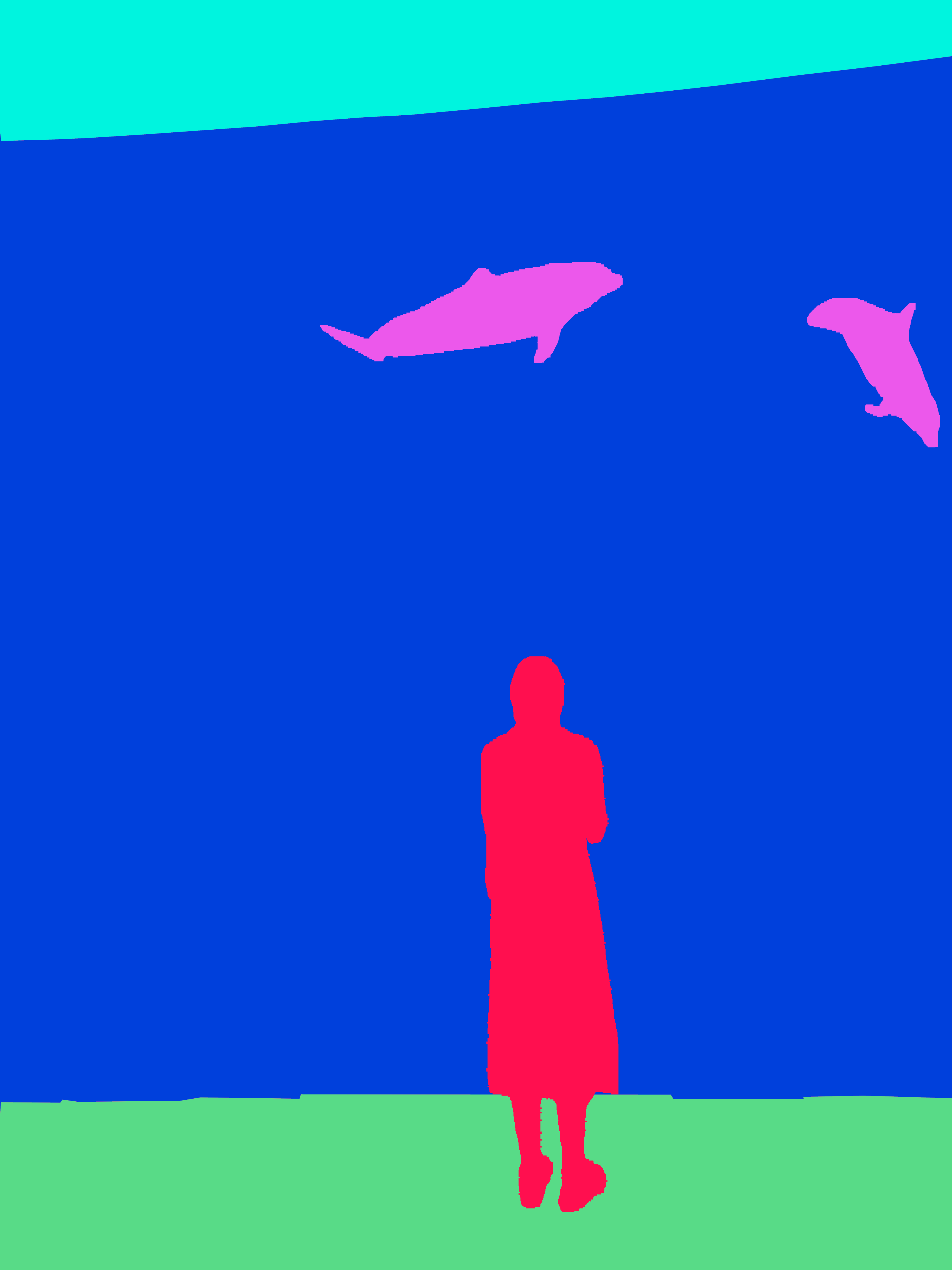}
    \captionsetup{width=0.9\textwidth}
    \caption{Semantic Segmentation}
    \label{fig:semantic_segmentation}
\end{subfigure} \\
\begin{subfigure}{.3\textwidth}
    \centering
    \captionsetup{justification=centering,margin=0cm}
    \includegraphics[width=0.6\linewidth]{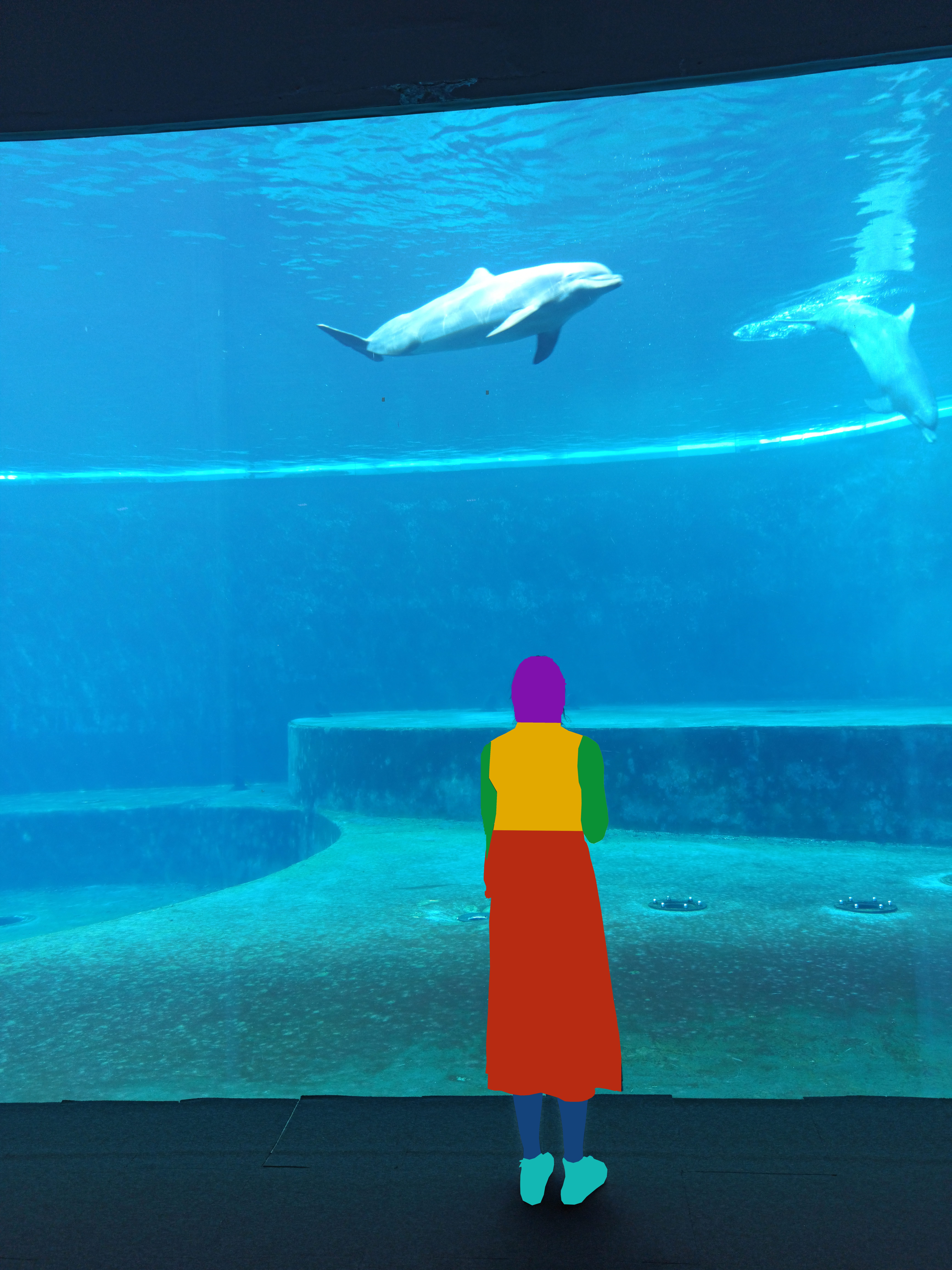}
    \captionsetup{width=0.9\linewidth}
    \caption{Part Segmentation}
    \label{fig:parts_segmentation}
\end{subfigure}
\begin{subfigure}{.3\textwidth}
    \centering
    \captionsetup{justification=centering,margin=0cm}
    \includegraphics[width=0.6\linewidth]{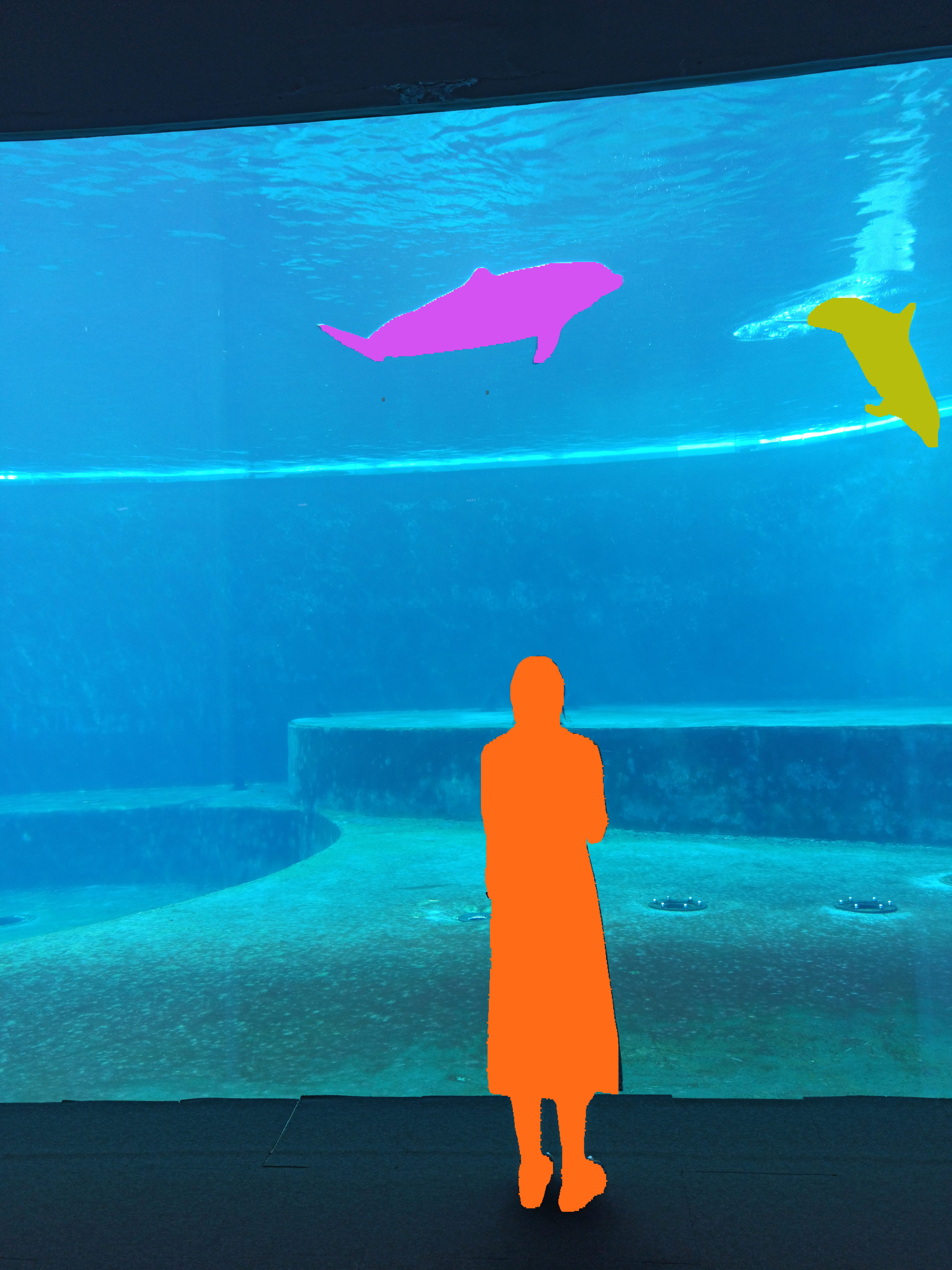}
    \captionsetup{width=0.9\textwidth}
    \caption{Instance Segmentation}
    \label{fig:instance_segmentation}
\end{subfigure}
\begin{subfigure}{.3\textwidth}
    \centering
    \captionsetup{justification=centering,margin=0cm}
    \includegraphics[width=0.6\linewidth]{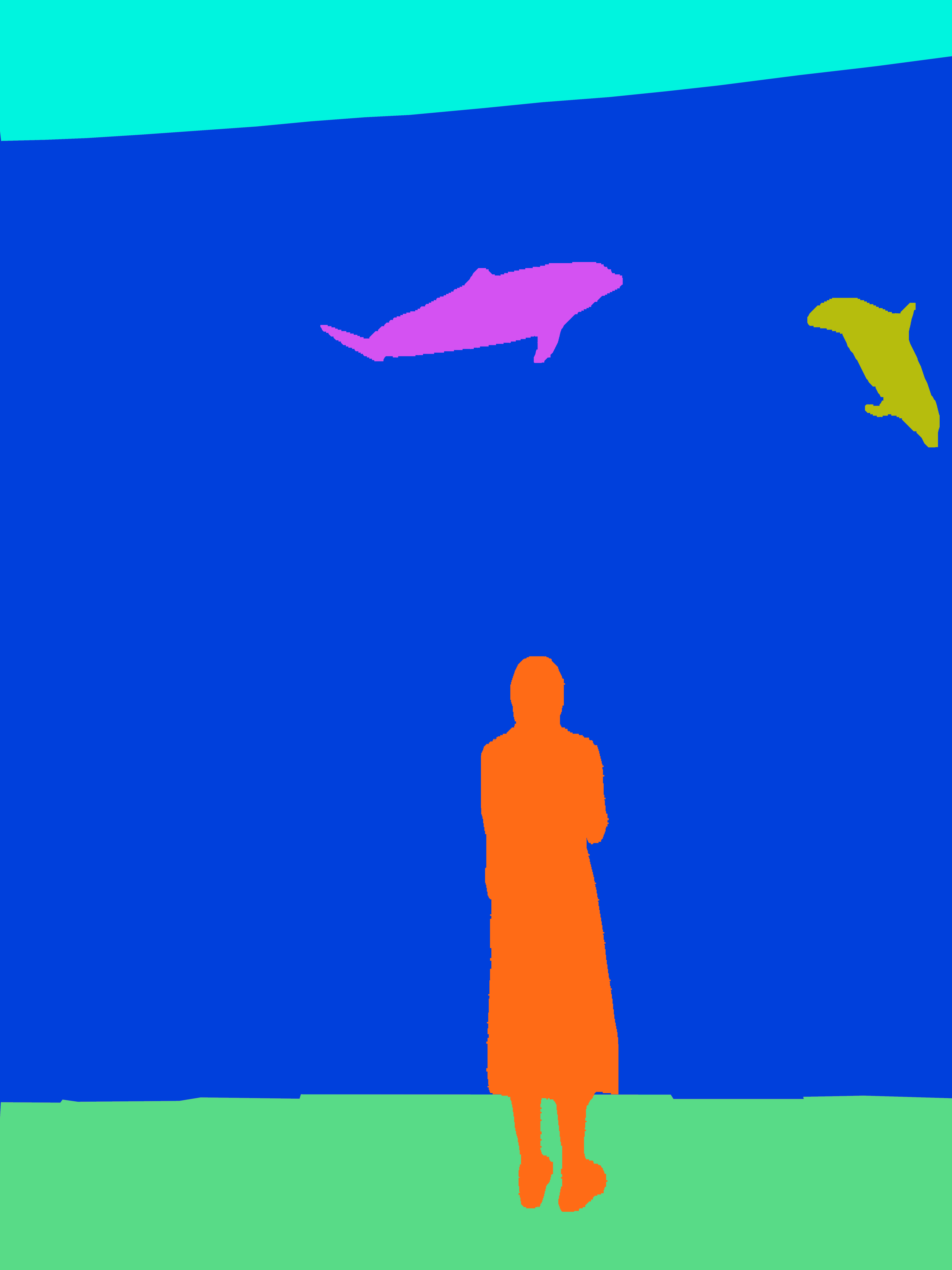}
    \captionsetup{width=0.9\textwidth}
    \caption{Panoptic Segmentation}
    \label{fig:panoptic_segmentation}
\end{subfigure} \\
\caption{Visualization of different computer vision tasks.}
\label{fig:comp_segmentation}
\end{figure}

Semantic segmentation plays a crucial role in computer vision and allows a deeper understanding of visual data. It creates a segmentation mask for an image by assigning exactly one class label to each pixel. This separates the picture into distinct regions and group together pixels that belong to the same object. A segmentation of an image results in a finite set of regions $R$~\cite{sonka13}:
\begin{equation}
    R = \bigcup_{i=1}^{N} R_i, \quad \text{where } R_i \cap R_j = \emptyset \quad \text{for all } i \ne j,\ i, j \in \{1, \dots, N\}.
\end{equation}
Humans are capable of performing segmentation without the need of explicitly knowing the objects in a scene. A model that understands visual context and can distinguish between different objects and regions without having encountered them before is a very powerful tool~\cite{guo18}. Semantic segmentation holds great potential as it offers strong contextual understanding of an image, while not include any analysis or interpretation. It can therefore be used as an intermediate step for other tasks. Semantic segmentation is closely interconnected with tasks such as image classification, object detection, and various segmentation techniques~\cite{catalano23}. The differences between these methods are illustrated in Figure~\ref{fig:comp_segmentation}.\\
\\
Image classification assigns one or more labels to an image to describe its overall content. Object detection~(\ref{fig:object_detection}) locates objects in an image, draws bounding boxes around them, and assigns class labels. Semantic segmentation~(\ref{fig:semantic_segmentation}) operates at the pixel level. It divides the image into separate regions by assigning each pixel to its semantic category. Part segmentation~(\ref{fig:parts_segmentation}) focuses on a single subject in the image and identifies its individual parts~\cite{catalano23}. Instance segmentation~(\ref{fig:instance_segmentation}) detects all relevant objects in an image and separates each individual object into a distinct region. It creates precise pixel-level masks showing their visible parts. Panoptic segmentation~(\ref{fig:panoptic_segmentation}) combines the concepts of semantic segmentation and instance segmentation. It assigns a class label to every pixel and distinguishes individual object instances~\cite{szeliski22}.


\section{Deep Learning Architectures for Semantic Segmentation}

\begin{figure}[H]
\centering
\captionsetup{justification=centering,margin=0cm}
\includegraphics[width=0.5\textwidth]{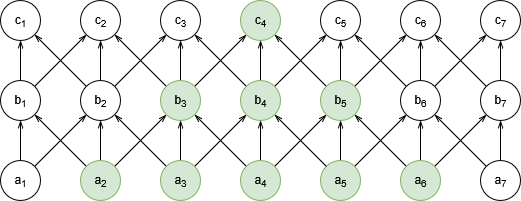}
\captionsetup{width=1.0\textwidth}
\caption{Although direct connections in a CNN are sparse, there are multiple indirectly connected elements that increase the receptive field for each subsequent layer.}
\label{fig:receptive_field}
\end{figure}

Numerous research was done to find new and innovative semantic segmentation algorithms. In the beginning, methods like thresholding~\cite{otsu1975}, region merging~\cite{nock04}, and k-means clustering~\cite{dhanachandra15} were popular choices. However, utilizing deep learning models outperformed all previous techniques and introduced new possibilities concerning semantic segmentation tasks~\cite{minaee22}.\\
\\
In 1998, LeCun et al.~\cite{lecun1998} introduced the first convolutional neural network (CNN) architecture used for document recognition. CNNs are a special kind of neural network that utilize the convolution operation instead of matrix multiplication in at least one of its layers, and they have received great appreciation in the computer vision field. Each convolution layer includes three stages: convolution stage, detector stage, and pooling stage. First, the convolution stage applies a convolution operation on the input to extract features as a set of linear activations. Next, the detector stage applies a nonlinear activation function on the extracted features. Lastly, the pooling stage reduces the number of spatial dimensions by aggregating statistical information, like max or mean, of a certain neighborhood. This information is then handed over to the next layer or used for the computation of the final output. The elements of different layers are locally connected, which means that the units from a certain layer get weighted inputs from a neighborhood of units from the previous layer. The neighborhood is also referred to as the receptive field. By increasing the number of layers, the receptive field is expanded, as visualized in Figure~\ref{fig:receptive_field}. Shallow layers learn features from a smaller receptive field than deeper layers. Although the units are sparsely connected, more elements are indirectly connected in deeper layers, resulting in a wider receptive field. After the last convolutional layer, fully connected layers produce the final predictions using the features extracted from the convolutional layers. In a fully connected layer, each unit is connected with each unit of the previous layer, allowing the model to make predictions using global context. An additional benefit of CNNs compared to traditional neural networks is that their parameters are shared within each layer. This means that the kernel contains learnable parameters that are used for every location to which the kernel is applied. This significantly reduces the number of learnable parameters, increasing efficiency and reducing memory requirements~\cite{goodfellow16cnn}. CNNs have already existed for a very long time~\cite{lecun89}. However, the first major breakthrough was achieved by Krizhevsky et al.~\cite{krizhevsky12} in 2012, by using a model consisting of eight layers and millions of parameters. It was trained on the ImageNet~\cite{deng09} dataset, consisting of a million images. Since then, larger models have been introduced that are trained on even more data to surpass previous state-of-the-art approaches. Although CNNs were originally introduced for classification tasks, they have also been used for semantic segmentation. In semantic segmentation tasks, it is often not feasible to have a large dataset. Therefore, techniques such as data augmentation are employed to improve the model's performance~\cite{ronneberger2015u}.

\noindent
CNNs have advanced the fields of object detection, image correspondence and part prediction. A few earlier approaches~\cite{farabet13, ganin14, gupta14, hariharan14, pinheiro14} utilized CNNs for semantic segmentation tasks by marking each pixel with the class label of the object or region enclosing the pixel. In 2015, Long et al.~\cite{long15} proposed a fully convolutional network (FCN) that performs end-to-end pixel-wise classification. This network outperformed all state-of-the-art methods at the time. Replacing all fully connected layers with convolutional ones allows the network to take an arbitrary-sized input and produce a segmentation mask with the same dimensions. Skip connections are used to combine feature maps from the final output layer with those from previous layers by upsampling and fusing them. Thus, the final output considers semantic information from deep layers and details about the visual appearance from shallow layers~\cite{minaee22}. \\
\\
FCNs such as U-Net~\cite{ronneberger2015u} and the Feature Pyramid Network (FPN)~\cite{lin2017feature} are common choices for models in semantic segmentation tasks. However, Vision Transformers (ViT) \cite{vaswani17, dosovitskiy21, liu21swin} also demonstrate great potential and are a popular alternative to CNNs. The following section explains these different architectures in more detail.

\subsection{Encoder-Decoder Network}\label{edn}
An encoder-decoder network (EDN) is a two-stage network consisting of an encoder and a decoder. The model learns a mapping from the input domain to the output domain, where first the encoder $z = f(x)$ creates a latent representation of the input $x$. The feature vector $z$ outlines the semantic characteristics of the input and is handed over to the decoder. The decoder $y = g(z)$ uses the latent representation of the input and creates the final output $y$. The output can be, for example, a denoised image or a segmentation mask~\cite{minaee22}.

\begin{figure}[H]
\centering
\captionsetup{justification=centering,margin=0cm}
\includegraphics[width=1.0\textwidth]{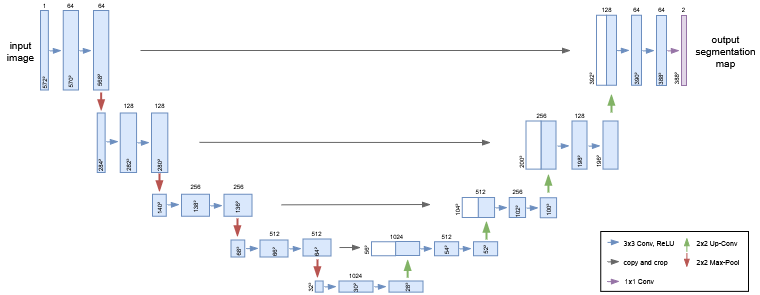}
\captionsetup{width=1.0\textwidth}
\caption{Visualization of the U-Net architecture where the input image is downsampled to 32x32 pixels in the contracting path. The blue boxes refer to multi-channel feature maps whereas the white ones represent the copied
feature maps from the contracting path to the expanding one. The number of channels are denoted on the top of the feature map and its size is stated inside or on its left~\cite{ronneberger2015u}.}
\label{fig:unet}
\end{figure}

\noindent
The U-Net~\cite{ronneberger2015u} architecture was introduced in 2015 for biomedical image segmentation. Building on the fundamental idea of FCN, it proposes an architecture that is more suitable for smaller training sets while still achieving state-of-the-art results. Figure~\ref{fig:unet} visualizes the U-Net architecture and its individual components and operations. The architecture consists of two steps: the contracting path and the expanding path. The contracting path extracts features and identifies the context of the input image. The expanding path is used for localization and generating the corresponding segmentation mask. The contracting path follows the architecture of a CNN and includes multiple downsampling steps. Each step consists of two $3\times3$ convolution operations, each followed by a rectified linear unit (ReLU) and a $2\times2$ max pooling operation. The number of feature channels doubles with each step to enable the model to learn more complex features. During the expanding path, $2\times2$ upsampling convolutions halve the number of feature channels again to restore the input dimensions for the final output. Additionally, the corresponding feature map from the downsampling path is copied, cropped, and fused with the upsampled map to improve localization. The copied feature map from the contracting path must be cropped because convolution causes the loss of border pixel information. As a last step, a $1\times1$ convolution maps the final feature vectors to the intended number of classes, generating the final segmentation mask that stores a class label for each pixel~\cite{minaee22}. \\
\\
Encoder-decoder architectures also have limitations. During the encoding step, fine-grained information that may be relevant for distinguishing between classes and creating a precise segmentation mask can get lost~\cite{minaee22}. Additionally, the model may have difficulty detecting features at different scales and handling variation in visual appearance~\cite{alshawi24b, alshawi24a}.

\subsection{Feature Pyramid Network}\label{theory_fpn}
\begin{figure}[H]
\centering
\captionsetup{justification=centering,margin=0cm}
\includegraphics[width=0.8\textwidth]{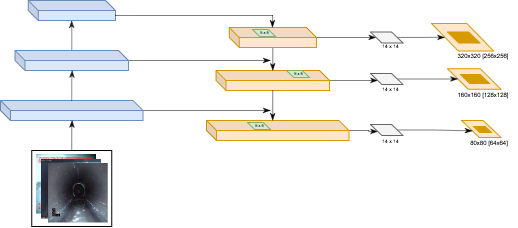}
\captionsetup{width=1.0\textwidth}
\caption{Visualization of the FPN architecture. The blue rectangles represent the feature maps at different pyramid levels. At each level, a small Multi-Layer Perceptron (MLP) is applied to sliding windows of size $5 \times 5$, shown as green boxes, to generate dense object proposals, each producing a segmentation mask of size $14 \times 14$, visualized in grey. The orange rectangles indicate the size of the image region that each proposal corresponds to. The light orange areas represent the image region size, while the dark orange areas visualize the typical object size at each pyramid level. This shows how deeper pyramid levels consider larger receptive fields~\cite{lin2017feature}.}
\label{fig:fpn}
\end{figure}

Lin et al.~\cite{lin2017feature} demonstrated the effectiveness of multi-scale analysis in a deep learning architecture with the Feature Pyramid Network (FPN). The main concept behind FPN is to create a scale-invariant feature representation, where variations in an object's scale are handled by shifting its level within the pyramid. This allows the model to detect objects of different sizes by scanning across various positions and pyramid levels. Although, FPN was originally developed for object detection, its application area has since been extended to semantic segmentation. It exploits the hierarchical, pyramid-like structure of deep CNNs to generate feature pyramids with minimal computational overhead. The pyramidal structure of deep convolutional neural networks (CNNs) provides feature maps at different resolutions. Shallow layers produce high-resolution features suitable for localization, but they do not provide semantic meaning. Deeper layers, on the other hand, produce low-resolution feature maps with strong semantic meaning but are less appropriate for localization tasks. FPN introduces a top-down pathway with additional lateral connections to combine low- and high-resolution feature maps, closing the semantic gap. This creates a feature pyramid with semantically rich features at each level~\cite{lin2017feature}.\\
\\
The FPN consists of two pathways: the bottom-up pathway and the top-down pathway. Its architecture is illustrated in Figure~\ref{fig:fpn}. The bottom-up pathway uses a deep CNN backbone to create a feature hierarchy, including feature maps of different scales. The top-down pathway, in combination with lateral connections, creates semantically strong feature maps at all levels. Spatially coarser, yet semantically meaningful, features from higher pyramid levels are upsampled and merged with feature maps from the bottom-up pathway via lateral connections. The bottom-up feature maps provide important spatial information relevant for localization. These feature maps are upsampled by a factor of two using the nearest neighbor method. The resulting feature map is then merged with the corresponding feature map from the bottom-up pathway via element-wise addition. Finally, a 3×3 convolution is applied to each merged feature map to create the final predictions and minimize the aliasing effect resulting from the upsampling process~\cite{lin2017feature}.

\noindent
FPNs struggle with class imbalance. A small number of positives and a high number of background pixels cause a significant imbalance. In that case, FPNs have difficulty focusing on the small number of positives, which can negatively impact their performance~\cite{gao18}.

\subsection{Vision Transformer}

\begin{figure}[H]
\centering
\captionsetup{justification=centering,margin=0cm}
\includegraphics[width=0.8\textwidth]{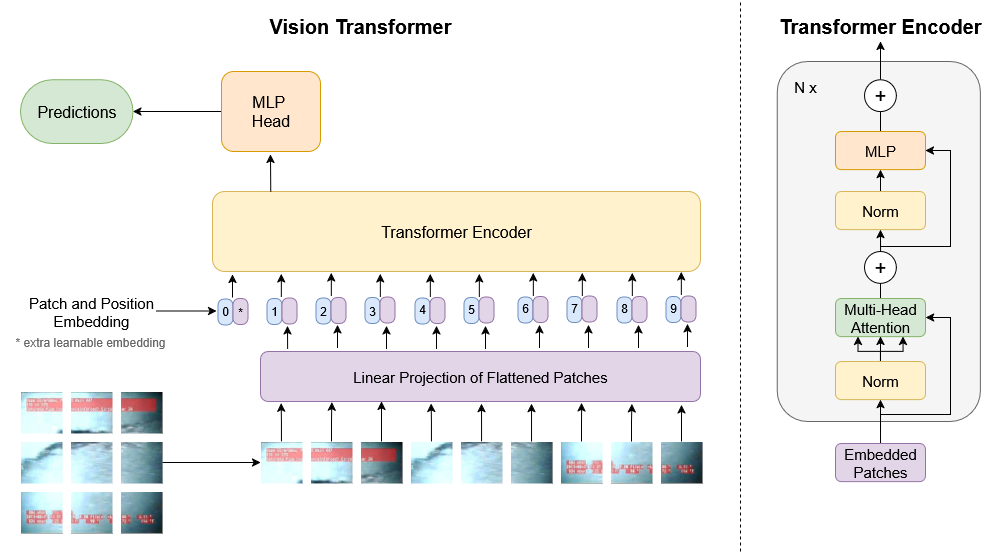}
\captionsetup{width=1.0\textwidth}
\caption{Visualization of the ViT architecture. The input image is divided into multiple image patches. For each patch a linear embedding with an added position embedding is handed over to the transformer encoder to create a rich feature representation. The transfomer encoder consists of multiple alternating multi-head self-attention and MLP blocks. An MLP head creates the final class predictions~\cite{dosovitskiy21}.}
\label{fig:vit}
\end{figure}

Vision Transformers (ViT) are a popular choice for computer vision tasks nowadays especially for object detection and image classification. They incorporate a self attention mechanism allowing the model to integrate global context even in the lower layers. Initially, transformers were proposed by Vasawani et al.~\cite{vaswani17} for natural language processing (NLP) tasks. They are often pretrained on a large dataset and then finetuned using a smaller dataset tailored to the specific task. Transformers convince with their self-attention mechanism, allowing to consider the global context of an image even in lower layers and across image patches. Dosovitskiy et al.~\cite{dosovitskiy21} demonstrated the applicability of transformers to computer vision tasks, proposing the first transformer-based vision model developed for image classification. They split the input images into image patches, and for each patch, a linear embedding is computed. The linear embedding is fed into a transformer encoder alongside a position embedding. Position embeddings are added to provide spatial information to the model. The transformer encoder consists of multiple blocks each including a multi-head self-attention and Multilayer Perceptron (MLP) block~\cite{dosovitskiy21}. The rich feature representations are then handed over to an MLP head for the final classification, as shown in Figure~\ref{fig:vit}.

\noindent
Self-attention takes a sequence of values as an input and computes for each value a weighted sum over all values of the sequence, where the weights are defined as the similarity of the two values. Multihead self-attention extends this concept by performing multiple parallel self-attention operations. These operations are also called heads, and each head focuses on different aspects and relationships. Their insights are combined by concatenating the individual results and performing a linear projection~\cite{dosovitskiy21}.\\
\\
Transformers handle images more globally than CNNs do, but they lack some of the helpful inductive biases that a CNN naturally possess. CNNs are translation equivariant, which helps them to identify objects even when their position in the image changes. Furthermore, they first focus on small neighborhoods to effectively extract fine-grained details and local patterns. In a ViT, however, only the MLP layers provide these characteristics. The multi-scale attention mechanism is global and lacks these inductive biases. For this reason, they require access to a lot of data to generalize well~\cite{dosovitskiy21}.

\subsection{Improvements relevant for Infrastructure Inspection}

Encoder-decoder networks (EDNs), feature pyramid networks (FPNs), and vision transformers perform well on standard segmentation tasks. However, additional challenges arise in water infrastructure systems that significantly harm their performance. Culverts and sewer pipes vary in their size, shape, and overall visual appearance due to different environmental conditions. Occlusion, debris, vegetation, and changing lighting conditions affect the model's performance enormously. Furthermore, defects are often small and hard to detect. Models have to be able to capture small defects as well as the entire pipe structure to provide accurate results. EDNs have difficulty identifying features at different scales, like detecting small defects while also considering larger pipe characteristics. FPNs handle this challenge well but struggle with data imbalance and prioritize the majority class in predictions. ViTs, on the other hand, lack some of the inductive biases that CNNs include by nature and are dependent on a large amount of data. The dataset size is a crucial point for culvert and sewer pipe inspection because collecting data is cumbersome. Also, the two challenges, class imbalance and scale variance, often arise in defect datasets and must be addressed. In recent years, different enhancements have been introduced in the last few years to improve the performance of segmentation models in sophisticated settings~\cite{alshawi24a}.\\
\\
\textbf{Attention Gates.} Attention Gates allow models to implicitly learn to focus on salient regions and ignore irrelevant ones.  The model becomes more sensitive and achieves a higher accuracy, while only a minimal computational overhead was added. They can be incorporated into various CNN architectures. In 2018, Oktay et al.~\cite{oktay18} introduced Attention U-Net, the first architecture integrating Attention Gates to a U-Net model for an efficient and accurate dense label predictions. They positioned the gates at skip connections to filter features using contextual information extracted from coarser representations. Attention gates do not require supervision and automatically learn to focus on specific structures and features. They are especially useful when small objects vary greatly in shape. Traditional models have difficulty identifying them correctly. Attention Gates are a valuable addition that allows proper detection with only a small computational overhead~\cite{oktay18}.\\
\\
\textbf{Inception Blocks and Residual Connections.} Inception Blocks allow multi-scale feature extraction by performing multiple convolution operations with different filter sizes in parallel. They often consist of $1\times1$, $3\times3$, $5\times5$ convolutions, as well as a separate max pooling operation. Their outputs are concatenated to achieve a richer feature representation. The $1\times1$ convolution is used for channel-wise pooling, while the $3\times3$ and $5\times5$ filters are used to capture features at different scales~\cite{alshawi23a}. Max pooling is a popular technique in several neural networks that reduces spatial dimensions, increases the receptive field in later layers, and introduces translation invariance~\cite{gholamalinezhad20}. Residual connections mitigate gradient-related issues by adding the input to the computed output, reducing vanishing gradients and training time~\cite{he16, szegedy17}. Combining inception blocks and residual connections results in faster and more efficient training of deep neural networks~\cite{alshawi24b}.  \\
\\
\textbf{Depthwise Separable Convolution.} Depthwise separable convolution is a factorized convolution that splits a standard convolution operation into two operations, namely into a depthwise convolution and a $1\times1$ convolution, also called pointwise convolution. First, the depthwise convolution processes each input channel individually by applying a separate filter to each channel. Then, the pointwise convolution combines the outputs. Standard convolutions perform filtering and aggregation in one step, whereas depthwise separable convolutions use two layers. This factorization substantially reduces the number of parameters and computational effort without affecting the performance negatively~\cite{howard17}.   \\
\\
\textbf{Squeeze-and-Excitation Blocks.} The convolution operation is a fundamental building block of CNNs that fuses spatial and channel-wise information to create representative features. Squeeze-and-Excitation (SE)~\cite{hu18} blocks use adaptive recalibration to improve feature responses per channel. Channel descriptors gather global information for each channel and recalibrate feature responses, highlighting relevant ones and suppressing less important ones. Consequently, the model focuses more on crucial channel-wise features leading to an overall enhanced performance~\cite{alshawi24b}.

\section{\germanenglish{Datenerweiterung}{Data Augmentation}}

A network needs access to a large amount of data. Otherwise, the risk of overfitting is significantly high. Overfitting occurs when the network learns a function that aligns perfectly with the training data points but has high variance and performs poorly on unseen data. The model memorizes the training data instead of learning patterns that apply to unseen data. Collecting a large dataset is often not feasible, since the process can be difficult and expensive. In addition to collecting data samples, the data needs to be annotated, which is often done manually. The manual annotation is very time-consuming and requires domain knowledge. Data augmentation offers a simple but effective solution by increasing the size and diversity of the training set. This helps train a more generalizable model, while reducing the risk of overfitting. The generalizability of a model refers to its performance on unseen data compared to already seen data. A model generalizes poorly when it performs well on the training data but poorly on the test data, indicating that it has overfitted on the training set. Data augmentation not only increases the size of the training set but also enhances its quality. The model can learn more meaningful information and key features from the training data. In real-time applications, it must be able to handle challenges such as different viewpoints, changing lighting, varying backgrounds, object sizes and proportions, as well as occlusions. Data augmentation mitigates these challenges. By incorporating more diverse data, the difference between the training and test sets is minimized, leading to improved performance. Augmenting data is an effective way to prevent overfitting by focusing not on the model architecture, but on the dataset itself, which is often the root of the problem. In the early days, the first data augmentation techniques applied were simple transformations such as horizontal flipping and color space adjustments. These early methods demonstrated great potential and paved the way for more advanced augmentation techniques~\cite{shorten19}.\\
\\
It is also important to mention the disadvantages of performing data augmentation, which need to be taken into account. First, applying data augmentation is time-consuming and computationally expensive. The final training dataset requires more memory than before. For instance, performing a single augmentation technique on a dataset of size $N$ increases its size to $2N$. With a larger dataset, the computational cost of training increases, and the training takes longer. However, data augmentation is an effective strategy to improve the generalizability and performance of a model. The optimal balance must be found between fast, resource-efficient training and a larger, more diverse dataset that yields better results. Ideally, a small set of complementary augmentation techniques should be selected, while keeping the number of additional samples as low as possible~\cite{shorten19}.\\
\\
There are different photometric and geometric transformations that are commonly used for data augmentation. Some popular techniques that were also used on our dataset are described below.\\
\\
\textbf{Horizontal and Vertical Flip.}
Horizontal flipping is a common technique that is easy to implement. It is used more frequently than vertical flipping and yields better results. Horizontal flipping introduces realistic variations, whereas vertical flipping often produces unnatural-looking images, which may be irrelevant in some application domains~\cite{shorten19}.\\
\\
\textbf{Color Space Adjustments.}
Images are commonly described by their dimensions (height x width x channels), where channels refers to the number of color channels, which is usually set to one or three. One channel indicates a binary image, and three channels indicate a colored image with red (R), green (G) and blue (B) components. By isolating one of these dimensions, a simple augmentation can be performed. Emphasizing color channels instead of completely removing them is a suitable choice when important features would otherwise disappear. Color channels can also be easily influenced using matrix multiplication, which results in adjusted brightness. Another common method is to derive a color histogram and use its distribution to adjust the brightness. Histogram equalization adjusts pixel values based on an image's histogram to achieve a roughly uniform distribution of intensity values. The result is an image with enhanced contrast~\cite{shorten19}.\\
\\
\textbf{Rotation.}
Rotation involves rotating an image left or right around an anchor point, typically the center of the image. In some scenarios, a small rotation performs best because the resulting image still represents a realistic scenario~\cite{shorten19}.\\
\\
\textbf{Kernel Filters.}
Kernel filters are commonly used to sharpen or blur an image by sliding an $n \times n$ kernel over it. Gaussian filters are a popular method for blurring images, while vertical and horizontal high-contrast filters are often used to sharpen image edges. Sharpening provides important structural details that the model benefits from during training. Conversely, blurring improves the model's robustness, especially against challenges such as motion blur~\cite{shorten19}.\\
\\
\textbf{Noise Injection.}
Noise injections benefit a model's robustness by allowing it to learn more salient features. The model should be able to recognize the content of an image after a small amount of noise was added~\cite{goodfellow16}. An easy way to inject noise is to take random numbers from a Gaussian distribution and add them to an image~\cite{shorten19}.\\
\\ 
\textbf{Perspective Transform.}
Perspective transformations simulate images captured from different viewpoints. It is a specific type of planar homography that models the relationship between two images of the same object, taken from varying camera angles and their respective projections. Each pixel in the input image is transformed using a projection matrix. The resulting output image retains the original dimensions. Pixels that fall outside the transformed boundaries are filled with black~\cite{wang20b}. \\
\\
\textbf{Elastic Deformation.}
Simard et al.~\cite{simard03} introduced elastic deformation as a data augmentation technique for datasets containing handwriting, aiming to simulate the natural tremor of hand muscles during writing. Since its introduction in 2003, this approach has been commonly used in various areas. The process begins by randomly generating a displacement field, which is then smoothed using a Gaussian filter and scaled to control the intensity of the deformation. \\
\\
\textbf{Randomly Cropping.}
Random cropping is used to adjust the spatial dimensions of an image. If the original dimensions must be maintained, the cropped image can be resized again to its original shape. This method often produces similar results to those from translations. However, it may also lead to the partial removal of the object of interest, which encourages the model to learn a more robust and diverse set of features~\cite{shorten19}.\\
\\
\textbf{Shadow and Highlight Injection.}
Mazhar and Kober~\cite{mazhar21} demonstrated that adding shadows and highlights to images is an effective data augmentation technique. It not only makes models more robust to varying lighting conditions but also helps prevent overfitting. They randomly add shadows and highlights to darken or brighten image areas. These effects can follow different patterns, such as linear or radial, to imitate real-world lighting scenarios.\\
\\
\textbf{Sample Mixing.}
Mixup~\cite{zhang18} is a widely used data augmentation technique, where two randomly selected images are blended to create new training samples. This can lead to better generalization and increased robustness. They stated that linearly interpolating two inputs should ideally result in the linear interpolation of their corresponding targets. CutMix~\cite{yun19} combines the ideas of Mixup and Cutout~\cite{devries17} and is better suited for discrete labels, such as segmentation masks. For this approach, a region from one image is cut out and replaced with a patch of the same size from another image. This forces the model to focus on different object features, similar to random cropping. Compared to Cutout, where removed pixels are colored in black, CutMix makes full use of the image pixels.

\begin{figure}[H]
    \centering   %
\setkeys{Gin}{width=\linewidth}
\begin{subfigure}{0.117\textwidth}
    \centering
    \includegraphics[width=1.0\linewidth]{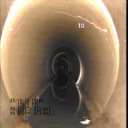}
    \caption{}
\end{subfigure}
\begin{subfigure}{0.117\textwidth}
    \centering
    \includegraphics[width=1.0\linewidth]{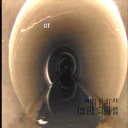}
    \caption{}
\end{subfigure}
\begin{subfigure}{0.117\textwidth}
    \centering
    \includegraphics[width=1.0\linewidth]{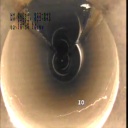}
    \caption{}
\end{subfigure}
\begin{subfigure}{0.117\textwidth}
    \centering
    \includegraphics[width=1.0\linewidth]{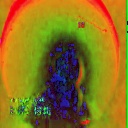}
    \caption{}
\end{subfigure}
\begin{subfigure}{0.117\textwidth}
    \centering
    \includegraphics[width=1.0\linewidth]{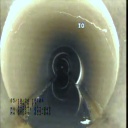}
    \caption{}
\end{subfigure}
\begin{subfigure}{0.117\textwidth}
    \centering
    \includegraphics[width=1.0\linewidth]{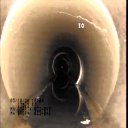}
    \caption{}
\end{subfigure}
\begin{subfigure}{0.117\textwidth}
    \centering
    \includegraphics[width=1.0\linewidth]{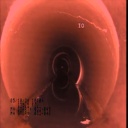}
    \caption{}
\end{subfigure}
\begin{subfigure}{0.117\textwidth}
    \centering
    \includegraphics[width=1.0\linewidth]{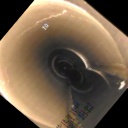}
    \caption{}
\end{subfigure}
\\
\begin{subfigure}{0.117\textwidth}
    \centering
    \includegraphics[width=1.0\linewidth]{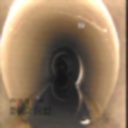}
    \caption{}
\end{subfigure}
\begin{subfigure}{0.117\textwidth}
    \centering
    \includegraphics[width=1.0\linewidth]{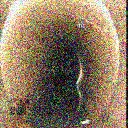}
    \caption{}
\end{subfigure}
\begin{subfigure}{0.117\textwidth}
    \centering
    \includegraphics[width=1.0\linewidth]{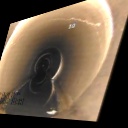}
    \caption{}
\end{subfigure}
\begin{subfigure}{0.117\textwidth}
    \centering
    \includegraphics[width=1.0\linewidth]{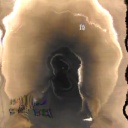}
    \caption{}
\end{subfigure}
\begin{subfigure}{0.117\textwidth}
    \centering
    \includegraphics[width=1.0\linewidth]{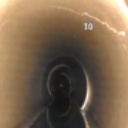}
    \caption{}
\end{subfigure}
\begin{subfigure}{0.117\textwidth}
    \centering
    \includegraphics[width=1.0\linewidth]{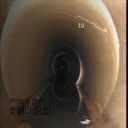}
    \caption{}
\end{subfigure}
\begin{subfigure}{0.117\textwidth}
    \centering
    \includegraphics[width=1.0\linewidth]{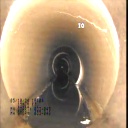}
    \caption{}
\end{subfigure}
\begin{subfigure}{0.117\textwidth}
    \centering
    \includegraphics[width=1.0\linewidth]{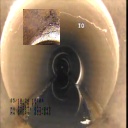}
    \caption{}
\end{subfigure}
\\
\caption{Visual comparison of different data augmentation techniques. (a) original image. (b) horizontal flip. (c) vertical flip. (d) color jittering. (e) linear color transformation. (f) histogram equalization. (g) emphasized color channel 'red'. (h) rotation. (i) kernel filters such as Gaussian blur. (j) random noise. (k) perspective transform. (l) elastic deformation. (m) random cropping (n) linear shadow injection. (o) radial highlight injection. and (p) CutMix. }
\label{fig:comp_augmentation_12}
\end{figure}

\section{\germanenglish{Class Imbalance}{Class Imbalance}}
Class imbalance is a common problem in machine learning and can significantly skew a model's performance. It refers to a situation where the classes in a dataset are distributed disproportionately, causing the model to be biased toward predicting the majority classes~\cite{shorten19}. Predicting the minority class can be compared to finding a needle in a haystack. The model struggles to distinguish between the majority and minority classes, and although it may mostly predict the majority class, the overall accuracy can still appear high due to the minimal impact of the minority class on the evaluation metrics. Class imbalance can either be intrinsic or extrinsic. Intrinsic imbalance reflects real-world distribution of samples and is most of the time unavoidable. For instance, in medical image processing, most pixels represent healthy tissue, while only a few belong to diseased ones. On the other hand, extrinsic imbalance occurs because of external factors, such as time constraints or sampling strategies. A dataset containing mostly urban scenes and only a few rural ones is a good example of extrinsic imbalance, as it results from how and where the data was collected. Class imbalance occurs in big data as well as traditional data. However, it has a larger influence on big data and can harm the predictions significantly, because a bigger skew of distribution is possible. But, for both dataset sizes, countermeasures may be necessary. Oversampling and undersampling are common strategies used to mitigate the issue of class imbalance. Oversampling increases the number of samples per minority class by replicating already existing ones. Undersampling reduces the number of samples per majority class to achieve a more balanced distribution. Randomly adding or removing samples works surprisingly well and better than some algorithmic approaches. The downsides of oversampling are that it increases the size of the training set, resulting in a longer training time and higher computational costs. Additionally, there is a higher risk of overfitting to the minority classes because the model learns to expect very similar samples, and may not generalize well to unseen ones. On the other hand, undersampling may remove valuable information, which harms the performance of the model and prevents the model from learning necessary patterns. Empirical results, however, suggest that random oversampling performs overall better than random undersampling~\cite{leevy18}. Data augmentation can help with class imbalance by oversampling minority classes. Oversampling creates new samples for underrepresented classes using augmentation techniques to make the class distribution more balanced. However, oversampling can cause the model to overfit, especially if the new samples are very similar to each other or to the originals. For this reason, it should be used carefully~\cite{shorten19}.

\section{\germanenglish{Dynamic Label Injection}{Dynamic Label Injection}}\label{dli_theory}
\begin{figure}[htp]
\centering
\captionsetup{justification=centering,margin=0cm}
\includegraphics[width=0.8\textwidth]{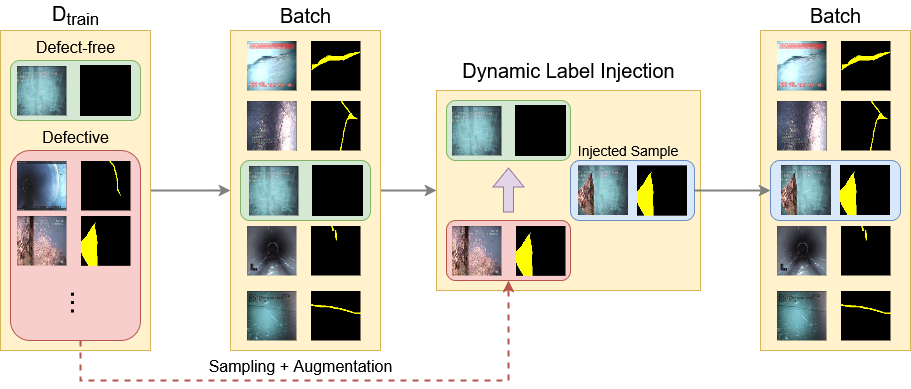}
\captionsetup{width=1.0\textwidth}
\caption{Pipeline of dynamic label injection. If a batch contains a defect-free sample, a defect from another image in the training set is injected to generate a new defective sample. The original defect-free image and its mask are then replaced with the newly created image-mask pair~\cite{caruso24}.}
\label{fig:dli_pipeline}
\end{figure}

Caruso et al.~\cite{caruso24} introduced dynamic label injection in 2024 and demonstrated the advantages of handling class imbalance dynamically during training. They created more uniformly distributed batches by extracting defects from images using the corresponding segmentation mask and adding them to defect-free images using Poisson-based seamless cloning or simple cut-paste. Dynamic label injection is an algorithmic data augmentation process that handles class imbalance in an online learning setting. During training, the occurrences of class labels in a batch are counted, and all defect-free images in the batch are injected with defects from underrepresented classes to achieve a more uniformly distributed batch~\cite{caruso24}. Figure~\ref{fig:dli_pipeline} visualizes the algorithm of dynamic label injection.\\

\noindent
A dataset $D$ consists of $N$ images and $N$ corresponding ground truth segmentation masks, so $D = {(I_i, M_i)}_{i=1}^N$. $I_i \in \mathbb{R}^{H \times W \times 3}$ represents a color image with width $W$ and height $H$, and $M_i$ is the corresponding segmentation mask with dimensions $M_i \in \{0, ..., C\}^{H \times W}$. The segmentation mask is the same size as the image, providing a pixel-level annotation, and includes class labels ranging from $0$ to $C$. Label $0$ indicates a defect-free pixel, and $C$ stands for the total number of different defect classes. A dataset $D$ is heavily imbalanced if it has two subsets, $D_i$ and $D_j$, where each set contains images with class labels $i$ or $j$ exclusively, and their sizes vary significantly as~\cite{caruso24}:
\begin{equation}\label{eq:imbalance}
    ||D_i| - |D_j|| \gg 0
\end{equation}
Dynamic label injection handles class imbalances dynamically by iterating over all defect-free images in a batch. For each one, it calculates the number of samples per defect in the current batch and then injects the least common one into the defect-free image. After adding a defect, the sample distribution changes and is recalculated in the next iteration step before handling the next defect-free image. To inject a defect, Poisson-based seamless image cloning or cut-paste is used. Cut-paste simply cuts the defect from one image using the segmentation mask and pastes it onto a different one. Poisson-based seamless cloning blends the area of the target image into the source image seamlessly~\cite{caruso24}. Its concept is described in more detail in Section~\ref{poisson}. 

\subsection{Algorithm}\label{dli_algorithm}
For each defect-free image $I_f$ in a batch $B$~\cite{caruso24}:
\begin{enumerate}[itemsep=0cm,parsep=0cm]
\item Count occurrences of each class $c \in \{1, \dots, C\}$ in $B$.
\item Identify class $c^*$ with the fewest samples in $B$.
\item Randomly select an image $I_d$ and the corresponding mask $M_d$ from $D_{train}$ such that $M_d \in \{0, c^*\}$.
\item Apply random transformations to $I_d$ and $M_d$.
\item Randomly choose between Poisson-based seamless cloning and cut-paste.
\item Inject defect from $I_d$ into $I_f$ using $M_d$, resulting in new image $I_f^*$.
\item Replace $I_f$ with $I_f^*$ in batch $B$.
\item Replace $M_f$ with $M_d$ in batch $B$.
\end{enumerate}
The above steps are repeated for all defect-free images to ensure that none are included after the injection process, and to achieve a more balanced sample distribution~\cite{caruso24}.

\subsection{Poisson-based Seamless Cloning}\label{poisson}
The idea of Poisson-based seamless cloning is to not only copy and paste a region $\Omega$ from one image $I_d$ to another image $I_f$, but also blend the region into the target image $I_f$ by matching the gradients of both images, while still maintaining the boundary conditions of $I_f$. This produces a realistic-looking image. The Poisson equation is defined as~\cite{caruso24}:
    
\begin{equation}
    \Delta g = \nabla v \quad \text{ in } \Omega,
\end{equation}
where $v$ describes the gradient of the defect image $I_d$, and $\Delta g$ represents the Laplace operator of the blended region $g$. The border condition is defined as~\cite{caruso24}:
\begin{equation}
    g = I_f \quad \text{ at } \; \partial \Omega,
\end{equation}
where $\partial \Omega$ describes the border region of $\Omega$ extracted from $I_d$. This equals the defect-free image $I_f$, indicating a seamless blend. To find a solution for $g$, the following energy function needs to be minimized~\cite{caruso24}:
\begin{equation}
    E(g) = \int_{\Omega} || \nabla g - \nabla I_f ||^2 dx
\end{equation}
This ensures a seamless integration of the extracted defect into the defect-free image by considering a gradual gradient transition of both images~\cite{caruso24}.

\section{\germanenglish{Few-Shot Semantic Segmentation}{Few-Shot Semantic Segmentation}}
The introduction of Convolutional Neural Networks (CNNs) has significantly advanced the field of computer vision and improved the accuracy of semantic segmentation models. However, their performance relies on the amount of data they have access to during training.  Large, diverse data supports the training process, while limited data will likely lead to poor performance. At the same time, collecting and annotating data is resource-intensive. In some application areas, creating a large dataset is not feasible. The annotation process often requires manual work and domain-specific knowledge. Furthermore, the training data must include high variability to represent various scenarios~\cite{catalano23, ferdaus25}.\\
\\
Few-shot learning is a machine learning task that focuses on an efficient training process when only a few annotated samples are available. Few-shot semantic segmentation is a specialized approach that focuses on semantic segmentation. It addresses the problem of designing a semantic segmentation model that is able to segment and identify classes when only a few examples are available for training. It predicts a segmentation mask $M^q$ for a query image $I^q$ given a support set $S$ with $N$ training pairs. Each pair is composed of one image $I^i$ and the corresponding segmentation mask $M^i$~\cite{catalano23}:
\begin{equation}
    S = \{(I^i, M^i)\}_{i=1}^{N}.
\end{equation}
The support set can be further subdivided into multiple subsets. Each subset $S(l)$ contains all image-mask pairs in $S$ for which the mask contains the label $l$. In a few-shot semantic segmentation setting, $k$ defines the number of pairs included in $S(l)$ and is rather small~\cite{catalano23}:
\begin{equation}
    S(l) = \{(I^i, M^i_l)\}_{i=1}^{k}.
\end{equation}
Then, the model learns a function, $f_{\theta}$, using the query image, $I^q$, and the support set, $S(l)$, to produce a segmentation mask, $M^q_l$, that should be similar to the ground truth mask, $\hat{M^q_l}$~\cite{catalano23}:
\begin{equation}
    \hat{M^q_l} = M^q_l = f_{\theta}(I^q, S(l)).
\end{equation}

\noindent
Because of the limited data available for few-shot segmentation, traditional training procedures are not suitable for this particular use case, and episodic training is commonly used instead. Episodic training splits the training process into a series of episodes. Instead of letting the model learn everything at once, it trains the model in smaller steps, also called episodes~\cite{catalano23}. A $n$-way $k$-shot segmentation task describes the segmentation process of query images, where the support set includes $n$ classes, and each class is represented by $k$ samples~\cite{liu20prototype}. For each episode, the model receives $k \cdot n$ samples in the form of a support set $S$ to learn from and creates a segmentation mask for all query images in the query set $Q$. The number of query images, as well as the definition of $k$ and $n$, are design choices and vary for different application areas and datasets. By learning from only a few examples, few-shot semantic segmentation is a practical solution that is relevant for many real-world scenarios where collecting and labeling large amounts of data is simply not feasible~\cite{catalano23}.\\
\\ 
The research performed on few-shot semantic segmentation can be divided into three different concepts: conditional networks, prototypical networks and advanced networks. Advanced networks include strategies such as latent space optimization and reflect the current state of Artificial Intelligence (AI) and synthetic data approaches~\cite{catalano23}. The following sections will discuss each of these strategies in more detail, with a particular focus on prototypical networks, as this is the approach we adopted in our research.

\subsection{\germanenglish{Conditional Networks}{Conditional Networks}}
\begin{figure}[htp]
\centering
\captionsetup{justification=centering,margin=0cm}
\includegraphics[width=0.7\textwidth]{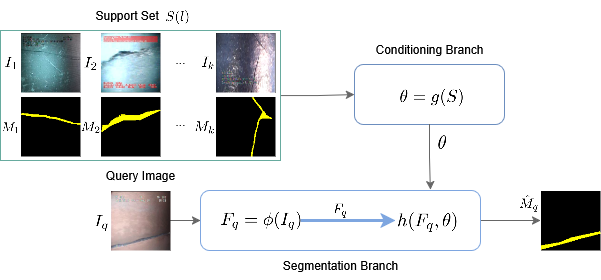}
\captionsetup{width=1.0\textwidth}
\caption{The conditional network consists of two branches: the conditional branch and segmentation branch. While the conditional branch takes a support set as an input and produces a parameter set $\theta$, the segmentation branch uses $\theta$ and a feature vector $F_q$ extracted from the query image $I_q$ to predict the corresponding segmentation mask $\hat{M_q}$~\cite{catalano23}.}
\label{fig:conditional}
\end{figure}

Shaban et al.~\cite{shaban17} addressed the few-shot semantic segmentation problem first and utilized a conditional network consisting of two branches, namely the conditional branch and the segmentation branch. The conditional branch receives a support set $S(l)$ as an input and generates a parameter set $\theta$. Then, the segmentation branch takes $\theta$ alongside a query image $I_q$ and predicts the corresponding segmentation mask $\hat{M_q}$. It extracts a feature embedding $F_q$ from the query image and uses $\theta$ to predict the segmentation mask for each pixel location in the image individually~\cite{catalano23}. Figure~\ref{fig:conditional} visualizes this process.\\
\\
The design of these two branches is critical to the performance of the model. The conditional branch and its predicted parameter set $\theta$ have great influence on the final prediction of the segmentation branch. Furthermore, the relation between the parameter set $\theta$, the query image $I_q$, and the segmentation mask $\hat{M_q}$ can be complex and hard to learn. The issues arising from using separate branches for the support and query sets are later alleviated by the adoption of prototypical networks~\cite{catalano23}.

\subsection{\germanenglish{Prototypical Networks}{Prototypical Networks}}\label{prototypical_network}
Prototypical networks were first introduced by Snell et al.~\cite{snell17} for few-shot classification. Dong et al.~\cite{dong18} adapted the idea for few-shot semantic segmentation, and it has become a well-known strategy used in many studies~\cite{wang19, zhang19, liu20part-aware, zhang20}. The idea is to represent images in an embedded space where similar images are positioned closer together than images from different scenes. By computing the centroid of all images in a given class, a prototype can be generated representing that class. Thus, the main idea of prototypical networks is to compute prototypes using the support set and then find the closest prototype for each pixel in the query image using a distance metric, labeling them individually~\cite{catalano23}.

\begin{figure}[htp]
\centering
\captionsetup{justification=centering,margin=0cm}
\includegraphics[width=0.8\textwidth]{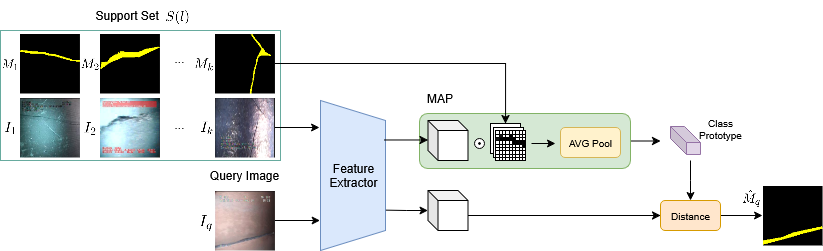}
\captionsetup{width=1.0\textwidth}
\caption{A prototypical network utilizes the same feature extractor for support and query images  to create feature embeddings. The class prototypes are generated by using the support set in combination with Masked Average Pooling (MAP), element-wise multiplication $\odot$ and Average Pooling. The segmentation mask $\hat{M_q}$ is computed by finding the closest class prototypes to the embedded features of the query image $I_q$~\cite{catalano23}.}
\label{fig:prototypical}
\end{figure}

\noindent
Figure~\ref{fig:prototypical} visualizes the internal process of a prototypical network. First, images are fed through an arbitrary feature extractor. To get a more informative feature representation, Masked Average Pooling (MAP) is applied. Then, the ground-truth mask $M_l^i$ is used to extract all relevant features belonging to a certain class. Class prototypes are constructed by averaging all extracted features of a given class, resulting in a single feature vector. The features of the query images are extracted using the same backbone as the support images. The final segmentation mask is created by finding the closest prototype for each query embedding~\cite{catalano23}. Snell et al.~\cite{snell17} proposed using squared Euclidean distance as a distance metric. However, Wang et al.~\cite{wang19} recommend using the cosine similarity distance metric because it improves performance and provides greater stability. \\
\\
In mathematically wording, the prototype $p_s^c$ of class $c$ is defined as~\cite{liu20prototype}:
\begin{equation}
\begin{split}
    p_s^c &= \frac{1}{k} \sum_{i = 1}^{k} MAP(\phi(I_{s}^{c,i}), M_{s}^{c,i}), \\
    &= \frac{1}{k} \sum_{i = 1}^{k} \frac{\sum_{x,y}F_{s;x,y}^{c,i} \; \mathbbm{1} (M_{s;x,y}^{c,i} = c)}{\sum_{x,y} \mathbbm{1} (M_{s;x,y}^{c,i} = c)},
\end{split}
\end{equation}
where the support set contains $k$ shots per class, and $F_{s;x,y}^{i,j}$ represents the extracted features of the $i$-th support image of class $c$ at position $(x,y)$. The prototype $p_s^c$ of class $c$ considers all extracted features at spatial locations, where the corresponding mask $M_{s;x,y}^{c,i}$ contains the label $c$. After computing suitable prototypes for each class, the query images are segmented by finding the closest prototype at each spatial position $(x, y)$~\cite{liu20prototype}:
\begin{equation}
    \hat{M}_{q;x,y}^{i} = \text{arg } \underset{n}{\text{max }} \sigma(d(F_{q;x,y}^{i}, p_s^c)),
\end{equation}
where $F_{q;x,y}^{i}$ represents the extracted features from the query image at pixel location $(x,y)$. $d(\cdot)$ denotes the distance metric used to compute the most similar class prototype, and $\sigma(\cdot)$ indicates the softmax function.\\
\\
To improve the training process further, Wang et al.~\cite{wang19} introduced the novel idea of creating bidirectional prototypes. In addition to generating class prototypes for the support set, they also compute prototypes for the query set using the query segmentation masks. These query set prototypes are then used to segment the support set images. This bidirectional approach ensures that the prototypes from the support and query data are better aligned, allowing the model to extract more knowledge from the support set. With the traditional technique, knowledge flows only from the support set to the query set. However, when information flows back from the query set to the support set, the model aligns the two sets of prototypes, thereby improving its generalization performance~\cite{wang19}. 

\subsection{\germanenglish{Advanced Networks}{Advanced Networks}}
The introduction of Generative Adversarial Networks (GANs)~\cite{goodfellow14} in 2014 created new opportunities for few-shot semantic segmentation. Various research has been performed and GANs have been utilized in different application areas. They are commonly used in areas where a lot of unlabeled data is available, as is often the case in partial segmentation. GANs represent features in a latent space, capturing important geometric and semantic details of an image~\cite{catalano23}. Karras et al.~\cite{karras19} demonstrated that combining multiple latent representations using interpolation results in a high-quality image. This led to the assumption, that images generated by GANs provide more relevant information than features extracted using other techniques~\cite{catalano23}. 

\begin{figure}[htp]
\centering
\captionsetup{justification=centering,margin=0cm}
\includegraphics[width=0.8\textwidth]{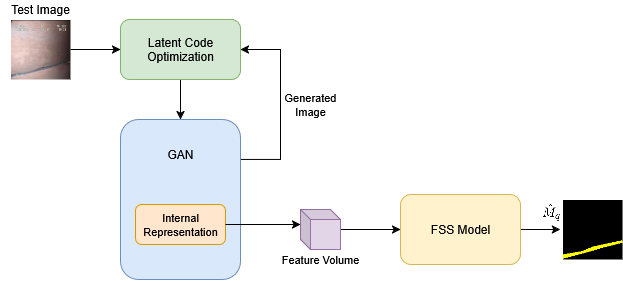}
\captionsetup{width=1.0\textwidth}
\caption{The shown GAN-based network uses latent code optimization during inference to iteratively improve the GAN-generated image to closely resemble the test image. The internal features of the generated image are then fed into a segmentation model to create the corresponding mask to the test image~\cite{catalano23}.}
\label{fig:gan_based}
\end{figure}

\noindent
Tritrong et al.~\cite{tritrong21} demonstrated the advantages of GANs in few-shot segmentation tasks. As shown in Figure~\ref{fig:gan_based}, they proposed using GANs during inference to generate an image similar to the test image and use its internal representation to create a segmentation mask corresponding to the original image. The synthetic image is iteratively refined using latent space optimization to generate an image similar to the test image. However, latent space optimization is computationally intensive and complex, motivating researchers to find alternatives. Saha et al.~\cite{saha22} demonstrated that including contrastive learning in the feature extractor enables the backbone to generate internal representations similar to the ones created by GANs. This approach achieves similar results with lower complexity. Contrastive learning is a self-supervised approach that uses image pairs, where each pair consists of an original image from the dataset and the same image after applying transformations. The model is trained to identify similarities between the two images by extracting suitable features. The model's feature extractor is then used to generate latent features of images for the segmentation task~\cite{catalano23}.\\
\\
Traditional prototypical networks have one prototype per class. This approach fails to capture class invariance and incorporates noise. Wang et al.~\cite{wang21} introduced Variational Auto Encoders (VAEs) to the few-shot semantic segmentation domain to mitigate this problem. With only one class prototype, class invariance cannot be satisfactorily captured, nor can noise be ignored. Wang et al. transformed the traditional approach into a probabilistic framework where the model learns class prototypes as probability distributions~\cite{catalano23}.


\chapter{\germanenglish{Datensatz}{Culvert Sewer Defect Dataset}} \label{culvert_sewer_defect_dataset}

This section describes the Culvert Sewer Defect Dataset (CSDD), proposed by Alshawi et al. in 2024~\cite{alshawi24b} and used to evaluate our new preprocessing pipeline. The dataset contains temporal information, as it was created from video sequences, and includes nine defect classes. It was developed by the University of New Orleans in cooperation with the US Army Corps of Engineers, and is specifically designed for autonomous defect detection in culverts and sewer pipes~\cite{alshawi24b}. Section~\ref{data_collection} describes the data collection process. Section~\ref{data_distribution} provides information about the distribution of samples per defect class and the corresponding class importance weights used during evaluation. Section~\ref{dataset_partitioning} outlines how we partitioned the data for training purposes. Following that, Section~\ref{data_implementation} outlines our implementation and evaluation details. Finally, Section~\ref{data_experiments} describes our preprocessing pipeline, which embraces standard data augmentation techniques and dynamic label injection. In addition, we report the obtained results and compare state-of-the-art methods that incorporate our proposed preprocessing steps. 

\section{\germanenglish{Datensatzerstellung}{Data Collection}}\label{data_collection}
The dataset was created using 580 annotated videos of culverts and sewer pipes. Individual frames were extracted from the videos every four to ten seconds, and experienced technicians inspected these frames to annotate them according to predefined standards. The masks are created using pixel-wise annotation, marking the position of the deficiency with the specific class label. The annotations required precise manual work to ensure that the deficiencies, especially their boundaries, were carefully captured. The annotations are used as a ground truth for training and evaluation and provide important information for precise and robust defect detection. The trained model is then able to detect defects on a pixel-level, ensuring a precise defect localization~\cite{alshawi24b}.\\
\\
The videos include different culverts and sewer pipes, ensuring a variation of shape, size, material and environmental conditions in the training data. Culverts and sewer pipes have similar structural characteristics, but they are designed to serve different purposes. Culverts are usually placed under roads or railways to let rainwater or stream flow pass through without causing flooding or damage. They help keep natural water paths intact and prevent water from building up on the surface. Sewer pipes, on the other hand, are an essential part of a much larger underground infrastructure system. They are responsible for transporting wastewater from their source, e.g. homes, industries and storm drains, to designated disposal facilities. Although both involve moving water through pipes, their structural requirements and contexts of use are quite different~\cite{butler18}. Including a high diversity of data ensures that the model is able to capture the variance of real-world scenarios and performs well in real-world inspection~\cite{alshawi24b}.

\section{\germanenglish{Datensatzverteilung}{Data Distribution}}\label{data_distribution}
The dataset contains nine class labels, one for the background and eight defect classes. There are a total of 12,230 images in the dataset. The class labels and their respective distribution are listed in Table~\ref{tab:total_class_distr}.
\begin{table}[ht]
\centering
\begin{tabular}{c l c c}
\hline \hline
\textbf{Label} & \textbf{Deficiency} & \textbf{Nr. of Samples} & \textbf{Proportion (\%)} \\
\hline
1 & Cracks & 2,981 & 24.13 \\
2 & Holes & 1,171 & 9.48 \\
3 & Roots & 406 & 3.29 \\
4 & Deformation & 385 & 3.12 \\
5 & Fracture & 3,235 & 26.19 \\
6 & Erosion & 78 & 0.63 \\
7 & Joint Problems & 3,991 & 32.31 \\
8 & Loose Gasket & 105 & 0.85 \\
\hline \hline
\end{tabular}
\caption{Class distribution in the dataset.}
\label{tab:total_class_distr}
\end{table}

\noindent
It is noticeable that there is a significant class imbalance in the dataset, where the deficiencies \textit{Roots}, \textit{Deformation}, \textit{Erosion}, and \textit{Loose Gasket} are underrepresented, with sample counts between 78 and 406. In contrast, the deficiencies \textit{Cracks}, \textit{Fracture}, and \textit{Joint Problems} are overrepresented, with sample counts ranging from 2,981 to 3,991. To address this class imbalance, we implemented various approaches described in Section~\ref{data_experiments}.\\
\\
The dataset includes eight types of pipe deficiencies. Below is a short description of each class.\\
\\
\textbf{Cracks.} Cracks are break lines that are visible but not open. The two edges remain close together, with no visible gap between them. Cracks can occur longitudinally, in a circular pattern, or as a combination of multiple intersecting cracks~\cite{nasscoPACP}.\\
\\
\textbf{Fractures.} Fractures are break lines that, unlike cracks, have a visible gap between the edges. However, the fractured parts remain in place and cannot be moved. Fractures can occur longitudinally, in a circular pattern, or as a combination of multiple intersecting lines. Cracks and fractures may also intersect~\cite{nasscoPACP}.\\
\\ 
\textbf{Holes.} Holes are defined as areas where pipe material is missing. If the material is completely removed and the hole is slightly larger, the surrounding soil may become visible. During defect detection, pipes that have been repaired with patches may be mistakenly classified as holes, which must be handled carefully~\cite{nasscoPACP}.\\
\\
\textbf{Roots.} Roots indicate an unintended open path to the surrounding soil when they are able to enter a pipe. Roots often point to weak spots in a pipe, such as joints, lateral connections, or existing defects that require repair. They grow quickly and can cause severe damage to the pipe~\cite{nasscoPACP}.\\
\\
\textbf{Deformations.} Deformations describe visible changes to the original geometry or cross-section of a pipe~\cite{nasscoPACP}.\\
\\
\textbf{Erosion.} Erosion refers to the gradual loss of the surface material of a pipe. This happens naturally, for example, due to water wearing off the material. However, it can lead to significant issues and damage~\cite{nasscoPACP}.\\
\\
\textbf{Joint problems.} Joint problems are structural deficiencies and happen at the connection point of two pipes, the so-called joint. Joints are designed to prevent surrounding soil and water from entering the pipe and to prevent leakage. Joint problems occur when the two pipe segments are misaligned~\cite{nasscoPACP}.\\
\\
\textbf{Loose gaskets.} Loose gaskets refer to issues with the sealing element between two connecting pipes. Unlike joint problems, a loose gasket is a structural deficiency. A loose gasket can cause leakage, and its detection is important but challenging due to its small size and the difficulty of spotting loose elements in images or videos~\cite{nasscoPACP}.

\begin{table}[ht]
\centering
\begin{tabular}{c l c}
\hline \hline
\textbf{Label} & \textbf{Deficiency} & \textbf{Class Importance Weight (CIW)} \\
\hline
1 & Cracks & 1.0000 \\
2 & Holes & 1.0000 \\
3 & Roots & 1.0000 \\
4 & Deformation & 0.1622 \\
5 & Fracture & 0.7100 \\
6 & Erosion & 0.3518 \\
7 & Joint Problems & 0.6419 \\
8 & Loose Gasket & 0.5419 \\
\hline \hline
\end{tabular}
\caption{The class importance weights capture the impact of specific deficiencies on the economy or public safety according to US standards.}
\label{tab:ciw}
\end{table}

\noindent
For evaluation purposes, professional civil engineers defined for each class an importance weight that captures the impact of the specific deficiency on the public safety and economy. Table~\ref{tab:ciw} includes a list of the different deficiencies alongside their class label and importance weight. A higher importance weight indicates a higher impact of the deficiency and a higher relevance for repairing the defect promptly. The weights are considered in the calculation of the frequency-weighted Intersection over Union (FWIoU). The FWIoU reflects the performance of the model by evaluating the overlapping areas of the ground-truth and the predicted deficiency mask while considering the different types of defects and their importance to society. The class importance weights are normalized by dividing each weight by the highest score~\cite{alshawi24b}.  The calculation of the FWIoU, along with other evaluation metrics, is described in Section~\ref{metrics}.

\section{\germanenglish{Datensatzteilung}{Dataset Partitioning}}\label{dataset_partitioning}
We reduced the number of training samples to ensure there were no identical or highly similar images in the training and test sets. As described in Section~\ref{data_collection}, the images were collected by extracting frames from videos every four to ten seconds. However, if the robot did not move during this time, the same image could appear twice in the dataset. Even minor movements by the robot result in similar images that can cause data leakage.

\begin{figure}[H]
\centering
\captionsetup{justification=centering,margin=0cm}
\includegraphics[width=0.8\textwidth]{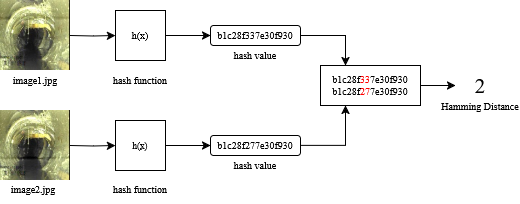}
\captionsetup{width=1.0\textwidth}
\caption{Pipeline of image hashing where two images are compared using their hash values. In this example, two characters in the hash strings differ, indicating a Hamming distance of two. A small visual difference is visible as an additional shadow line appears in \texttt{image2.jpg}.}
\label{fig:imagehashing}
\end{figure}

\noindent
To ensure a non-overlapping train and test split, we first divided the dataset into training, validation, and test sets using a 70/15/15 split. Then, we compared the training and test sets using a combination of image hashing and Hamming distance. Image hashing computes a hash value for each image that uniquely identifies it. Identical images receive the same hash value. The Hamming distance between two strings indicates the number of operations required to transform one string into the other. These operations can include changing, adding, or removing a character. Therefore, the Hamming distance between two identical strings is zero. In our case, the strings are represented by the hash values of the images, so computing the Hamming distance between two hash values evaluates the similarity of the images. Figure~\ref{fig:imagehashing} visualizes the process of comparing two images using image hashing and Hamming distance.\\
\\
We tested different thresholds for the Hamming distance in order to exclude all identical and highly similar images, while preserving as many training samples as possible to avoid unnecessary reduction. The final threshold was set to seven. As a result, the number of training samples decreased from 8,561 to 2,922, while the number of validation and test samples remained at 1,834 and 1,835, respectively. The differences in sample counts and class distributions in the training set before and after the reduction, as well as the final class distribution, are listed in Table~\ref{tab:final_class_distr}.


\begin{table}[H]
\centering
\begin{tabular}{c l c c c c}
\hline \hline
\textbf{Label} & \textbf{Deficiency} & \makecell{\textbf{Train} \\ \textbf{(Before)}}  & \makecell{\textbf{Train} \\ \textbf{(After)}} & \textbf{Validation} & \textbf{Test} \\
\hline
1 & Cracks & 2,059 & 603 & 453 & 469 \\
2 & Holes & 807 & 205 & 179 & 185 \\
3 & Roots & 285 & 99 & 66 & 55 \\
4 & Deformation & 256 & 33 & 71 & 58 \\
5 & Fracture & 2,290 & 861 & 458 & 487 \\
6 & Erosion & 52 & 11 & 15 & 11 \\
7 & Joint Problems & 2,826 & 1,075 & 593 & 572 \\
8 & Loose Gasket & 75 & 64 & 19 & 11 \\
\hline \hline
\end{tabular}
\caption{Class distribution for training, validation and test set. Distributions are shown for the training set both before and after the reduction.}
\label{tab:final_class_distr}
\end{table}

\section{Implementation and Evaluation Details}\label{data_implementation}

In this section, we present the implementation and evaluation details of our experiments on the preprocessed dataset. We begin with the training setup, then describe the E-FPN~\cite{alshawi24b} model used for evaluation, and finally define the metrics and loss functions employed.

\subsection{Training Setup}\label{data_training_setup}
We conducted training for our model and the other state-of-the-art models on an NVIDIA A100 GPU (80GB) supported by 128 CPU cores and 1,007.6 GB of RAM. We used Python 3.12.2, PyTorch 2.5.1, and CUDA 12.1. Training and testing included 25 epochs with a batch size of 16. The experiments used the culvert sewer defect dataset described in Section~\ref{culvert_sewer_defect_dataset}, containing 6,591 images with a resolution of 128 $\times$ 128. We used the E-FPN model proposed by Alshawi et al.~\cite{alshawi24b} and included the preprocessing steps explained in~\ref{data_experiments}. We tested four settings: no modifications for the baseline, using only the augmented dataset, using only dynamic label injection, and using both methods together. For training, we used a combined loss function with cross-entropy, dice, and focal loss. These losses were weighted with predefined values: 0.5 for cross-entropy, 0.3 for dice, and 0.2 for focal loss. We set the reduction parameter of cross-entropy loss to \textit{'None'}. This allows a deterministic implementation, while we compute the weighted mean manually after each loss step to align with PyTorch’s default \textit{'Mean'} behavior. We optimized using AdamW with a learning rate of 0.001 and a weight decay of 1e-5. A CosineAnnealingLR scheduler was applied with 50 maximum iterations and a minimum learning rate of 1e-6. To stabilize training, we integrated gradient clipping with a maximum norm of 1.0. With the mentioned settings and a random seed of 42, we achieved the results listed in Section~\ref{data_experiments}.

\subsection{Model}\label{data_model}
We employed the Enhanced Feature Pyramid Network (E-FPN), proposed by Alshawi et al.~\cite{alshawi24b} in 2024,
to develop our data preprocessing pipeline and find the optimal combination of data augmentation techniques and dynamic label injection. The E-FPN was designed for culvert and sewer pipe defect detection and produced strong results in that area. 

\begin{figure}[htp]
    \centering
    \captionsetup{justification=centering,margin=0cm}
    \includegraphics[width=0.9\textwidth]{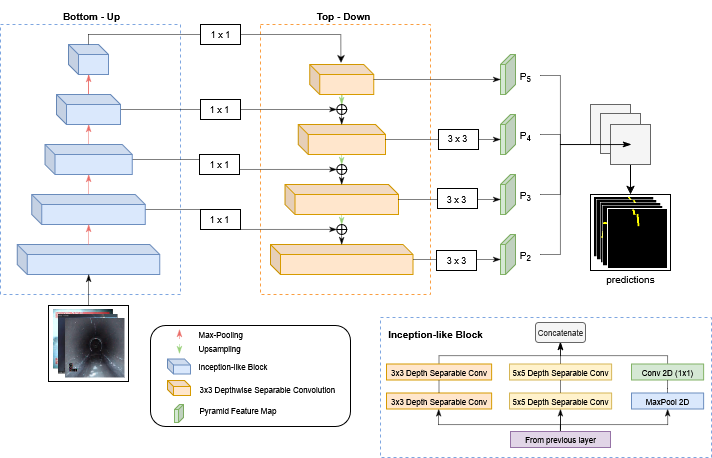}
    \captionsetup{width=1.0\textwidth}
    \caption{E-FPN~\cite{alshawi24b} architecture. The model includes a dual-pathway design. The bottom-up pathways extract features using Inception-like blocks and downsampling by a factor of 2. The top-down pathway performs feature fusion and upsampling operations to create the final segmentation mask. The Inception-like block uses three parallel paths, including two different-sized kernels and a max-pooling layer.}
    \label{fig:efpn}
\end{figure}

\noindent
The E-FPN is an extension of the feature pyramid network (FPN) introduced by Lin et al.~\cite{lin2017feature}. The structure and components of a standard FPN are explained in Section~\ref{theory_fpn}. E-FPN was built for the segmentation of culverts and sewer pipes, where defects vary in appearance and are often very small. These conditions cause a strong foreground-background imbalance and require multi-scale feature extraction. Traditional methods struggle with these challenges. E-FPN addresses them by introducing sparsely connected blocks and depthwise separable convolutions. Its architecture is shown in Figure~\ref{fig:efpn}. Like the traditional FPN, E-FPN follows a dual pathway design. The bottom-up pathway extracts multi-scale features through convolutions and downsampling. To improve this process, they replaced standard convolution layers with Inception-based blocks~\cite{alshawi24b}. Each block has three parallel paths. Two of them use depthwise separable convolution with either a $3 \times 3$ or a $5 \times 5$ kernel. The third one is a parallel max-pooling layer. The $3 \times 3$ filter is able to extract fine-grained details and textures, while the $5 \times 5$ filter captures large-scale features and offers a wider receptive field. Together, these filters provide a better understanding of the overall pipe structure. The top-down pathway performs feature fusion and upsampling to generate the final predictions~\cite{alshawi23a}. It begins with a $1 \times 1$ convolution on the output of the bottom-up pathway to reduce the channel depth to 128 while preserving important information. At each level, the output from the previous top-down layer is upsampled by a factor of two and merged with the corresponding bottom-up feature map. A $3 \times 3$ convolution is applied to the merged layers to reduce aliasing effects and retain fine details. Throughout the entire framework, depthwise separable convolutions are used instead of standard convolutions to lower the parameter count without reducing performance~\cite{alshawi24b}.

\subsection{Loss functions}\label{loss_functions}
The choice of loss function for a semantic segmentation task is important because it strongly affects the training process and, therefore, the final prediction accuracy~\cite{jadon20}. For our implementation, we used a combination of three well-known losses, namely cross-entropy, dice and focal loss. The three loss values are combined using predefined weights: $0.5$ for cross-entropy, 0.3 for dice, and 0.2 for focal loss.\\
\\
\textbf{Cross-Entropy Loss.} Cross-entropy loss is a well-established, straightforward and stable loss in semantic segmentation tasks. In multi-class settings, it is defined as~\cite{terven23}:
\begin{equation}
    L_{CE} = - \frac{1}{N} \sum_{i=1}^N \sum_{i=1}^C y_{i,c} \text{ log}(p_{i,c}),
\end{equation}
where $N$ is the total number of pixels and $C$ indicates the number of semantic classes. $p_{i,c} \in [0,1]$ denotes the predicted probability of pixel $i$ for class $c$, while $ y_{i,c} \in \{0,1\}$ is set to one if pixel $i$ actually belongs to class $c$ in the ground-truth, and zero otherwise. Cross-entropy loss works well in standard segmentation tasks. However, class imbalance affects it negatively, and complementary losses are necessary for a stable training~\cite{terven23}.\\
\\
\textbf{Dice Loss.} Dice loss is particularly established in medical image segmentation and follows an overlap-based approach. It compares the overlapped region of the ground truth, $y_i$, and the prediction mask, $\hat{y_i}$, using Dice Similarity Coefficient ($DSC$)~\cite{dice1945}. The $DSC$ is defined as~\cite{terven23}:
\begin{equation}
    DSC(\hat{y_i}, y_i) = \frac{2 |\hat{y}_i \cap y_i|}{|\hat{y}_i| + |y_i|},
\end{equation}
while the dice loss is formulated as~\cite{terven23}:
\begin{equation}
    L_{Dice} = 1 - DSC(\hat{y_i}, y_i).
\end{equation}
If the prediction perfectly matches the ground truth, $DSC$ results in one, while $L_{Dice}$ is zero. Dice loss is well-suited for class-imbalanced datasets, where foreground pixels appear much less than background ones, because the loss function penalizes poor region overlaps directly and includes higher sensitivity to rare classes. However, if the prediction and ground-truth only overlap slightly, the dice loss might be unstable and needs to be combined with other loss functions to work properly~\cite{terven23}.\\
\\
\textbf{Focal Loss.} Lin et al.~\cite{lin17} propose focal loss to handle foreground-background class imbalance effectively. For a binary segmentation, the prediction probability $p~\in~[0,1]$ states how likely the pixel belongs to the foreground class. The focal loss is defined as~\cite{terven23}:
\begin{equation}
    L_{Focal}(p_t) = -\alpha_t (1 - p_t) ^\gamma \text{ log}(p_t),
\end{equation}
\begin{equation}
        p_t = \left\{
        \begin{array}{ll}
            p & \text{if } y = 1, \\
            1 - p & \text{if } y = 0,
        \end{array} 
        \right.
\end{equation}
where $\alpha$ is a class weight, weighting the importance of foreground and background, $\gamma \geq 0$ denotes the focusing parameter, and $y$ represents the ground truth. In a multi-class setting, the binary focal loss is calculated for each class and then averaged. The focal loss focuses on hard examples, but might emphasize them too much. To mitigate this problem, the focal loss is often combined with other loss functions~\cite{terven23}.

\subsection{Metrics}  \label{metrics}
To evaluate and compare our different approaches, we employ the following metrics: Intersection over Union (IoU), Frequency-weighted IoU (FWIoU), F1 score, Balanced Accuracy (BA), and Matthew's Correlation Coefficient (MCC). The following section outlines their concepts and formulations in more detail. Prior to that, we shortly explain the concepts of true positives, true negatives, false positives and false negatives, since they are the foundation for all used metrics.\\
\\
We consider a multi-class segmentation task. For each class $c \in \{1, ..., k\}$, we compare the ground-truth mask $\hat{Y_c}$ and the predicted mask $Y_c$. The number of true positives $TP_c$ describes the number of pixels correctly predicted as class $c$~\cite{vujovic21, rainio24}:
\begin{equation}
    TP_c = |Y_c \cap \hat{Y_c}|.
\end{equation}
True negatives $TN_c$ represent the number of pixels correctly predicted as any class $l \neq c$:
\begin{equation}
    TN_c = \sum_{l = 1, l \neq c}^k |Y_l \cap \hat{Y_l}|.
\end{equation}
False positives $FP_c$ denote the number of pixels incorrectly predicted as class $c$:
\begin{equation}
    FP_c = |\hat{Y_c} \setminus Y_c|.
\end{equation}
Finally, false negatives $FN_c$ correspond to the number of pixels that should have been classified as class $c$ but were not predicted as such:
\begin{equation}
    FN_c = |Y_c \setminus \hat{Y_c}|.
\end{equation}
\\
\textbf{F1 Score.} The F1 score harmonically balances precision ($P$) and recall ($R$). $P_c$ defines the number of correctly predicted positives (TP) divided by the total number of predicted positives (TP + FP). It shows how many of the pixels labeled as class $c$ were actually correct. $R_c$, on the other hand, represents the number of correctly predicted positives (TP) divided by the the total number of actual positives (TP + FN). It reveals how many of the pixels that should have been classified as class $c$ were correctly predicted~\cite{vujovic21}. They are calculated as~\cite{davis06}:
\begin{equation}
    P_c = \frac{TP_c}{TP_c + FP_c} \in [0, 1], \\
\end{equation}
\begin{equation}
    R_c = \frac{TP_c}{TP_c + FN_c} \in [0, 1]. \\
\end{equation}
\\
The F1 score provides a single metric that aggregates precision and recall values. The F1 score is computed as~\cite{rainio24}:
\begin{equation}
    F1_c = \frac{2 \cdot P_c \cdot R_c}{P_c + R_c} \in [0, 1].
\end{equation}
\\
\textbf{Intersection over Union (IoU).} The IoU is a commonly used evaluation metric for segmentation tasks and measures the overlap between the predicted masks $Y$ and the ground truth ones $\hat{Y}$. It is defined as the ratio of the intersection and union of these two sets. Typically, the IoU is computed for each class individually and then averaged to receive a performance indicator representing all classes~\cite{garcia17}. The IoU for a single class is computed as~\cite{rainio24}:
\begin{equation}
    IoU_c = \frac{TP_c}{TP_c + FP_c + FN_c} = \frac{|Y_c \cap \hat{Y_c}|}{|Y_c \cup \hat{Y_c}|} \in [0, 1]
\end{equation}
\\
\textbf{Frequency-Weighted Intersection over Union (FWIoU).} FWIoU is a weighted adaptation of the Mean IoU and is computed over the entire dataset. It incorporates class-specific weights $w_i$, which reflect the frequency and importance of each class. In our case, we use the class importance weights listed in Table~\ref{tab:ciw}. The FWIoU is calculated as~\cite{garcia17}:
\begin{equation}
    FWIoU = \frac{\sum_{i=1}^k w_i \cdot|Y_i \cap \hat{Y_i}|}{\sum_{i=1}^k w_i \cdot|Y_i \cup \hat{Y_i}|}  \in [0, 1].
\end{equation}
\\
\textbf{Balanced Accuracy (BA).} Similar to the F1 score, balanced accuracy utilizes per-class recall values to obtain a reliable measure suitable for imbalanced datasets. In a segmentation task, recall represents the proportion of pixels of a certain class that are correctly classified. Balanced accuracy captures the mean of these individual recall values and is calculated as~\cite{grandini20}:
\begin{equation}
    BA = \frac{1}{k} \sum_{i=1}^k R_i \in [0, 1].
\end{equation}
\\
\textbf{Matthew's Correlation Coefficient (MCC).} The MCC metric is used to evaluate the correlation between predicted and ground-truth labels. It values range from negative one to positive one, where positive one indicates perfect correlation and a strong model performance. A value of zero implies that there is no correlation, while a value of negative one suggests an inverse correlation. This suggests that the model systematically swaps predicted values and indicates an implementation issue. The Mean MCC is computed by averaging the MCC scores for each individual class, while the MCC for class $c$ is calculated as~\cite{rainio24}:
\begin{equation}
    MCC_c = \frac{TN_c \cdot TP_c - FN_c \cdot FP_c}{\sqrt{(TP_c + FP_c)(TP_c + FN_c)(TN_c + FP_c)(TN_c + FN_c)}} \in [-1, 1].
\end{equation}

\section{Experiments}\label{data_experiments}
To address the challenges of class imbalance and limited sample size, we explored classic data augmentation techniques, downsampling strategies, and a new approach called dynamic label injection~\cite{caruso24}. These methods are described in detail below. For each approach, we provide a performance comparison between the original and the adapted data. Then, we determined and tested the best combination of methods on various state-of-the-art models to assess the applicability of our proposed method to different architectures. All experiments were conducted using the implementation details described in Section~\ref{data_implementation}.

\subsection{\germanenglish{Datenerweiterung}{Data Augmentation}}\label{data_augmentation}

To increase the number of samples per class and improve model generalization, we applied several data augmentation techniques to the training set. All selected methods were applied to each image in the dataset. After evaluating the techniques individually and in combination, the best results were achieved with horizontal flipping, rotation by 30° and 50°, perspective transformation, elastic deformation, and histogram equalization. We added the augmented images to the original dataset to form the final training set. Table~\ref{tab:img_data_augmentation} visualizes the final combination of data augmentation techniques.
\clearpage

\begin{table}[htp]
  \centering
  \begin{tabular}{|>{\centering\arraybackslash}m{1.6cm}|>{\centering\arraybackslash}m{1.6cm}|>{\centering\arraybackslash}m{1.6cm}|>{\centering\arraybackslash}m{1.6cm}|>{\centering\arraybackslash}m{1.6cm}|>{\centering\arraybackslash}m{1.6cm}|>{\centering\arraybackslash}m{1.6cm}|}
    \hline
    {\small Original} & {\small Horizontal Flip} & {\small Rotation by 30°} & {\small Rotation by 50°} & {\small Histogram Equalization} & {\small Perspective Transform} & {\small Elastic Deformation} \\
    \hline
    \includegraphics[width=1.0\linewidth]{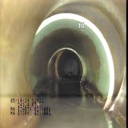} & \includegraphics[width=1.0\linewidth]{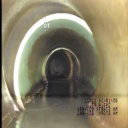} & \includegraphics[width=1.0\linewidth]{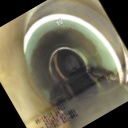} & \includegraphics[width=1.0\linewidth]{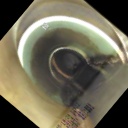} & \includegraphics[width=1.0\linewidth]{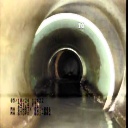} & \includegraphics[width=1.0\linewidth]{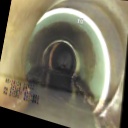} & \includegraphics[width=1.0\linewidth]{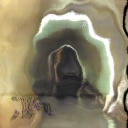} \\ 
    \includegraphics[width=1.0\linewidth]{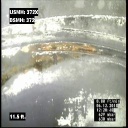} & \includegraphics[width=1.0\linewidth]{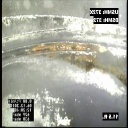} & \includegraphics[width=1.0\linewidth]{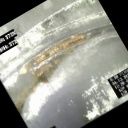} & \includegraphics[width=1.0\linewidth]{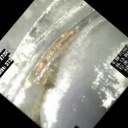} & \includegraphics[width=1.0\linewidth]{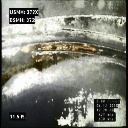} & \includegraphics[width=1.0\linewidth]{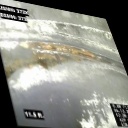} & \includegraphics[width=1.0\linewidth]{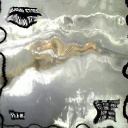} \\ \hline
  \end{tabular}
  \caption{The combination of the original image and the applied data augmentation techniques horizontal flip, rotation by 30 and 50 degrees, histogram equalization, perspective transform and elastic deformation perform the best.}
  \label{tab:img_data_augmentation}
\end{table}

\noindent
This combination of augmentation strategies significantly improved performance. On average, both the IoU and F1 score increased by over four percentage points across all classes. The most notable improvements were observed in the \textit{Cracks}, \textit{Holes}, \textit{Fracture}, and \textit{Loose Gasket} classes, each showing an increase between eight and ten percentage points. The detailed results, including the F1 score and IoU for both the original and augmented datasets, are presented in Table~\ref{tab:data_aug}.

\begin{table}[H]
\centering
\begin{tabular}{c l|c c|c c}
\hline \hline
& & \multicolumn{2}{c|}{\textbf{Original Dataset}} & \multicolumn{2}{c}{\textbf{Augmented Dataset}} \\
\textbf{Label} & \textbf{Deficiency} & \textbf{F1 Score (\%)} & \textbf{IoU (\%)} & \textbf{F1 Score (\%)} & \textbf{IoU (\%)} \\
\hline
0 & Background & 97.21 & 94.57 & 97.51 & 95.14 \\
1 & Cracks & 42.51 & 26.99 & 51.89 & 35.04 \\
2 & Holes & 90.66 & 82.91 & 95.32 & 91.06 \\
3 & Roots & 82.33 & 69.96 & 84.88 & 73.73 \\
4 & Deformation & 77.08 & 62.70 & 78.18 & 64.18 \\
5 & Fracture & 57.57 & 40.42 & 65.04 & 48.19\\
6 & Erosion & 73.24 & 57.78 & 73.87 & 58.57 \\
7 & Joint Problems & 77.91 & 63.82 & 78.61 & 64.75 \\
8 & Loose Gasket & 76.02 & 61.32 & 83.11 & 71.09 \\
\hline
\multicolumn{2}{r|}{Avg. F1 (w/ bg)} & \multicolumn{2}{c|}{74.95} & \multicolumn{2}{c}{78.71} \\
\multicolumn{2}{r|}{Avg. F1 (w/o bg)} & \multicolumn{2}{c|}{72.16} & \multicolumn{2}{c}{76.36} \\
\multicolumn{2}{r|}{Avg. IoU (w/ bg)} & \multicolumn{2}{c|}{62.27} & \multicolumn{2}{c}{66.86} \\
\multicolumn{2}{r|}{Avg. IoU (w/o bg)} & \multicolumn{2}{c|}{58.24} & \multicolumn{2}{c}{63.33} \\
\multicolumn{2}{r|}{FWIoU} & \multicolumn{2}{c|}{63.92} & \multicolumn{2}{c}{67.71} \\
\multicolumn{2}{r|}{Balanced Accuracy} & \multicolumn{2}{c|}{70.49} & \multicolumn{2}{c}{75.08} \\
\multicolumn{2}{r|}{Mean MCC} & \multicolumn{2}{c|}{73.38} & \multicolumn{2}{c}{76.77} \\
\hline \hline
\end{tabular}
\caption{Results with original and augmented dataset using E-FPN. Augmenting the data improves the results for all classes, increasing the average F1 score and IoU by over four percent.}
\label{tab:data_aug}
\end{table}


\subsection{\germanenglish{Dynamic Label Injection}{Dynamic Label Injection}}
To include more variety in our dataset, we incorporated the Dynamic Label Injection (DLI) algorithm, introduced by Caruso et al.~\cite{caruso24} in 2024. This method injects defects, extracted from images from the training set, into defect-free images, after applying a variety of transformations. Chapter~\ref{dli_theory} explains the underlying theory and concept of dynamic label injection. Our dataset does not include any defect-free images. In order to integrate dynamic label injection into our training pipeline, we expanded the training set by adding new defect-free samples. We did this by randomly cropping images and their corresponding masks from the training set into the shapes $(100 \times 100)$, $(90 \times 90)$ and $(80 \times 80)$. If the resulting mask does not include a defect class label and consists only of zeros, we add the new defect-free image-mask pair to the set. We added a total of 155 defect-free images to the training set, increasing its size from 2,951 to 3,106. The randomly cropped images provide similar variations in backgrounds and conditions as the dataset, yielding the most promising results. Each epoch introduces additional randomness by adjusting the defect-free image differently with respect to class, defect sample, and transformations. This creates more data variety, which prevents the model from overfitting. 

\begin{table}[H]
\centering
\begin{tabular}{l|c c c c c c c}
\hline \hline
& \multicolumn{7}{c}{\textbf{Probability of Poisson Image Editing}} \\
& 0.0 & 0.1 & 0.3 & \textbf{0.5} & 0.7 & 0.9 & 1.0 \\
\hline
Avg. F1 (w/ bg) & 75.88 & 76.14 & 74.52 & \textbf{76.35} & 75.99 & 74.30 & 74.06 \\
Avg. F1 (w/o bg) & 73.21 & 73.50 & 71.68 & \textbf{73.72} & 73.32 & 71.43 & 71.16 \\
Avg. IoU (w/ bg)  & 63.55 & 63.83 & 61.96 & \textbf{64.01} & 63.55 & 61.69 & 61.08 \\
Avg. IoU (w/o bg)  & 59.64 & 59.96 & 57.87 & \textbf{60.17} & 59.65 & 57.56 & 56.89 \\
FWIoU  & \textbf{65.11} & 63.52 & 63.40 & 64.14 & 63.91 & 64.19 & 62.76 \\
Balanced Accuracy  & \textbf{71.17} & 70.59 & 67.63 & 71.09 & 70.14 & 69.47 & 67.76 \\
Mean MCC  & 73.82 & 74.23 & 72.52 & \textbf{74.54} & 74.11 & 72.28 & 72.32 \\
\hline \hline
\end{tabular}
\caption{Comparison of E-FPN results using DLI with different probabilities for Poisson image blending and cut-paste. A probability of $1.0$ indicates that Poisson image blending was always used, whereas $0.0$ means that only the cut-paste method was applied. The best results are achieved with equal probability ($0.5$).}
\label{tab:diff_PROB_POISSON}
\end{table}

\noindent
During training, each batch is first adjusted using dynamic label injection before being fed into the model. The class to be injected is determined by counting the number of occurrences of each class in the batch and selecting one from those with the lowest count. Random transformations are applied to the defect image. These include horizontal flipping, translation, rotation, and scaling, where the exact combination and parameters are customized for each class label.  Both methods, Poisson-based seamless image cloning and cut-paste, were used to inject the defect with a 50 percent probability each. The results for different probability settings are listed in Table~\ref{tab:diff_PROB_POISSON}. The table clearly shows that the best results are achieved when both methods are used with equal probability. It also indicates that using only the simple cut-paste method is more effective than applying Poisson image blending exclusively.

\begin{figure}[htp]
\centering
\captionsetup{justification=centering,margin=0cm}
\includegraphics[width=1.0\textwidth]{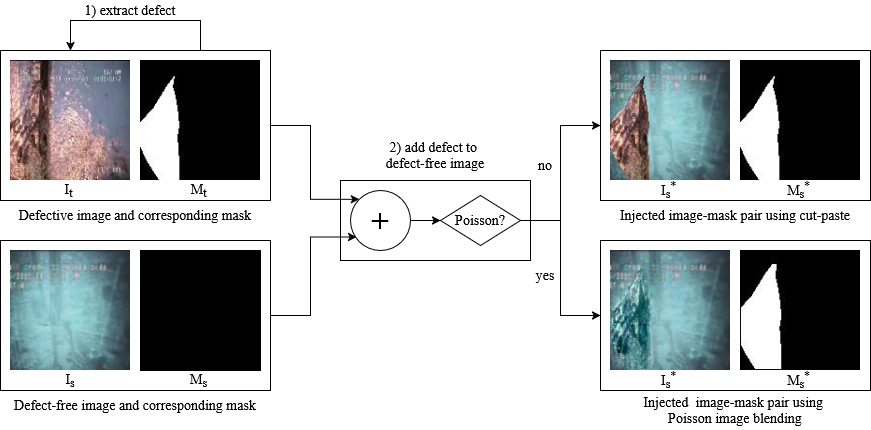}
\captionsetup{width=1.0\textwidth}
\caption{Visual comparison of the two blending methods: cut-paste and Poisson image blending. Poisson image blending matches the gradients of the source and target images, while cut-paste simply copies the extracted region to the new image.}
\label{fig:dli_diff_blending}
\end{figure}

\noindent
Figure~\ref{fig:dli_diff_blending} shows a visual comparison of the two methods and their respective outputs. Poisson-based blending is noticeable for its ability to adapt the defect to the surrounding area of the defect-free image, thereby reducing sharp edges and pixel artifacts at the boundaries. This approach modifies the image and adjusts the mask using max pooling to smooth and expand the defect region. Conversely, cut-paste does not adjust the appearance of the defect, which can result in visible boundaries. It uses the original mask from the defective image without modification. Both methods are valid and commonly used options in various application areas.

\noindent
Table~\ref{tab:nr_injections} compares the number of injections for each class on defect-free images throughout the entire 25-epoch training process. Notably, the majority classes \textit{Cracks}, \textit{Fracture}, and \textit{Joint Problems} are injected less frequently than the minority ones. However, there is a favor for lower-number classes, as visible in the provided table. That happens because of the use of \textit{argmin()} to select the class with the fewest occurrences. Using random sampling of the minority classes did indeed slightly lower the performance. Therefore, we decided to keep the original implementation. This suggests that dynamic label injection enhances the performance of the model of lower-numbered classes and that higher-numbered classes do not require as much attention.
\begin{table}[H]
\centering
\begin{tabular}{c|c c c c c c c c c}
\hline \hline
Label & 1 & 2 & 3 & 4 & 5 & 6 & 7 & 8 \\
Class & \tiny Cracks & \tiny Holes & \tiny Roots & \tiny Deformation & \tiny Fracture & \tiny Erosion & \tiny Joint Problems & \tiny Loose Gasket \\
\hline
\# Injections & 93 & 863 & 1,176 & 935 & 7 & 361 & 1 & 64 \\
\hline \hline
\end{tabular}
\caption{Number of injections during 25 epoch training for all eight defect classes.}
\label{tab:nr_injections}
\end{table}

\noindent
Adding dynamic label injection led to an overall performance improvement. On average, both the IoU and F1 score increased by nearly two percentage points across all classes. Several classes benefited from the introduction of DLI, with the most notable improvements achieved for the classes \textit{Cracks}, \textit{Holes}, and \textit{Loose Gasket}, each showing an increase between four and ten percentage points. Table~\ref{tab:dli_results} presents the detailed results, including the F1 score and IoU for both the baseline and the adjusted model incorporating DLI.

\begin{table}[H]
\centering
\begin{tabular}{c l|c c|c c}
\hline \hline
& & \multicolumn{2}{c|}{\textbf{E-FPN w/o DLI}} & \multicolumn{2}{c}{\textbf{E-FPN w/ DLI}} \\
\textbf{Label} & \textbf{Deficiency} & \textbf{F1 Score (\%)} & \textbf{IoU (\%)} & \textbf{F1 Score (\%)} & \textbf{IoU (\%)} \\
\hline
0 & Background & 97.21 & 94.57 & 97.32 & 94.79 \\
1 & Cracks & 42.51 & 26.99 & 48.13 & 31.69 \\
2 & Holes & 90.66 & 82.91 & 94.01 & 88.70 \\
3 & Roots & 82.33 & 69.96 & 82.22 & 69.81 \\
4 & Deformation & 77.08 & 62.70 & 74.44 & 59.28 \\
5 & Fracture & 57.57 & 40.42 & 57.68 & 40.52 \\
6 & Erosion & 73.24 & 57.78 & 74.86 & 59.82 \\
7 & Joint Problems & 77.91 & 63.82 & 75.56 & 60.71 \\
8 & Loose Gasket & 76.02 & 61.32 & 82.90 & 70.80 \\
\hline
\multicolumn{2}{r|}{Avg. F1 (w/ bg)} & \multicolumn{2}{c|}{74.95} & \multicolumn{2}{c}{76.35} \\
\multicolumn{2}{r|}{Avg. F1 (w/o bg)} & \multicolumn{2}{c|}{72.16} & \multicolumn{2}{c}{73.72} \\
\multicolumn{2}{r|}{Avg. IoU (w/ bg)} & \multicolumn{2}{c|}{62.27} & \multicolumn{2}{c}{64.01} \\
\multicolumn{2}{r|}{Avg. IoU (w/o bg)} & \multicolumn{2}{c|}{58.24} & \multicolumn{2}{c}{60.17} \\
\multicolumn{2}{r|}{FWIoU} & \multicolumn{2}{c|}{63.92} & \multicolumn{2}{c}{64.14} \\
\multicolumn{2}{r|}{Balanced Accuracy} & \multicolumn{2}{c|}{70.49} & \multicolumn{2}{c}{71.08} \\
\multicolumn{2}{r|}{Mean MCC} & \multicolumn{2}{c|}{73.38} & \multicolumn{2}{c}{74.53} \\
\hline \hline
\end{tabular}
\caption{Results for the original E-FPN model and the E-FPN model with integrated dynamic label injection (DLI). DLI improves the results of several classes, increasing the average F1 score and IoU by 1.5~\%.}
\label{tab:dli_results}
\end{table}

\subsection{\germanenglish{Downsampling}{Downsampling}}

\noindent
Downsampling removes instances of the majority classes to handle class imbalance~\cite{leevy18}. In our dataset, the classes \textit{Cracks}, \textit{Fracture}, and \textit{Joint Problems} are considered majority classes, each containing between 603 and 1,075 samples. In contrast, all remaining classes have significantly fewer samples, ranging from 11 to 205. We experimented with different strategies to reduce the number of images in the majority classes, either by removing a fixed percentage of samples or by setting a maximum number of samples per class. Samples were either randomly selected or identified by searching for similar images using image hashing. The latter approach prevents the loss of important information or unique patterns. However, neither approach led to significantly improved results. Additionally, we attempted to combine downsampling with data augmentation and DLI. Nevertheless, the results were worse or nearly identical to those achieved without downsampling. Table~\ref{tab:diff_downsampling} compares different downsampling approaches performed on the original dataset. Only the approach in which each majority class was reduced to 500 images via random sampling showed slightly better results than the approach without downsampling. This approach is compared to the original implementation without downsampling in Table~\ref{tab:downsampling_results}.

\begin{table}[H]
\centering
\begin{tabular}{l|c c c|c c c}
\hline \hline
& \multicolumn{3}{c|}{\textbf{max. samples}} & \multicolumn{3}{c}{\textbf{percentage}} \\
& 300 & 500 & 700 & 0.25 & 0.5 & 0.7 \\
\hline
Avg. F1 (w/ bg) & 69.02 & \textbf{75.51} & 66.32 & 64.88 & 73.27 & 73.61 \\
Avg. F1 (w/o bg) & 65.57 & \textbf{72.81} & 62.50 & 60.94 & 70.31 & 70.67 \\
Avg. IoU (w/ bg)  & 56.57 & \textbf{63.37} & 54.52 & 52.19 & 60.87 & 60.97 \\
Avg. IoU (w/o bg)  & 51.95 & \textbf{59.51} & 49.60 & 47.10 & 56.70 & 56.79 \\
FWIoU  & 55.41 & \textbf{61.58} & 57.67 & 52.01 & 59.85 & 61.00 \\
Balanced Accuracy  & 63.90 & \textbf{71.06} & 57.53 & 64.18 & 67.81 & 68.46 \\
Mean MCC  & 67.07 & \textbf{73.22} & 65.75 & 62.74 & 71.52 & 71.66 \\
\hline \hline
\end{tabular}
\caption{Comparison of different downsampling strategies integrated to E-FPN. \textit{max. samples} describes the maximum number of samples allowed, meaning random samples from the majority classes were removed. \textit{percentage} describes the fraction of each majority class used. Notably, a maximum of 500 samples performs best, achieving results that are slightly better than the baseline.}
\label{tab:diff_downsampling}
\end{table}

\begin{table}[H]
\centering
\begin{tabular}{c l|c c|c c}
\hline \hline
& & \multicolumn{2}{c|}{\textbf{w/o Downsampling}} & \multicolumn{2}{c}{\textbf{w/ Downsampling}} \\
\textbf{Label} & \textbf{Deficiency} & \textbf{F1 Score (\%)} & \textbf{IoU (\%)} & \textbf{F1 Score (\%)} & \textbf{IoU (\%)} \\
\hline
0 & Background & 97.21 & 94.57 & 97.08 & 94.33 \\
1 & Cracks & 42.51 & 26.99 & 44.29 & 28.44 \\
2 & Holes & 90.66 & 82.91 & 93.59 & 87.95 \\
3 & Roots & 82.33 & 69.96 & 84.58 & 73.28 \\
4 & Deformation & 77.08 & 62.70 & 78.69 & 64.87 \\
5 & Fracture & 57.57 & 40.42 & 53.15 & 36.19 \\
6 & Erosion & 73.24 & 57.78 & 71.52 & 55.66 \\
7 & Joint Problems & 77.91 & 63.82 & 72.27 & 56.58 \\
8 & Loose Gasket & 76.02 & 61.32 & 84.47 & 73.12 \\
\hline
\multicolumn{2}{r|}{Avg. F1 (w/ bg)} & \multicolumn{2}{c|}{74.95} & \multicolumn{2}{c}{75.51} \\
\multicolumn{2}{r|}{Avg. F1 (w/o bg)} & \multicolumn{2}{c|}{72.16} & \multicolumn{2}{c}{72.82} \\
\multicolumn{2}{r|}{Avg. IoU (w/ bg)} & \multicolumn{2}{c|}{62.27} & \multicolumn{2}{c}{63.38} \\
\multicolumn{2}{r|}{Avg. IoU (w/o bg)} & \multicolumn{2}{c|}{58.24} & \multicolumn{2}{c}{59.51} \\
\multicolumn{2}{r|}{FWIoU} & \multicolumn{2}{c|}{63.92} & \multicolumn{2}{c}{61.58} \\
\multicolumn{2}{r|}{Balanced Accuracy} & \multicolumn{2}{c|}{70.49} & \multicolumn{2}{c}{71.06} \\
\multicolumn{2}{r|}{Mean MCC} & \multicolumn{2}{c|}{73.38} & \multicolumn{2}{c}{73.22} \\
\hline \hline
\end{tabular}
\caption{Results for the original E-FPN model and the E-FPN model incorporating a downsampling technique, where all majority classes were randomly reduced to 500 samples each.}
\label{tab:downsampling_results}
\end{table}

\clearpage
\noindent
Although the IoU and F1 scores of the minority classes \textit{Holes}, \textit{Roots}, \textit{Deformation}, and \textit{Loose Gasket} improved, performance for the majority classes \textit{Fracture} and \textit{Joint Problems} decreased. Since downsampling, when combined with data augmentation, DLI, or with both methods, reduced the model’s performance, we decided to use only data augmentation and DLI. In our case, the dataset exhibits an intrinsic class imbalance, and the distribution of classes is consistent across the training, validation, and test sets, as well as in new data. Maintaining this natural distribution in the training set appears to benefit the model and improve performance. Therefore, downsampling does not support the training process in our specific case. Still, it remains a simple and relevant method in other domains, particularly when aiming to balance datasets, reduce training size or lower computational costs~\cite{shorten19}.  

\subsection{Final Preprocessing Pipeline}\label{data_results}

\noindent
Table~\ref{tab:combined_results} demonstrates the advantage of combining data augmentation with DLI, as well as the increase in performance compared to the baseline model. On average, the IoU and F1 score increased by around five percentage points across all classes. All classes benefit from the combination of data augmentation and DLI. The most notable improvements were achieved for the classes \textit{Cracks}, \textit{Fracture}, and \textit{Loose Gasket}, each showing an increase between nine and thirteen percentage points.
\begin{table}[H]
\centering
\begin{tabular}{c l|c c|c c}
\hline \hline
& & \multicolumn{2}{c|}{\makecell{\textbf{E-FPN w/o DLI} \\ \textbf{and original dataset}}} & \multicolumn{2}{c}{\makecell{\textbf{E-FPN w/ DLI} \\ \textbf{and augmented dataset}}} \\
\textbf{Label} & \textbf{Deficiency} & \textbf{F1 Score (\%)} & \textbf{IoU (\%)} & \textbf{F1 Score (\%)} & \textbf{IoU (\%)} \\
\hline
0 & Background & 97.21 & 94.57 & 97.59 & 95.30  \\
1 & Cracks & 42.51 & 26.99 & 53.41 & 36.43 \\
2 & Holes & 90.66 & 82.91 & 95.17 & 90.79 \\
3 & Roots & 82.33 & 69.96 & 85.29 & 74.35 \\
4 & Deformation & 77.08 & 62.70 & 79.60 & 66.12 \\
5 & Fracture & 57.57 & 40.42 & 66.62 & 49.95 \\
6 & Erosion & 73.24 & 57.78 & 76.27 & 61.64 \\
7 & Joint Problems & 77.91 & 63.82 & 79.97 & 66.63  \\
8 & Loose Gasket & 76.02 & 61.32 & 84.74 & 73.52 \\
\hline
\multicolumn{2}{r|}{Avg. F1 (w/ bg)} & \multicolumn{2}{c|}{74.95} & \multicolumn{2}{c}{79.85} \\
\multicolumn{2}{r|}{Avg. F1 (w/o bg)} & \multicolumn{2}{c|}{72.16} & \multicolumn{2}{c}{77.63} \\
\multicolumn{2}{r|}{Avg. IoU (w/ bg)} & \multicolumn{2}{c|}{62.27} & \multicolumn{2}{c}{68.30} \\
\multicolumn{2}{r|}{Avg. IoU (w/o bg)} & \multicolumn{2}{c|}{58.24} & \multicolumn{2}{c}{64.93} \\
\multicolumn{2}{r|}{FWIoU} & \multicolumn{2}{c|}{63.92} & \multicolumn{2}{c}{69.15} \\
\multicolumn{2}{r|}{Balanced Accuracy} & \multicolumn{2}{c|}{70.49} & \multicolumn{2}{c}{77.76} \\
\multicolumn{2}{r|}{Mean MCC} & \multicolumn{2}{c|}{73.38} & \multicolumn{2}{c}{77.99} \\
\hline \hline
\end{tabular}
\caption{Results of E-FPN model incoporating DLI and trained on augmented dataset. Data augmentation and DLI helps to improve the results for multiple classes, increasing the average F1 score and IoU by around 5~\%. }
\label{tab:combined_results}
\end{table}

\noindent
To emphasize the applicability of our preprocessing steps, we evaluated their impact on different state-of-the-art models. These included the original U-Net architecture, several of its adaptations, different types of feature pyramid networks, and vision transformers. For a better comparison between our implementation and other state-of-the-art models, we included information about parameter counts, floating-point operations (FLOPs), and size. Table~\ref{tab:model_comparison_100} shows the results using the augmented dataset with dynamic label injection, while Table~\ref{tab:model_comparison_baseline} presents the baseline results using the original dataset without any augmentation or label injection. A comparison of these tables shows that all models benefit from the proposed preprocessing steps. The median F1-score improved by eleven percentage points, while the median IoU increased by thirteen percentage points.

\begin{table}[htp]
\centering
\scriptsize
\setlength{\tabcolsep}{3.5pt}
\resizebox{1.0\textwidth}{!}{%
\begin{tabular}{l|ccc|cc|cc|ccc}
\toprule
\multirow{2}{*}{\textbf{Model}}
 & \multicolumn{3}{c}{\textbf{Params}}
 & \multicolumn{2}{c}{\textbf{F1 Score}}
 & \multicolumn{2}{c}{\textbf{mIoU}}
 & \multirow{2}{*}{\textbf{Bal. Acc.}}
 & \multirow{2}{*}{\textbf{Mean MCC}}
 & \multirow{2}{*}{\textbf{FW IoU}} \\
\cmidrule(lr){2-4}\cmidrule(lr){5-6}\cmidrule(lr){7-8}
& \textbf{(M)} & \textbf{GFLOPS} & \textbf{Size (MB)}
 & \textbf{w/bg} & \textbf{w/o}
 & \textbf{w/bg} & \textbf{w/o}
& & & \\
\midrule
U-Net\cite{ronneberger2015u}   & 31.04 & 13.69 & 118.47 & 75.40 & 72.64 & 63.14 & 59.14 & 73.02 & 73.33 & 67.01 \\
FPN\cite{lin2017feature}       & 21.20 & 7.81 & 80.89 & 76.84 & 74.26 & 64.51 & 60.68 & 73.38 & 74.77 & 67.32 \\
Att. U-Net\cite{oktay2018attention} & 31.40 & 13.97 & 119.85 & 78.81 & 76.46 & 66.92 & 63.35 & 77.80 & 76.74 & 68.47 \\
UNet++\cite{zhou2018unet++}    & 4.98 & 6.46 & 19.03 & 73.85 & 70.92 & 61.27 & 57.08 & 66.58 & 72.23 & 63.64 \\
BiFPN\cite{tan2020efficientdet} & 4.46 & 17.76 & 17.07 & 77.41 & 74.89 & 65.35 & 61.62 & 73.59 & 75.38 & 67.00 \\
SA-UNet\cite{guo2021sa}   & 7.86 & 3.62 & 1.85 & 78.79 & 76.43 & 66.87 & 63.30 & \textbf{78.65} & 76.87 & 68.16 \\
UNet3+\cite{huang2020unet}     & 25.59 & 33.04 & 97.68 & 78.50 & 76.11 & 66.63 & 63.04 & 76.89 & 76.41 & 68.29 \\
UNeXt\cite{valanarasu2022unext}   & 6.29 & 1.16 & 24.02 & 76.58 & 74.00 & 64.33 & 60.52 & 74.34 & 74.26 & 64.14 \\
EGE-UNet\cite{ruan2023ege} & 3.02 & 0.31 & 11.54 & 68.90 & 65.47 & 55.08 & 50.35 & 63.46 & 66.04 & 53.40 \\
Rolling UNet-L\cite{liu2024rolling} & 28.33 & 8.22 & 108.08 & 75.24 & 72.46 & 62.46 & 58.39 & 76.75 & 73.22 & 66.87 \\
\hline
HierarchicalViT U-Net\cite{ghahremani2024h} & 14.58 & 1.31 & 55.63 & 53.97 & 48.84 & 41.64 & 35.53 & 53.70 & 50.02 & 40.79 \\
Swin-UNet\cite{cao2021swin}      & 14.50 & 0.98 & 55.42 & 70.95 & 67.75 & 57.68 & 53.22 & 71.43 & 68.07 & 58.08 \\
MobileUNETR\cite{perera2024mobileunetr}           & 12.71 & 1.07 & 48.62 & 74.69 & 71.89 & 62.05 & 58.01 & 76.11 & 72.53 & 62.78  \\
Segformer\cite{xie2021segformer}             & 13.67 & 0.78 & 52.15 & 66.63 & 62.95 & 53.14 & 48.22 & 68.05 & 63.33 & 53.63 \\
FasterVit\cite{hatamizadeh2024fastervit}             & 25.23 & 1.57 & 96.27 & 68.35 & 64.78 & 55.21 & 50.37 & 62.68 & 66.20 & 58.30 \\
\hline
U-KAN\cite{li2024ukan}          & 25.36 & 1.73 & 96.98 & 77.41 & 74.93 & 65.27 & 61.59 & 77.65 & 75.19 & 64.50 \\
\hline
E-FPN~\cite{alshawi24b} &  8.45 & 1.83 & 32.26 & \textbf{79.85} & \textbf{77.63} & \textbf{68.30} & \textbf{64.93} & 77.76 & \textbf{77.99} & \textbf{69.15} \\ 
\hline
\bottomrule
\multicolumn{11}{l}{Note: M = Million, MB = Mega Byte, bg = background, Bal. Acc. = Balanced Accuracy, FW = Frequency Weighted,}\\
\multicolumn{11}{l}{MCC = Matthews Correlation Coefficient}
\end{tabular}%
}
\caption{Comparison of performance metrics across different models \\ with  100~\% of training data, data augmentation and dynamic label injection.}
\label{tab:model_comparison_100}
\end{table}

\begin{table}[htp]
\centering
\scriptsize
\setlength{\tabcolsep}{3.5pt}
\resizebox{1.0\textwidth}{!}{%
\begin{tabular}{l|ccc|cc|cc|ccc}
\toprule
\multirow{2}{*}{\textbf{Model}}
 & \multicolumn{3}{c}{\textbf{Params}}
 & \multicolumn{2}{c}{\textbf{F1 Score}}
 & \multicolumn{2}{c}{\textbf{mIoU}}
 & \multirow{2}{*}{\textbf{Bal. Acc.}}
 & \multirow{2}{*}{\textbf{Mean MCC}}
 & \multirow{2}{*}{\textbf{FW IoU}} \\
\cmidrule(lr){2-4}\cmidrule(lr){5-6}\cmidrule(lr){7-8}
& \textbf{(M)} & \textbf{GFLOPS} & \textbf{Size (MB)}
 & \textbf{w/bg} & \textbf{w/o}
 & \textbf{w/bg} & \textbf{w/o}
& & & \\
\midrule
U-Net\cite{ronneberger2015u}   & 31.04 & 13.69 & 118.47 & 60.18 & 55.67 & 46.14 & 40.31 & 61.97 & 57.27 & 49.28 \\
FPN\cite{lin2017feature}       & 21.20 & 7.81 & 80.89 & 59.39 & 54.87 & 45.71 & 39.98 & 59.23 & 56.53 & 49.02 \\
Att. U-Net\cite{oktay2018attention} & 31.40 & 13.97 & 119.85 & 67.54 & 63.92 & 53.55 & 48.59 & 69.79 & 64.91 & 53.97 \\
UNet++\cite{zhou2018unet++}    & 4.98 & 6.46 & 19.03 & 65.77 & 61.91 & 52.83 & 47.75 & 58.55 & 64.09 & 53.52 \\
BiFPN\cite{tan2020efficientdet} & 4.46 & 17.76 & 17.07 & 69.18 & 65.73 & 55.59 & 50.82 & 64.29 & 66.73 & 57.26 \\
SA-UNet\cite{guo2021sa}   & 7.86 & 3.62 & 1.85 & 68.75 & 65.27 & 54.97 & 50.18 & \textbf{73.15} & 66.22 & 57.24 \\
UNet3+\cite{huang2020unet}     & 25.59 & 33.04 & 97.68 & 72.12 & 69.04 & 58.85 & 54.49 & 71.63 & 69.38 & 58.15 \\
UNeXt\cite{valanarasu2022unext}   & 6.29 & 1.16 & 24.02 & 71.83 & 68.72 & 58.72 & 54.35 & 69.17 & 68.95 & 57.31 \\
EGE-UNet\cite{ruan2023ege} & 3.02 & 0.31 & 11.54 & 55.91 & 50.97 & 42.44 & 36.35 & 47.97 & 52.96 & 39.44 \\
Rolling UNet-L\cite{liu2024rolling} & 28.33 & 8.22 & 108.08 & 12.06 & 01.87 & 10.66 & 01.00 & 12.03 & 04.53 & 02.61 \\
\hline
HierarchicalViT U-Net\cite{ghahremani2024h} & 14.58 & 1.31 & 55.63 & 53.73 & 48.53 & 40.23 & 33.87 & 52.16 & 50.88 & 41.69 \\
Swin-UNet\cite{cao2021swin}      & 14.50 & 0.98 & 55.42 & 58.09 & 53.41 & 44.18 & 38.27 & 54.59 & 54.20 & 43.75 \\
MobileUNETR\cite{perera2024mobileunetr}           & 12.71 & 1.07 & 48.62  & 11.87 & 01.65 & 10.55 & 00.88 & 11.92 & 03.94 & 02.37   \\
Segformer\cite{xie2021segformer}             & 13.67 & 0.78 & 52.15 & 55.31 & 50.48 & 41.10 & 35.16 & 56.02 & 50.85 & 34.78 \\
FasterVit\cite{hatamizadeh2024fastervit}             & 25.23 & 1.57 & 96.27 & 13.11 & 03.04 & 11.32 & 01.73 & 12.85 & 05.33 & 05.09 \\
\hline
U-KAN\cite{li2024ukan}          & 25.36 & 1.73 & 96.98 & 69.81 & 66.43 & 56.37 & 51.69 & 68.48 & 67.27 & 57.62 \\
\hline
E-FPN~\cite{alshawi24b}  & 8.45 & 1.83 & 32.26 & \textbf{74.95} & \textbf{72.16} & \textbf{62.27} & \textbf{58.24} & 70.49 & \textbf{73.38} & \textbf{63.92} \\
\hline
\bottomrule
\multicolumn{11}{l}{Note: M = Million, MB = Mega Byte, bg = background, Bal. Acc. = Balanced Accuracy, FW = Frequency Weighted,}\\
\multicolumn{11}{l}{MCC = Matthews Correlation Coefficient}
\end{tabular}%
}
\caption{Comparison of performance metrics across different models \\ with the original dataset, without data augmentation and dynamic label injection.}
\label{tab:model_comparison_baseline}
\end{table}

\noindent
For a better comparison, Table~\ref{tab:visual_comp_data} visualizes the predicted segmentation masks produced by four models: (c) the baseline E-FPN model trained on the original dataset, (d) the baseline E-FPN model trained on the augmented dataset, (e) an advanced E-FPN model incorporating DLI trained on the original dataset, and (f) the advanced E-FPN model incorporating DLI trained on the augmented dataset. In addition to the predictions, the table also includes the input image (a) and the ground truth (b). Each row and color represents one defect class, with the class label shown in the first column.
It can be observed that the baseline E-FPN model (column c) shows the highest number of misclassifications, indicated by the presence of multiple incorrect colors in the segmentation output. While the model incorporating DLI (column e) performs better, it still predicts some incorrect defect classes. In contrast, our proposed model (column f) produces segmentation results most similar to the ground truth.\\
\\
Additional results supporting the effectiveness of our preprocessing pipeline are presented in the appendix~\ref{appx_data_preprocessing}. We evaluated experiments with reduced training data, or using either standard data augmentation techniques or dynamic label injection.

\begin{table}[htp]
  \centering
  \begin{tabular}{>{\centering\arraybackslash}m{0.5cm}>{\centering\arraybackslash}m{2cm}>{\centering\arraybackslash}m{2cm}>{\centering\arraybackslash}m{2cm}>{\centering\arraybackslash}m{2cm}>{\centering\arraybackslash}m{2cm}>{\centering\arraybackslash}m{2cm}}
    1) & \includegraphics[width=1.0\linewidth]{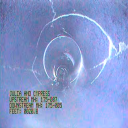} & \includegraphics[width=1.0\linewidth]{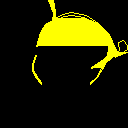} & \includegraphics[width=1.0\linewidth]{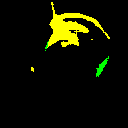} & \includegraphics[width=1.0\linewidth]{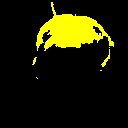} & \includegraphics[width=1.0\linewidth]{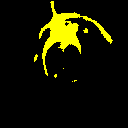} & \includegraphics[width=1.0\linewidth]{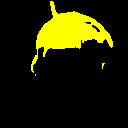} \\
    2) & \includegraphics[width=1.0\linewidth]{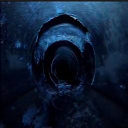} & \includegraphics[width=1.0\linewidth]{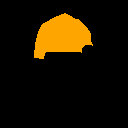} & \includegraphics[width=1.0\linewidth]{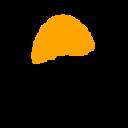} & \includegraphics[width=1.0\linewidth]{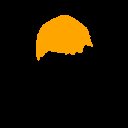} & \includegraphics[width=1.0\linewidth]{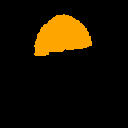} & \includegraphics[width=1.0\linewidth]{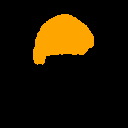} \\
    3) & \includegraphics[width=1.0\linewidth]{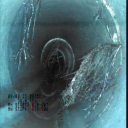} & \includegraphics[width=1.0\linewidth]{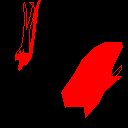} & \includegraphics[width=1.0\linewidth]{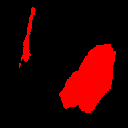} & \includegraphics[width=1.0\linewidth]{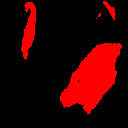} & \includegraphics[width=1.0\linewidth]{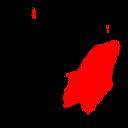} & \includegraphics[width=1.0\linewidth]{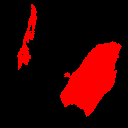} \\
    4) & \includegraphics[width=1.0\linewidth]{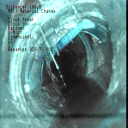} & \includegraphics[width=1.0\linewidth]{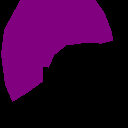} & \includegraphics[width=1.0\linewidth]{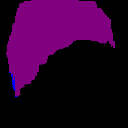} & \includegraphics[width=1.0\linewidth]{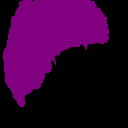} & \includegraphics[width=1.0\linewidth]{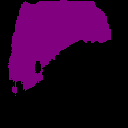} & \includegraphics[width=1.0\linewidth]{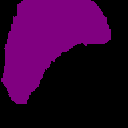} \\
    5) & \includegraphics[width=1.0\linewidth]{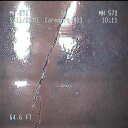} & \includegraphics[width=1.0\linewidth]{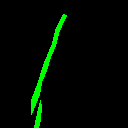} & \includegraphics[width=1.0\linewidth]{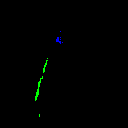} & \includegraphics[width=1.0\linewidth]{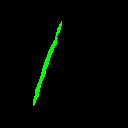} & \includegraphics[width=1.0\linewidth]{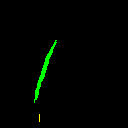} & \includegraphics[width=1.0\linewidth]{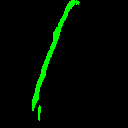} \\
    6) & \includegraphics[width=1.0\linewidth]{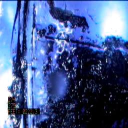} & \includegraphics[width=1.0\linewidth]{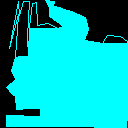} & \includegraphics[width=1.0\linewidth]{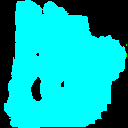} & \includegraphics[width=1.0\linewidth]{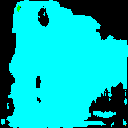} & \includegraphics[width=1.0\linewidth]{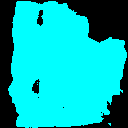} & \includegraphics[width=1.0\linewidth]{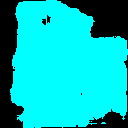} \\
    7) & \includegraphics[width=1.0\linewidth]{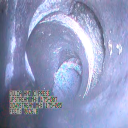} & \includegraphics[width=1.0\linewidth]{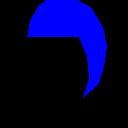} & \includegraphics[width=1.0\linewidth]{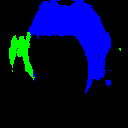} & \includegraphics[width=1.0\linewidth]{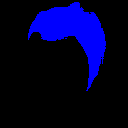} & \includegraphics[width=1.0\linewidth]{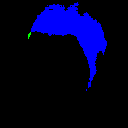} & \includegraphics[width=1.0\linewidth]{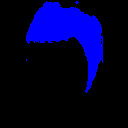} \\
    8) & \includegraphics[width=1.0\linewidth]{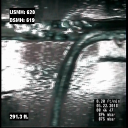} & \includegraphics[width=1.0\linewidth]{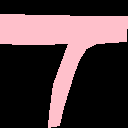} & \includegraphics[width=1.0\linewidth]{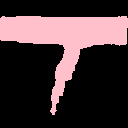} & \includegraphics[width=1.0\linewidth]{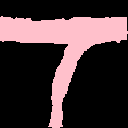} & \includegraphics[width=1.0\linewidth]{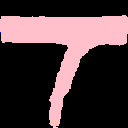} & \includegraphics[width=1.0\linewidth]{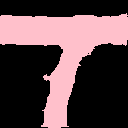} \\
    & (a) & (b) & (c) & (d) & (e) & (f)\\
  \end{tabular}
  \caption{Visual comparison of different methods to improve the model's performance. Each row and color represents a different defect class. (a) original image. (b) ground-truth. (c) baseline E-FPN model using original dataset. (d) baseline E-FPN model using augmented dataset. (e) E-FPN model incorporating DLI using original dataset. (f) E-FPN model incorporating DLI using augmented dataset. }
  \label{tab:visual_comp_data}
\end{table}

\subsection{Conclusion}
In this chapter, we evaluated different preprocessing strategies to improve defect segmentation performance. Traditional data augmentation was very effective, significantly improving both IoU and F1 score across all classes, with particularly strong improvements for the classes \textit{Cracks}, \textit{Holes}, \textit{Fracture}, and \textit{Loose Gasket}. Dynamic label injection further enhanced performance, increasing the results between four and ten percentage points for \textit{Cracks}, \textit{Holes}, and \textit{Loose Gasket}. Downsampling, however, did not improve performance, regardless of the technique used or whether it was combined with augmentation or dynamic label injection. We believe this is the case because of the intrinsic class imbalance in our dataset, which is consistent across training, validation, and test splits, as well as in real-world data. Preserving this natural distribution appears to be more beneficial than artificially balancing the dataset. The combination of data augmentation and dynamic label injection provided the best overall results. An average performance improvement of around five percentage points was achieved in both IoU and F1 score. Results for every class improved, with the largest enhancements being observed for the classes \textit{Cracks}, \textit{Fracture}, and \textit{Loose Gasket}. These findings show that carefully designed augmentation strategies are essential for improving segmentation in defect detection tasks, and demonstrate the advantage of combining multiple methods, including dynamic label injection, rather than relying on a single approach.

\chapter{FORTRESS}\label{fortress}

We developed FORTRESS (Function-composition Optimized Real-Time Resilient Structural Segmentation)~\cite{thrainer25}, a new architecture balancing accuracy and efficiency through the combination of depthwise separable convolutions and adaptive Kolmogorov-Arnold Network (KAN) integration. FORTRESS is based on three key innovations. First, we replaced standard convolutions with depthwise separable convolutions, reducing the number of parameters for each layer. Second, we introduced adaptive KAN integration, which composes functions more efficiently and provides accurate feature representations while keeping the computational cost low. Third, we added a multi-scale attention mechanism to the decoder that fuses spatial, channel-wise, and KAN-enhanced features for better performance. Thanks to these improvements, we reduced the total number of parameters and the computational complexity by 91 percent. In this chapter, we will explain the model, its implementation, and the underlying concepts in more detail.

\section{Problem Formulation}\label{fortress_problem}
Let $\mathcal{I} = \{I_i\}_{i=1}^N$ be a set of $N$ structural inspection images. Each image $I_i \in \mathbb{R}^{H \times W \times C}$ shares the same dimensions, with the height and width being $H$ and $W$ respectively, and the image includes $C$ channels. $M_i\in \mathbb{R}^{H \times W \times K}$ represents the corresponding segmentation mask, where $K$ represents the number of semantic classes. These masks can be combined to a set $\mathcal{M} = \{M_i\}_{i=1}^N$. The objective is to learn an optimal mapping function $g_\phi : \mathcal{I}\rightarrow \mathcal{M}$ where $\phi$ represents the model parameters. This function should provide accurate segmentation results while remaining efficient enough for real-time use. FORTRESS achieves this by using a dual-optimization approach that balances segmentation performance and computational cost:
\begin{equation}
    \phi^* = \arg\min_\phi \left[ \mathcal{L}_{seg}(\phi)+ \lambda_{eff} \mathcal{C}_{comp}(\phi) \right],
\end{equation}
where $\mathcal{L}_{seg}$ denotes the segmentation loss, $\mathcal{C}_{comp}$ is a penalty term reflecting computational complexity, and $\lambda_{eff}$ is a weighting factor that controls the trade-off between accuracy and efficiency. To meet this objective, FORTRESS uses two key strategies. First, it replaces standard convolutions with depth-wise separable convolutions, which reduces the number of parameters per layer. Second, it only applies KAN function composition transformations when they provide a computational advantage.

\clearpage

\section{Architecture Overview}\label{fortress_architecture}
\begin{figure}[H]
\centering
\captionsetup{justification=centering,margin=0cm}
\includegraphics[width=1.0\textwidth]{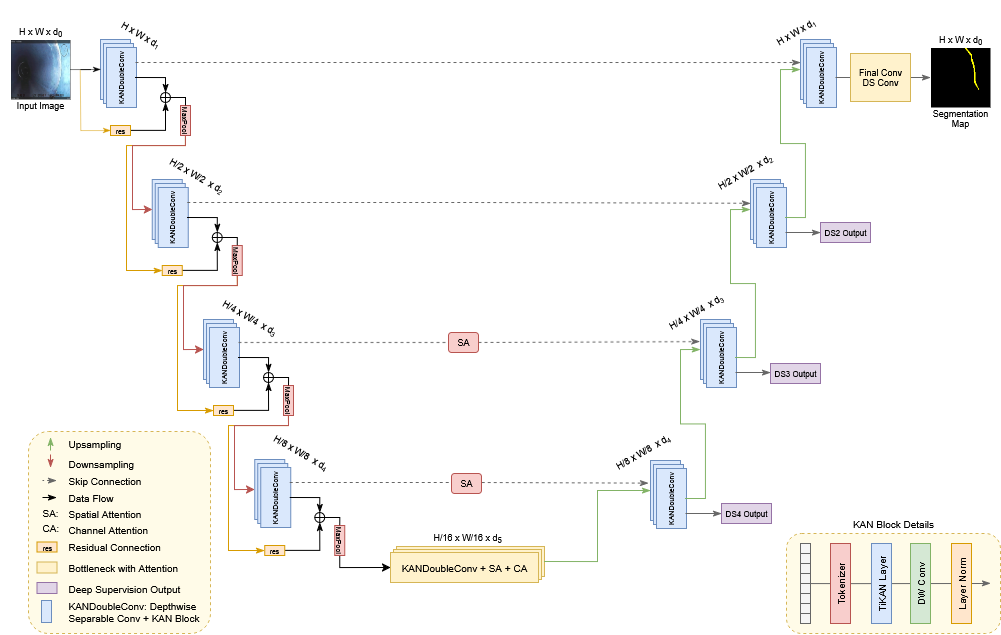}
\captionsetup{width=1.0\textwidth}
\caption{Architecture of FORTRESS. It follows an encoder-decoder structure. The encoder uses KAN-enhanced Double Convolution (KANDoubleConv) blocks with downsampling operations and residual connections. The final encoder layer integrates spatial attention (SA) and channel attention (CA) before handing over features to the decoder. Each decoder layer employs KANDoubleConv blocks and combines upsampled features from the previous decoder layer with encoder features from the corresponding level, which are passed through skip connections. An additional prediction head is added to each decoder level to enhance gradient stability and performance. The KANDoubleConv block uses depthwise separable convolutions and adaptive TiKAN transformations.}
\label{fig:fortress_architecture}
\end{figure}

The Fortress architecture efficiently addresses the unique challenges of structural image segmentation by integrating depthwise separable convolutions, adaptive KAN modules, and a multi-scale attention fusion mechanism. The architecture's novelty lies in its combination of spatial attention mechanisms, adaptive KAN-based feature representation, and computationally efficient depthwise separable convolutions. The architecture follows an encoder-decoder structure, as detailed in Section~\ref{edn}, and is illustrated in Figure~\ref{fig:fortress_architecture}. 

\noindent
Given an input image $I \in \mathbb{R}^{H \times W \times 3}$, FORTRESS generates a corresponding segmentation mask $\hat{M} \in \mathbb{R}^{H \times W \times K}$ by hierarchically processing the image through its encoder-decoder pipeline. The complete transformation can be described as:

\begin{equation}
    \hat{M} = \mathcal{G}_{FORTRESS} (I, \Theta) = \mathcal{D}_{\text{decoder}}(\mathcal{E}_{\text{encoder}}(I; \Theta_E); \Theta_D),
\end{equation}
where $\mathcal{D}_{\text{decoder}}$ and $\mathcal{E}_{\text{encoder}}$ describe the decoder and encoder components of the network, and $\Theta = [\Theta_E; \Theta_D]$ is the full set of learnable parameters. These parameters are optimized using depthwise separable convolutions to reduce computational cost without sacrificing accuracy. The encoder consists of five hierarchical layers, each of which includes a KANDoubleConv block. These blocks combine depthwise separable convolutions with adaptive KAN transformations for efficient and expressive feature extraction. Each encoder level $E_j$ is defined as:
\begin{equation}
    E_j = \mathcal{F}^{(j)}_{KANDouble}(\mathcal{MP}(E_{j-1}); \Theta_j) + \mathcal{R}_j(E_{j-1}), \quad j = \{1, \dotsc, 5\},
\end{equation}
where $\mathcal{F}^{(j)}_{KANDouble}$ represents the KAN-enhanced Double Convolution Blocks (KANDoubleConv) block at level $j$, applied to the output of the previous layer$E_{j-1}$. $E_0$ is set to the input image $I$. $\mathcal{MP}$ performs max pooling for spatial downsampling. Each level also includes a residual connection $\mathcal{R}_j$, which adds the previous layer's output directly to the result of the KANDoubleConv block. This residual path helps to prevent bottlenecks and supports stable gradient flow during training~\cite{alshawi24a}. Hierarchical feature processing allows the model to extract features across different spatial scales and to capture patterns more efficiently. In higher-resolution levels, such as level one and two, fine-grained features are extracted to capture defect boundaries and small defects. Coarse-grained features from lower-resolution levels, such as level four and five, capture a larger receptive field making it suitable to capture large-scale defects as well as contextual information~\cite{lin2017feature}. \\
\\
The decoder component includes upsampling mechanism and feature fusion operations to combine multi-scale information and preserve spatial precision. Each decoder level $D_d$ is defined as:
\begin{equation}
    D_d = \mathcal{F}^{(d)}_{\text{decoder}} (\mathcal{U}(D_{d+1}) \oplus E_{d}; \Theta_d),
\end{equation}
where the function $\mathcal{F}^{(d)}_{\text{decoder}}$ represents the decoder block at level $d$, parameterized by $\Phi_d$, which contains the learnable weights. $\mathcal{U}$ denotes the upsampling operation applied to the decoder output from the next deeper level $D_{d+1}$. This upsampled feature map is then concatenated, denoted by $\oplus$, with the corresponding encoder feature map $E_d$ at the same level to fuse semantic and spatial information. 

\section{Kolmogorov-Arnold Networks and Function Composition}\label{fortress_kan}
Fully convolutional networks like U-Net-based architectures are widely used for structural defect detection due to their ability to preserve spatial precision and capture fine structural details~\cite{sabet22}. However, these models face limitations when dealing with the complexity and variability of real-world structural defects. Defects often vary strongly in their visual appearance, requiring extensive computational resources to achieve accurate segmentation results~\cite{alshawi24b}. Additionally, class imbalance is a frequent challenge in defect datasets, as defects typically occupy only a small fraction of the image, while the majority remains defect-free~\cite{mahmoudi25}. The recent introduction of Kolmogorov-Arnold Networks (KANs) offers promising opportunities to address these challenges using a function composition-based learning paradigm. KANs are based on the Kolmogorov-Arnold representation theorem~\cite{kolmogorov1961}. This theorem states that any continuous multivariate function can be represented as a combination of a finite number of continuous univariate functions~\cite{brateanu24}. In contrast, traditional neural networks approximate functions through layers of linear transformations followed by fixed activation functions~\cite{liu24}. Liu et al.~\cite{liu24} introduced KANs in 2024 and demonstrated that the method can effectively approximate different functions. Decomposing complex multivariate functions into multiple univariate functions achieves a simpler and more interpretable representation while enhancing efficiency.

\section{FORTRESS Module Details}\label{fortress_modules}
This section outlines the key components of FORTRESS. First, we introduce the KAN-enhanced Double Convolution Blocks, which serve as the fundamental building blocks of the architecture. Next, we describe feature enhancement techniques using KAN transformations, spatial attention, and channel attention. Additionally, we detail the residual connections, which help maintain stable gradient flow. Finally, we describe the deep supervision mechanism, which further improves feature representation.

\subsection{KAN-enhanced Double Convolution Block}
The KAN-enhanced Double Convolution Block (KANDoubleConv) is the fundamental building block of FORTRESS. It incorporates depthwise separable convolutions for parameter-efficient implementation and integrates Kolmogorov–Arnold transformations. The KANDoubleConv block begins with two depthwise separable convolutions to extract spatial features, followed by adaptive TiKAN transformations. These transformations selectively enhance features by modeling complex functional relationships, but only if it is computationally efficient.  The full block can be defined as:

\begin{equation}
    \text{KANDoubleConv}(F_{in}) = \mathcal{R}(F_{in}) + \mathcal{T}_{\text{TiKAN}}(\mathcal{C}_{DS2}(\mathcal{C}_{DS1}(F_{in}))),
\end{equation}
where $\mathcal{C}_{DS2}$ and $\mathcal{C}_{DS1}$ are the two depthwise separable convolutions, $\mathcal{T}_{\text{TiKAN}}$ applies the adaptive TiKAN transformations, $F_{in}$ is the input feature map, and $\mathcal{R}$ represents a residual connection that supports stable gradient flow. The motivation behind KANDoubleConv is that CNNs perform well at detecting spatial patterns. However, this may be problematic for KANs~\cite{cang24, moradi24}. Incorporating depthwise separable convolution in KAN transformation block ensures that spatial locality is preserved while complex modeling of functional relationships is possible.

\subsection{Depthwise Separable Convolution}
The key innovation of FORTRESS is the replacement of standard convolutions with depthwise separable ones. Depthwise separable convolutions factorize a standard convolution in two operations, one depthwise convolution and a pointwise convolution~\cite{howard17}. The two depthwise separable convolutions in FORTRESS for an input $F_{in} \in \mathbb{R}^{H \times W \times C_{in}}$ can be split into four steps, which are defined as:
\begin{equation}
    F_{dw1} = \mathcal{DW}_{3\times3}(F_{in}; W_{dw1}),
\end{equation}
\begin{equation}
    F_{conv1} = \sigma_{ReLU}(\mathcal{BN}(\mathcal{PW}_{1\times1}(F_{dw1}; W_{pw1}, b_1))),
\end{equation}
\begin{equation}
    F_{dw2} = \mathcal{DW}_{3\times3}(F_{conv1}; W_{dw2}),
\end{equation}
\begin{equation}
    F_{conv2} = \sigma_{ReLU}(\mathcal{BN}(\mathcal{PW}_{1\times1}(F_{dw2}; W_{pw2}, b_2))),
\end{equation}
where $\mathcal{DW}_{3\times3}$ denotes a depthwise convolution with kernel size $3 \times 3$, $\mathcal{PW}_{1\times1}$ implements a pointwise convolution with an $1 \times 1$ kernel, $\sigma_{ReLU}$ applies a ReLU activation, and $\mathcal{BN}$ represents batch normalization. $W_{dw1}$, $W_{pw1}$, $W_{dw2}$, $W_{pw2}$ are the filter kernels used for the corresponding operations, while $b_1$ and $b_2$ are the associated bias terms for the pointwise convolutions.\\
\\
Depthwise convolution applies a $3 \times 3$ kernel independently to each feature channel, allowing the model to capture spatial context within each channel separately. Unlike standard convolutions, which compute across both spatial and channel dimensions simultaneously, depthwise convolution avoids cross-channel operations. This significantly reduces computational complexity while maintaining spatial locality. The operation can be defined as~\cite{kaiser17}:
\begin{equation}
    F_{dw1}[:,:,c] = \sum_{i=0}^2\sum_{j=0}^2 W_{dw1}[i,j,c] \cdot F_{in}[* + i, * + j, c],
\end{equation}
where $*$ indicates spatial convolution applied across all spatial locations, $W_{dw1}$ includes the filter kernel, and $c$ represents the individual channel that is processed. Pointwise convolution performs cross-channel feature blending by applying a $1\times1$ kernel. This allows the model to learn more complex relationships between channels, on top of the spatial features already captured by the depthwise convolution~\cite{kaiser17}. This two-stage approach provides similar representational capacity to standard convolution, but with significantly fewer parameters. A standard convolution with a $K \times K$ kernel, $C_{in}$ input channels, and $C_{out}$ output channels has in total $K^2 \cdot C_{in} \cdot C_{out}$ learnable parameters. Conversely, depthwise separable convolution reduces the number of parameters to $K^2 \cdot C_{in} + C_{in} \cdot C_{out}$, resulting in a reduction factor of~\cite{howard17}:
\begin{equation}
    \text{Reduction factor} = \frac{K^2 \cdot C_{in} \cdot C_{out}}{K^2 \cdot C_{in} + C_{in} \cdot C_{out}} = \frac{K^2 \cdot C_{out}}{K^2 + C_{out}}.
\end{equation}
With a typical kernel size of $K=3$, the theoretical maximum reduction factor is nine~\cite{howard17}. However, the actual savings may be lower.

\subsection{Adaptive TiKAN Integration Mechanism}
FORTRESS includes adaptive Kolmogorov-Arnold Network (KAN) transformations to enhance its ability to model the complex functional dependencies that typically occur in structural defect datasets.  To achieve a good balance between computational efficiency and expressive power, we integrated adaptive Tiny Kolmogorov-Arnold Networks (TiKAN).  The Canizaro Livingston Gulf States Center for Environmental Informatics, University of New Orleans, implemented TiKAN~\cite{ferdaus24, ferdaus2025karma} as a first approach to maintain KANs' representational power while reducing its computational requirements. Instead of applying KAN transformations uniformly across the architecture, FORTRESS uses the following adaptive activation criteria:
\begin{equation}
    \text{TiKAN\_Active}(F) = \left\{
        \begin{array}{ll}
            \text{True} & \text{if } C(F) \geq \gamma_c \text{ and } H(F) \times (W(F) \leq \gamma_s, \\
            \text{False} & \text{otherwise},
        \end{array} 
        \right.
\end{equation}
where $C(F)$, $H(F)$ and $W(F)$ represent the channels, height and width of feature map $F$. $\gamma_c = 16$ represent the minimum channel threshold, and $\gamma_s = 1024$ denotes the maximum spatial resolution threshold. Both thresholds are empirically determined to find balance between performance and efficiency. When TiKAN is activated, the input feature map $F$ is processed by a sequence of operations that implement the Kolmogorov-Arnold transformation principles. Adaptive TiKAN integration is defined as:
\begin{equation}
    \mathcal{T}_{TiKAN}(F) = \alpha \cdot \tau (F; \Phi) + (1 - \alpha) \cdot F,
\end{equation}
where $\alpha$ represents a learnable parameter, and $\tau (F; \Phi)$ applies the KAN transformations to $F$ while parameterized by $\Phi$. The transformed features are combined with the original features to produce the final output. The KAN transformation function $\tau$ is defined as~\cite{liu24}:
\begin{equation}
    \tau(F;\Phi) = \sum_{k=1}^K w_k \cdot [\phi_{base}(F) \cdot s_{base} + \phi_{spline}(\mathcal{DW}_{enhance}(F)) \cdot s_{spline}],
\end{equation}

where $w_k$ are learnable combination weights, $\phi_{base}$ and $\phi_{spline}$ represent base and spline-based univariate functions, where $s_{base}$ and $s_{spline}$ are the corresponding scaling factors. $\mathcal{DW}_{enhance}$ performs depthwise separable convolution to enhance the feature map $F$. The spline-based univariate functions $\phi_{spline}$ is composed of B-spline basis functions, where its control points are learnable. The spline-based univariate function is formulated as~\cite{liu24}:
\begin{equation}
    \phi_{spline}(x) = \sum_{i=0}^n c_i \cdot B_{i,p}(x),
\end{equation}
where $c$ represent the learnable control points, $n$ denotes the total number of control points, and $B_{i,p}$ defines a B-spline univariate function of degree $p$. The final KAN-enhanced features $F_{KAN}$ are defined as:
\begin{equation}
    F_{TiKAN} = \mathcal{T}_{TiKAN}(F_{attention}, \Phi),
\end{equation}
where $F_{attention}$ describes features processed by spatial and channel attention and $\Phi$ represents the KAN parameters.

\subsection{Residual Connection and Gradient Flow}
KANDoubleConv additionally incorporates residual connections to provide a stable gradient flow through the architecture while also incorporating KAN-based enhancements and depthwise separable convolutions. Residual connections are placed on multiple occasions in the feature processing pipeline to enhance training stability and convergence. Residual connections are especially important for complex relations provided by depthwise separable convolutions and TiKAN transformations. In FORTRESS, we utilized a dual optimization approach. First, a residual connection is placed to bypass the complete KANDoubleConv block as:
\begin{equation}
    F_{out} = F_{in} + \mathcal{KAN}_{Double} (F_{in}; \Theta),
\end{equation}
to ensure that the gradient information from the input is directly shared with the output. This is especially important for the TiKAN component, where complex computations may hinder stable gradient flow~\cite{alshawi24b}. Additionally, macro residual connections are placed inside the KAN enhancement blocks to bypass the depthwise separable convolutions as:
\begin{equation}
    F_{conv} = F_{conv} + \mathcal{C}_{DS}(F_{conv}; W_{ds}).
\end{equation}
 This approach allows parameter-efficient depthwise separable convolution while ensuring training stability through residual connections. The macro and micro residual connections ensure stable gradient flow by providing multiple possibilities for the gradient information to bypass complex computations. This is relevant for training stability and convergence. 
 
\subsection{Spatial Attention Mechanism}
The spatial attention mechanism is a helpful tool for improving the representation of features in images. By emphasizing relevant regions of an image and suppressing irrelevant ones, the model learns crucial features more effectively. This process mimics the way humans focus on areas of interest while overlooking those that are irrelevant. The SA-UNet~\cite{guo2021sa} architecture includes spatial attention mechanisms at various scales and achieved significant improvements in segmenting medical images. Its great performance in medical image segmentation suggests a potential adaptability for infrastructure defect detection. SA-UNet reduced the number of trainable parameters of the original UNet implementation from 31.04 million to a total of 7.86 million parameters. We used the SA-UNet architecture as an inspiration point due to its successful parameter reduction and its effective integration of spatial attention modules to enhance feature representation. \\
\\
Class imbalance is an omnipresent problem in defect detection datasets. Defects typically are only represented by a small part of the image, while a larger amount of pixels should be labeled as background~\cite{mahmoudi25}. FORTRESS incorporates a specialized spatial attention mechanism to mitigate this issue and offer precise localization and boundary detection. FORTRESS incorporates max and mean pooling strategies to enhance the feature representation, while enhancing its parameter efficiency with depthwise separable convolution. Max and mean pooling is performed in parallel on a feature map $F \in \mathbb{R}^{H \times W \times C}$ and defined as~\cite{hu18, luo24}:
\begin{equation}
    F_{avg}(i,j) = \frac{1}{C} \sum_{c = 1}^C F[i,j,c] \in \mathbb{R}^{H \times W \times 1},  
\end{equation}
\begin{equation}
    F_{max}(i,j) = \max_{c} F[i,j,c] \in \mathbb{R}^{H \times W \times 1} \quad \text{ where } c \in \{1, \dots, C\}.
\end{equation}

\noindent
First, feature statistics are created by pooling along the channel dimension. Average pooling provides contextual information by capturing the mean response, while max pooling highlights the most salient feature response at each spatial location~\cite{dougan23}. Its individual results are concatenated and passed on to a depthwise separable convolution, defined as:
\begin{equation}
    F_{concat} = [F_{avg}, F_{max}] \in \mathbb{R}^{H \times W \times 2},
\end{equation}
\begin{equation}
    A_{spatial} = \sigma_{sigmoid} (\mathcal{DS}_{k \times k} (F_{concat}; W_{spatial})),
\end{equation}
where $\mathcal{DS}$ denotes a depthwise separable convolution with adjusted kernel sizes $k \in\{3,5,7\}$ for different decoder levels. Fine-grained defects are detected using a small receptive field with a $3\times3$ kernel, whereas large-scale defects can be captured using a bigger receptive field with a $7\times7$ kernel. This allows an effective capture of defects at different scales~\cite{alshawi23a}. The resulting spatial attention weights $A_{spatial}$ are applied to the feature map $F$ as:
\begin{equation}
    F_{spatial} = A_{spatial} \odot F,
\end{equation}
where the symbol $\odot$ indicates element-wise multiplication, which selectively emphasizes relevant regions and suppresses irrelevant ones.

\subsection{Channel Attention Mechanism}
The channel-wise attention mechanism in FORTRESS uses solely average pooling to gather relevant information about the overall intensity distribution. The average pooling operation is defined as~\cite{hu18}:
\begin{equation}
    C_{avg} = \frac{1}{H \times W} \sum_{i=1}^H \sum_{j=1}^W F[i,j,:] \in \mathbb{R}^{C}.
\end{equation}
The resulting channel descriptors are processed by a multi-layer perceptron (MLP), which enables modelling of complex nonlinear relationships to emphasize important feature channels while suppressing irrelevant and noisy ones. Additionally, it reduces the channel dimensions by a factor of 16 and thus lowers the number of trainable parameters. The MLP is defined as~\cite{woo18}:
\begin{equation}
    \mathcal{MLP}(C) = W_2 \cdot \sigma_{ReLU}(W_1 \cdot C + b_1) + b_2,
\end{equation}
where $C$ denotes the channel descriptors, $W_1 \in \mathbb{R}^{(C/16)\times C }$ and $W_2 \in \mathbb{R}^{C \times (C/16)}$ denote trainable weight matrices, and $b_1$ and $b_2$ represent learnable bias vectors. The channel attention weights are computed by applying a sigmoid activation as\cite{woo18}:
\begin{equation}
    A_{channel} = \sigma_{sigmoid}(\mathcal{MLP}(C_{avg})).
\end{equation}
The sigmoid activation ensures that the attention scores remain between zero and one, avoiding negative or excessively large values~\cite{woo18}. Similar to the spatial attention weights, the resulting attention weights are applied to the feature map $F$ to receive its enhanced feature representation $F_{channel}$ as:
\begin{equation}
  F_{channel} = A_{channel} \odot F,  
\end{equation}
where the symbol $\odot$ denotes an element-wise multiplication. Channel attention mechanisms are beneficial for defect detection, as different types of defects often exhibit unique spectral characteristics that are best captured by different feature channels. For example, detecting cracks requires the model to focus on edge-sensitive channels, whereas identifying erosion relies more on textural information provided by different channels. The channel attention mechanism enhances FORTRESS’s ability to detect defects effectively by emphasizing the most relevant feature channels.


\subsection{Deep Supervision}
FORTRESS incorporates a deep supervision approach. Deep supervision improves gradient stability and enhances the model’s ability to learn feature representations at multiple resolutions, improving its capability to capture multi-scale defects~\cite{lee15}. Traditional deep supervision methods typically apply uniform supervision to all layers. In contrast, FORTRESS adopts an adaptive approach, recognizing that different decoder levels capture different levels of detail and semantic complexity.\\
\\
At each decoder level, an additional prediction head is added to generate segmentation predictions suitable for that level’s resolution and semantic complexity. These prediction heads are implemented using depthwise separable convolutions to remain consistent with the model’s parameter-efficient design.  The prediction head $P$ at decoder level $d$ is defined as:
\begin{equation}
    P^{(d)} = \sigma_{softmax}(\mathcal{DS}_{1 \times 1} (D_d; W_H^{(d)})),
\end{equation}
where $\sigma_{softmax}$ denotes the softmax activation, $\mathcal{DS}_{1 \times 1}$ is a depthwise separable convolution with a $1 \times 1$ kernel, $D_d$ represents the decoder features at level $d$, and $W_H^{(d)}$ are the trainable weights of the convolution. Each decoder level focuses on different spatial and semantic scales. Higher-resolution layers capture fine-grained information as defect boundaries and small-scale defects, while lower-resolution layers concentrate on large-scale features and contextual details~\cite{lin2017feature}. To consider this hierarchical structure, we apply different supervision weights:
\begin{equation}\label{eq:supervision_weights}
    \beta_d = \left\{
        \begin{array}{ll}
            \text{1.0} & \text{if } d = 1 \text{ (final output)}, \\
            \text{0.4} & \text{if } d = 2, \\
            \text{0.3} & \text{if } d = 3, \\
            \text{0.2} & \text{if } d = 4.
        \end{array} 
        \right.
\end{equation}
The final output receives the highest attention, while intermediate results receive progressively lower weights. This ensures that the final segmentation output remains the main optimization goal while enhancing its performance through intermediate supervision.

\section{Implementation and Evaluation}\label{fortress_implementation}
In this section, we report the experiments conducted with our proposed model, FORTRESS. We first describe the training setup. Then, we present the results for different settings and discuss their impact. Finally, we summarize the main findings and highlight the most important insights.

\subsection{Training Setup}
We conducted training for our model and the other state-of-the-art models on an NVIDIA A100 GPU (80GB) supported by 128 CPU cores and 1,007.6 GB of RAM. We used Python 3.12.2, PyTorch 2.5.1, and CUDA 12.1. Training and testing included 50 epochs with a batch size of 16. We used the culvert sewer defect dataset described in Section~\ref{culvert_sewer_defect_dataset}, containing 6,591 images with a resolution of 128 $\times$ 128. We tested our implementation of the original dataset, as well as our proposed preprocessing pipeline, which includes standard data augmentation techniques and dynamic label injection. We employed the FORTRESS architecture, as outlined in this chapter. For training, we used a deep supervision approach with a combined loss function consisting of cross-entropy, dice, and focal loss. At each decoder level, an additional prediction head generates a segmentation mask appropriate for that level’s resolution. For each level, the combined loss is calculated using predefined weights: 0.5 for cross-entropy, 0.3 for dice, and 0.2 for focal loss. Then, the loss values from all levels are combined using supervision weights, with the final output receiving the highest weight. The weights are shown in Equation~\ref{eq:supervision_weights}. Furthermore, we set the reduction parameter of cross-entropy loss to \textit{'None'}. This allows a deterministic implementation, while we compute the weighted mean manually after each loss step to align with PyTorch’s default \textit{'Mean'} behavior. We provide more details about loss functions in Section~\ref{loss_functions}. We optimized using AdamW with a learning rate of 0.001 and a weight decay of 1e-5. A CosineAnnealingLR scheduler was applied with 50 maximum iterations and a minimum learning rate of 1e-6. To stabilize training, we integrated gradient clipping with a maximum norm of 1.0. With the mentioned settings and a random seed of 42, we achieved the results listed in Section~\ref{fortress_results}. Our used metrics are described in detail in Section~\ref{metrics}.

\subsection{Results}\label{fortress_results}
Table~\ref{tab:fortress_model_comp} provides the comprehensive performance comparison on the culvert sewer pipe defect dataset using our preprocessing pipeline with standard data augmentation and dynamic label injection, as described in Section~\ref{data_experiments}. FORTRESS achieves state-of-the-art results with an F1-score of 0.771 and a mean IoU of 0.643, excluding background, surpassing SA-UNet by 0.7 percent in F1 and 1.0 percent in mean IoU while using 63 percent fewer parameters. SA-UNet has 7.86 million parameters and requires 3.62 GFLOPS, whereas FORTRESS achieves better performance with only 2.89 million parameters and 1.17 GFLOPS. In comparison, Table~\ref{tab:fortress_comp_without_aug} shows the results obtained using the original dataset without any data augmentation. Notably, SA-UNet’s performance dropped by more than ten percent, while U-Net and FPN declined by over fifteen percent. In contrast, FORTRESS remained stable, with only a minor decrease of one percent.

\begin{table}[H]
\centering
\scriptsize
\setlength{\tabcolsep}{3.5pt}
\resizebox{1.0\textwidth}{!}{%
\begin{tabular}{l|cc|cc|cc|ccc}
\toprule
\multirow{2}{*}{\textbf{Model}} 
 & \multicolumn{2}{c}{\textbf{Params}} 
 & \multicolumn{2}{c}{\textbf{F1 Score}} 
 & \multicolumn{2}{c}{\textbf{mIoU}} 
 & \multirow{2}{*}{\textbf{Bal. Acc.}} 
 & \multirow{2}{*}{\textbf{Mean MCC}} 
 & \multirow{2}{*}{\textbf{FW IoU}} \\
\cmidrule(lr){2-3}\cmidrule(lr){4-5}\cmidrule(lr){6-7}
 & \textbf{(M)} & \textbf{GFLOPS} 
 & \textbf{w/bg} & \textbf{w/o} 
 & \textbf{w/bg} & \textbf{w/o}
 & & & \\
\midrule
U-Net\cite{ronneberger2015u}   & 31.04 & 13.69 & 0.754 & 0.726 & 0.631 & 0.591 & 0.730 & 0.733 & 0.670 \\
FPN\cite{lin2017feature}       & 21.20 & 7.81 & 0.768 & 0.743 & 0.645 & 0.607 & 0.734 & 0.748 & 0.673 \\
Att. U-Net\cite{oktay2018attention} & 31.40 & 13.97 & 0.788 & 0.765 & 0.669 & 0.634 & 0.778 & 0.767 & 0.685 \\
UNet++\cite{zhou2018unet++}    & 4.98 & 6.46 & 0.739 & 0.709 & 0.613 & 0.571 & 0.666 & 0.722 & 0.636 \\
BiFPN\cite{tan2020efficientdet} & 4.46 & 17.76 & 0.774 & 0.749 & 0.654 & 0.616 & 0.736 & 0.754 & 0.670 \\
UNet3+\cite{huang2020unet}     & 25.59 & 33.04 & 0.785 & 0.761 & 0.666 & 0.630 & 0.769 & 0.764 & 0.683 \\
UNeXt\cite{valanarasu2022unext}   & 6.29 & 1.16 & 0.766 & 0.740 & 0.643 & 0.605 & 0.743 & 0.743 & 0.641 \\
EGE-UNet\cite{ruan2023ege} & 3.02 & 0.31 & 0.689 & 0.655 & 0.551 & 0.504 & 0.635 & 0.660 & 0.534 \\
Rolling UNet-L\cite{liu2024rolling} & 28.33 & 8.22 & 0.752 & 0.725 & 0.625 & 0.584 & 0.768 & 0.732 & 0.669 \\
\hline
HierarchicalViT U-Net\cite{ghahremani2024h} & 14.58 & 1.31 & 0.540 & 0.488 & 0.416 & 0.355 & 0.537 & 0.500 & 0.408 \\
Swin-UNet\cite{cao2021swin}      & 14.50 & 0.98 & 0.710 & 0.678 & 0.577 & 0.532 & 0.714 & 0.681 & 0.581 \\
MobileUNETR\cite{perera2024mobileunetr}           & 12.71 & 1.07 & 0.747 & 0.719 & 0.621 & 0.580 & 0.761 & 0.725 & 0.628 \\
Segformer\cite{xie2021segformer}             & 13.67 & 0.78 & 0.666 & 0.630 & 0.531 & 0.482 & 0.681 & 0.633 & 0.536 \\
FasterVit\cite{hatamizadeh2024fastervit}             & 25.23 & 1.57 & 0.684 & 0.648 & 0.552 & 0.504 & 0.627 & 0.662 & 0.583 \\
\hline
U-KAN\cite{li2024ukan}          & 25.36 & 1.73 & 0.774 & 0.749 & 0.653 & 0.616 & 0.777 & 0.752 & 0.645 \\
SA-UNet\cite{guo2021sa}       & 7.86 & 3.62 & 0.788 & 0.764 & 0.669 & 0.633 & 0.787 & 0.769 & 0.682 \\
\textbf{FORTRESS} & \textbf{2.89}  & \textbf{1.17}   & \textbf{0.793} & \textbf{0.771} & \textbf{0.677} & \textbf{0.643} & \textbf{0.787} & \textbf{0.772} & \textbf{0.690} \\
\bottomrule
\multicolumn{10}{l}{\scriptsize Note: M = Million, bg = background, Bal. Acc. = Balanced Accuracy, FW = Frequency Weighted,}\\
\multicolumn{10}{l}{\scriptsize MCC = Matthews Correlation Coefficient}
\end{tabular}%
}
\caption{Comparison of performance metrics across different models, including our proposed model, FORTRESS, and other state-of-the-art methods, using a preprocessing pipeline with standard data augmentation and dynamic label injection.}
\label{tab:fortress_model_comp}
\end{table}

\begin{table}[ht]
\centering
\scriptsize
\setlength{\tabcolsep}{3.5pt}
\resizebox{1.0\textwidth}{!}{%
\begin{tabular}{l|cc|cc|cc|ccc}
\toprule
\multirow{2}{*}{\textbf{Model}} 
 & \multicolumn{2}{c}{\textbf{Params}} 
 & \multicolumn{2}{c}{\textbf{F1 Score}} 
 & \multicolumn{2}{c}{\textbf{mIoU}} 
 & \multirow{2}{*}{\textbf{Bal. Acc.}} 
 & \multirow{2}{*}{\textbf{Mean MCC}} 
 & \multirow{2}{*}{\textbf{FW IoU}} \\
\cmidrule(lr){2-3}\cmidrule(lr){4-5}\cmidrule(lr){6-7}
 & \textbf{(M)} & \textbf{GFLOPS} 
 & \textbf{w/bg} & \textbf{w/o} 
 & \textbf{w/bg} & \textbf{w/o}
 & & & \\
\midrule
U-Net\cite{ronneberger2015u}   & 31.04 & 13.69 & 0.602 & 0.557 & 0.461 & 0.403 & 0.620 & 0.573 & 0.493 \\
FPN\cite{lin2017feature}       & 21.20 & 7.81 & 0.594 & 0.549 & 0.457 & 0.400 & 0.592 & 0.565 & 0.490 \\
Att. U-Net\cite{oktay2018attention} & 31.40 & 13.97 & 0.675 & 0.639 & 0.536 & 0.486 & 0.698 & 0.649 & 0.540 \\
UNet++\cite{zhou2018unet++}    & 4.98 & 6.46 & 0.658 & 0.619 & 0.528 & 0.478 & 0.586 & 0.641 & 0.535 \\
BiFPN\cite{tan2020efficientdet} & 4.46 & 17.76 & 0.692 & 0.657 & 0.556 & 0.508 & 0.643 & 0.667 & 0.573 \\
UNet3+\cite{huang2020unet}     & 25.59 & 33.04 & 0.721 & 0.690 & 0.589 & 0.545 & 0.716 & 0.694 & 0.582 \\
UNeXt\cite{valanarasu2022unext}   & 6.29 & 1.16 & 0.718 & 0.687 & 0.587 & 0.544 & 0.692 & 0.690 & 0.573 \\
EGE-UNet\cite{ruan2023ege} & 3.02 & 0.31 & 0.559 & 0.510 & 0.424 & 0.364 & 0.480 & 0.530 & 0.394 \\
Rolling UNet-L\cite{liu2024rolling} & 28.33 & 8.22 & 0.121 & 0.019 & 0.107 & 0.010 & 0.120 & 0.045 & 0.026 \\
\hline
HierarchicalViT U-Net\cite{ghahremani2024h} & 14.58 & 1.31 & 0.537 & 0.485 & 0.402 & 0.339 & 0.522 & 0.509 & 0.417 \\
Swin-UNet\cite{cao2021swin}      & 14.50 & 0.98 & 0.581 & 0.534 & 0.442 & 0.383 & 0.546 & 0.542 & 0.438 \\
MobileUNETR\cite{perera2024mobileunetr}           & 12.71 & 1.07 & 0.119 & 0.017 & 0.106 & 0.009 & 0.119 & 0.039 & 0.024 \\
Segformer\cite{xie2021segformer}             & 13.67 & 0.78 & 0.553 & 0.505 & 0.411 & 0.352 & 0.560 & 0.509 & 0.348 \\
FasterVit\cite{hatamizadeh2024fastervit}             & 25.23 & 1.57 & 0.131 & 0.030 & 0.113 & 0.017 & 0.129 & 0.053 & 0.051 \\
\hline
U-KAN\cite{li2024ukan}          & 25.36 & 1.73 & 0.698 & 0.664 & 0.564 & 0.517 & 0.685 & 0.673 & 0.576 \\
SA-UNet\cite{guo2021sa}       & 7.86 & 3.62 & 0.688 & 0.653 & 0.550 & 0.502 & 0.732 & 0.662 & 0.572 \\
\textbf{FORTRESS} & \textbf{2.89} & \textbf{1.17} & \textbf{0.783} & \textbf{0.759} & \textbf{0.663} & \textbf{0.627} & \textbf{0.791} & \textbf{0.763} & \textbf{0.674} \\
\bottomrule
\multicolumn{10}{l}{\scriptsize Note: bg = background, Bal. Acc. = Balanced Accuracy, FW = Frequency Weighted,}\\
\multicolumn{10}{l}{\scriptsize MCC = Matthews Correlation Coefficient}
\end{tabular}%
}
\caption{Comparison of performance metrics across different models, including our proposed model, FORTRESS, and other state-of-the-art methods, without data preprocessing.}
\label{tab:fortress_comp_without_aug}
\end{table}

\noindent
Figure~\ref{fig:comp_fortress} visualizes the predicted segmentation masks produced by FORTRESS on the culvert sewer pipe defect dataset. The results show precise boundary detection and reliable defect classification, even under challenging imaging conditions such as changing lighting, different surface textures, and varying defect scales.

\begin{figure}[H]
    \centering   %
    \captionsetup{justification=centering,margin=0cm}
    \includegraphics[width=0.4\textwidth]{figures/fortress_csdd_segmentation_results.png}
    \captionsetup{width=1\linewidth}
    \caption{Visual comparison of FORTRESS performance across various structural defect categories. Each row includes (left) the original image, (center) the corresponding ground truth, and (right) the predicted output generated by FORTRESS.}
    \label{fig:comp_fortress}
\end{figure}

\noindent
Additional results from experiments with reduced training data are presented in Appendix~\ref{appx_fortress}. These results demonstrate FORTRESS's effectiveness in data-sparse application areas.


\subsection{Conclusion}
In this chapter, we introduce FORTRESS, a novel architecture designed to balance accuracy and efficiency in defect segmentation. FORTRESS significantly reduces the number of trainable parameters without sacrificing performance by combining depthwise separable convolutions, adaptive Kolmogorov–Arnold Networks (KAN) integration, and a multi-scale attention mechanism. Additional components, such as deep supervision and residual connections, further improve feature representation and gradient flow. Experiments demonstrate that FORTRESS achieves state-of-the-art results on the culvert sewer pipe defect dataset. It reaches an F1 score of 0.771 and a mean IoU of 0.643 without the background class. This is while requiring 63 percent fewer parameters and substantially lower computational cost than SA-UNet. Notably, the model's performance remains stable even without data augmentation, unlike competing methods such as U-Net, FPN, and SA-UNet. They experience significant performance drops. Visual comparison to the ground-truth confirms that FORTRESS produces precise segmentation masks with accurate boundary detection and robust classification, handling diverse appearances and conditions well. These results highlight the effectiveness of integrating efficient convolutional operations with function composition and attention mechanisms. FORTRESS not only advances the state-of-the-art in sewer defect segmentation, but also demonstrates strong potential for real-time and data-sparse applications where both accuracy and efficiency are critical.

\chapter{Few-Shot Semantic Segmentation}\label{few_shot_segmentation}
Due to the limited availability of data in defect detection datasets, few-shot learning has considerable potential for this specific use case. Having a large structural defect dataset is not feasible because creating a large data collection is resource-intensive and requires manual annotations. Few-shot learning focuses on an efficient training process by only using a few annotated samples. Since no research has been conducted on its potential for structural defect detection, we implemented a few-shot semantic segmentation model and performed various experiments to verify its applicability in our specific situation. Section~\ref{fss_problem} defines the problem addressed by few-shot semantic segmentation, while Section~\ref{fss_implementation} outlines implementation details, including the encoder, prediction head, optimized loss calculation, and additional attention mechanisms to enhance the predictions. Section~\ref{fss_experiments} depicts the training setup and results obtained by various experiments as well as a final conclusion.

\section{Problem Definition}\label{fss_problem}
The goal is to train a segmentation model that is able to learn an appropriate feature representation and segment input images with only a few annotated images available. To fit the requirements of few-shot learning, we had to adjust the dataset. The training process now consists of multiple episodes instead of epochs. Rather than processing every training image during an epoch, the model now only sees a few samples per episode. We separate our dataset $\mathcal{D}$ into training set $\mathcal{D}_{train}$ and test set $\mathcal{D}_{test}$, so that $\mathcal{D}_{train} \cap \mathcal{D}_{test} = \emptyset$. Each set consists of several episodes, where each episode consists of a support set and a query set. In particular, $\mathcal{D}_{train} = \{(S_i, Q_i)\}^N_{i=1}$ and $\mathcal{D}_{test} = \{(S_i, Q_i)\}^N_{i=1}$, where $N$ defines the total number of episodes. Each episode includes an $n$-way $k$-shot segmentation task, where $n$ denotes the number of classes included per episode and $k$ represents the number of examples per class. Each episode during training and inference consists of a support set $S = \{(I_{c,j}, M_{c,j}) \; | \; c \in \{1, \dots, n\}, \; j\in\{1, \dots, k\}\}$ including $k$ pairs per class. The samples per class are selected randomly. The query set $Q = \{(I_c, M_c)\}_{c=1}^n$ includes one sample per class. The query set includes the same classes as the support set. If the number of classes per episode, $n$, is smaller than the total number of classes in the dataset, the classes per episode are selected randomly. The model first learns relevant features from the support set and then uses the extracted knowledge to perform segmentation on the query images. Because each episode includes different samples and classes, the model generalizes well. After training the segmentation model on $\mathcal{D}_{train}$, the model is evaluated using $\mathcal{D}_{test}$ over multiple episodes.  

\section{Implementation}\label{fss_implementation}
Our proposed few-shot semantic segmentation model is a prototypical network inspired by PANet~\cite{wang19}. As depicted in Section~\ref{prototypical_network}, the few-shot model consists of a feature extractor, as well as a prediction head. The feature extractor creates a suitable feature representation for the input images, and the prediction head uses the feature representation to generate a segmentation mask for the query image. The prediction head creates robust and suitable prototypes for each class label in an embedded space. Then the segmentation process is performed in the embedded space in a non-parametric manner. In our implementation, only the encoder is trainable. The prediction head is non-trainable, however in later experiments, we added different attention mechanisms and compared their influence to the original implementation, which makes the prediction head trainable as well. The attention mechanisms are outlined in Section~\ref{fss_attention}.

\begin{figure}[H]
    \centering
    \captionsetup{justification=centering,margin=0cm}
    \includegraphics[width=1.0\textwidth]{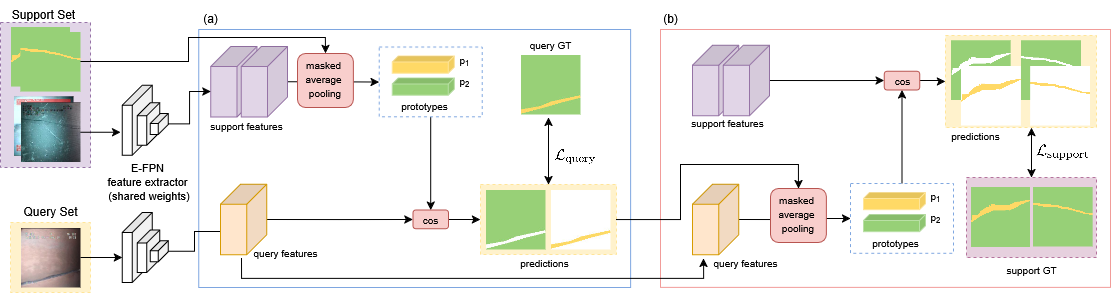}
    \captionsetup{width=1.0\textwidth}
    \caption{Architecture of our few-shot semantic segmentation model. The implementation follows a bidirectional approach: (a) First, masked average pooling creates representative prototypes using the support features and ground truth masks. The query features are compared to the prototypes using cosine similarity to find the most similar class label to create a suitable segmentation mask. (b) These predictions are used in the second step for a query-to-support few-shot semantic segmentation. This bidirectional approach ensures a better alignment of support and query features and allows the model to extract richer knowledge about the semantic classes. The loss values $\mathcal{L}_{\text{query}}$ and $\mathcal{L}_{\text{support}}$ are computed between the predictions and the ground truth.}
    \label{fig:fss_architecture}
\end{figure}

\noindent
Figure~\ref{fig:fss_architecture} illustrates the architecture of our few-shot semantic segmentation model. Each episode, the model receives a support and query set. The support set consists of images and corresponding ground-truth masks, whereas the query set only consists of images. We feed the images of both sets through a shared feature extractor to create an embedded feature representation. Masked average pooling creates representative prototypes out of the support features.  Each semantic class is represented in the embedded space by one prototype. To create a suitable segmentation for the query images, each pixel is labeled as the nearest prototype. We provide a detailed description of prototype learning and the segmentation process in Section~\ref{fss_predictionhead}. Our prototypical network follows a bidirectional approach. After we create segmentation masks for the provided query images using the support set, we use the generated query masks to segment the support images again. This means that few-shot learning is performed in a reverse manner, where the predicted masks and query images are the new support set, and the support images form the new query set. Intuitively, if the prototypes from support features create good segmentation masks for the query images, then the prototypes extracted from those query masks should also produce good segmentations on the support set. This forces an alignment of the support and query features, allowing the model to extract richer knowledge for each semantic class~\cite{wang19}. First, Section~\ref{fss_encoder} outlines the used feature extractor, while Section~\ref{fss_predictionhead} includes a detailed description of the prediction head, including prototype learning and segmentation, and Section~\ref{fss_loss} presents the loss calculation. Section~\ref{fss_attention} describes additional attention mechanisms for the prediction head to improve the final segmentation.

\subsection{Feature Extractor}\label{fss_encoder}
\begin{figure}[H]
    \centering
    \captionsetup{justification=centering,margin=0cm}
    \includegraphics[width=0.7\textwidth]{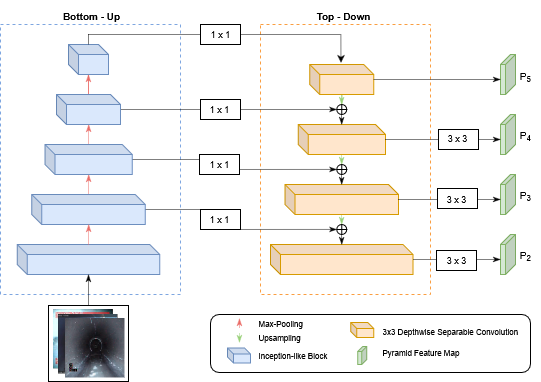}
    \captionsetup{width=1.0\textwidth}
    \caption{Feature Extractor part of E-FPN. The bottom-up pathway filters and downsamples feature maps using Inception-Like Blocks. The top-down pathway performs upsampling as well as feature fusion to receive final feature maps $\text{P}_2$,~$\dots$,~$\text{P}_5$.}
    \label{fig:fss_encoder}
\end{figure}

We extracted the encoder part of the Enhanced Feature Pyramid Network (E-FPN) architecture, introduced by Alshawi et al.~\cite{alshawi24b}. E-FPN is a feature pyramid network specialized for structural defect detection. Figure~\ref{fig:fss_encoder} illustrates the encoder and its components. The bottom-up pathway consists of Inception-like blocks and max pooling operations. It increases the number of channels to capture more abstract and richer features. An input image has three feature channels, and a height and width of $128$. In each level, the feature representation is downsampled by a factor of two. So, in the end, we achieve 1024 channels and a height and width of $8 \times 8$. The top-down pathway performs feature fusion and upsampling operations to generate the final pyramid feature maps $\text{P}_2$,~$\dots$,~$\text{P}_5$. These feature maps are returned and used for the prediction head.

\noindent
Before the prediction head creates pixel-wise predictions, the pyramid feature maps are fused to combine their information and provide the prediction head with more details and intricacies of individual defect classes. The feature maps $\text{P}_3$, and $\text{P}_4$ are interpolated using bilinear mode to match the dimensions of $\text{P}_2$ before simply adding them up. The above mentioned steps are performed on the support and query set individually. The final support feature and query features are used for the prototype creation and predictions described in the next section.

\subsection{Prediction Head}\label{fss_predictionhead}
Our prediction head uses the features provided by the feature extractor to first create accurate prototypes and then segment the query image by finding the nearest prototype for each pixel. Our implementation follows the prototypical approach introduced by Snell et al.~\cite{snell17}, where one well-separated and representative prototype is computed for each semantic class, including the background. Masked average pooling~\cite{zhang20} averages the feature vectors only over pixels belonging to a certain class. These feature vectors are extracted using the provided segmentation mask, and the prototype is created by averaging those masked features. Masked average pooling for class $c$ is defined as~\cite{wang19}:
\begin{equation}
    p_c = \frac{1}{k} \sum_{i=1}^k \frac{\sum_{x,y} F_{c,i}^{x,y} \mathbbm{1}(M_{c,i}^{x,y} = c)}{\sum_{x,y} \mathbbm{1}(M_{c,i}^{x,y} = c)} ,
\end{equation}
where $k$ indicates the number of samples per class, and $p_c$ denotes the prototype for class $c$. Given an input pair $(I_{c,i}, M_{c,i})$, $F_{c,i}^{x,y}$ represents the corresponding masked feature map at pixel position $(x,y)$. $\mathbbm{1}$ is an indicator function being one if the argument is true, or zero otherwise. The prototypes are optimized end-to-end and follow a non-parametric metric learning approach. It relies directly on the data and does not introduce new trainable parameters. Instead, it optimizes the model by measuring similarities and finetuning the feature extractor. The non-parametric metric learning performs the segmentation according to the computed prototypes as~\cite{wang19}:  

\begin{equation}
    M_{q,c}^{x,y} = \frac{\text{exp}(-\alpha \; d(F_q^{x,y}, p_c))}{\sum_{p_c \in \mathcal{P}}\text{exp}(-\alpha \; d(F_q^{x,y}, p_c))}.
\end{equation}

\noindent
The feature vector $F_q^{x,y}$ at spatial location $(x,y)$ is compared to each prototype $p_c \in \mathcal{P}$ using distance function $d$. $c \in \mathcal{C}_i$ indicates the semantic class. The scaling factor $\alpha$ is fixed at 20 following the PANet~\cite{wang19} implementation. We then deploy a softmax function over the distances to calculate the probability map $M_{q,c}$ for class $c$. The prediction mask $\hat{M}_q$ is computed as~\cite{wang19}:
\begin{equation}
    \hat{M}_q^{x,y} = \arg\max_c M_{q,c}^{x,y}.
\end{equation}
We employed cosine similarity as our distance function $d$. Snell et al.~\cite{snell17} suggested the use of squared Euclidean distance. However, Oreshkin et al.~\cite{oreshkin18} and Wang et al.~\cite{wang19} recommend using cosine similarity because, by multiplying the distance by a scaling factor, similar results to the squared Euclidean distance are achieved. Furthermore, Wang et al. state that cosine similarity exhibits a higher stability and enhances the model's performance because it is bounded, leading to an easier optimization. After segmenting the query images, the predictions are used to build prototypes from the query features and segment the support images. The same definitions apply, with the support set now acting as the query set, and vice versa.

\subsection{Loss Calculation}\label{fss_loss}
We use the cross-entropy loss to guide the training process. We calculate two loss values, namely $\mathcal{L}_{\text{query}}$ and $\mathcal{L}_{\text{support}}$. $\mathcal{L}_{\text{query}}$ is calculated by comparing the predicted segmentation masks and ground truth of the query set and $\mathcal{L}_{\text{support}}$ similarly for the support set. The query loss  $\mathcal{L}_{\text{query}}$ is calculated as~\cite{wang19}:
\begin{equation}
    \mathcal{L}_{\text{query}} = -\frac{1}{n \cdot H \cdot W} \sum_{q = 1}^n \sum_{y=1}^H \sum_{x=1}^W \left( \sum_{p_c \in \mathcal{P}} \mathbbm{1} (\tilde{M}_q^{x,y} = c) \text{ log } M_{q,c}^{x,y} \right) , 
\end{equation}
where $\tilde{M}_q$ denotes the ground truth mask belonging to the query image and $M_{q,c}$ represents the predicted segmentation mask for class $c$. The query set consists of $n$ query images, and we calculate the final loss value by considering the average of all samples. $\mathbbm{1}$ is an indicator function being one if the argument is true, and zero otherwise. The support loss $\mathcal{L}_{\text{support}}$ is calculated similarly, but the support set consists of $k \cdot n$ samples, where $k$ denotes the number of samples per class. Its computation is defined as~\cite{wang19}:
\begin{equation}
    \mathcal{L}_{\text{support}} = -\frac{1}{n \cdot k \cdot H \cdot W} \sum_{s = 1}^n \sum_{i = 1}^k \sum_{y=1}^H \sum_{x=1}^W \left( \sum_{p_c \in \mathcal{P}} \mathbbm{1} (\tilde{M}_{s,i}^{x,y} = c) \text{ log } M_{s,i}^{x,y} \right).
\end{equation}

\noindent
The total loss is calculated by simply adding the two individual terms as~\cite{wang19}:
\begin{equation}
    \mathcal{L} = \mathcal{L}_{\text{query}} + \mathcal{L}_{\text{support}}.
\end{equation}

\noindent
The two loss values are weighted uniformly. By incorporating a bidirectional approach, information from the query set flows back to the support set and forces the model to align the prototypes of the two sets in the embedded space. The model extracts richer information from the support set and exhibits higher generalizability as demonstrated by Wang et al.~\cite{wang19} and our experiments, described in Section~\ref{fss_results}.

\subsection{Attention Mechanisms in Prediction Head}\label{fss_attention}
To improve the segmentation results, we added different attention mechanisms inspired by transformers~\cite{vaswani17} to the prediction head. This transforms the prediction head to be trainable. We split the training process into two steps. First, we pretrain the feature extractor for $m$ episodes with the original non-trainable prediction head. Then, we train the prediction head for another $m$ episodes with the pretrained weights of the encoder while we also finetune the feature extractor. \\
\\
Attention allows models to focus on relevant areas in an image and suppress irrelevant ones. It follows an idea inspired by the human biological system. Our vision system tends to focus on relevant areas and overlooks irrelevant information to perceive our surroundings. In language and vision tasks, specific parts of the input are more important for the prediction than others. Attention mechanisms allow the model to dynamically adjust the attention weights depending on the input to achieve better performance on the task at hand~\cite{chaudhari21}. They are commonly used in natural language processing (NLP)~\cite{galassi20}, computer vision (CV)~\cite{khan22}, and recommender systems~\cite{zhang19b}, and are part of many state-of-the-art solutions. We employed self-attention, local self-attention and cross-attention. In the following, we will describe each approach individually.\\
\\
\textbf{Self-Attention.} Self-attention captures global context and long-range dependencies in an image and allows the model to focus on relevant parts while suppressing irrelevant ones. Self-attention connects each input token with all the others and allows a receptive field including the whole image. The importance of one pixel is therefore influenced by all other pixels~\cite{petit21}. The attention mechanism is defined as~\cite{petit21, zhou21}:
\begin{equation}\label{eq:self_attention}
    f_i =  \sum_{j = 1}^N \text{softmax}_j \left(\frac{q_i k_j^T}{\sqrt{d_k}}\right)v_j = \sum_{j=1}^N A_{i,j} v_j,
\end{equation}
where $N = H \cdot W$ is the total number of pixels, and $f_i$ represents the attention-weighted feature vector at pixel $i$. The features $q_i$, $k_j$, and $v_j$ denote the query, key and value vectors at location $i$ and $j$. The attention matrix $A$ includes the similarities between the query $q$ and key $k$ vector, and multiplying it with the value vectors $v$ produces a weighted sum over all pixels. The vectors $q$, $k$, and $v$ are different linear projections of the input feature map. They are obtained via the learned embedding matrices $W_q$, $W_k$, and $W_v$. The softmax function is applied over all $j$ for a fixed $i$ to bound the sum of similarity weights for pixel $i$ and its neighbors to one~\cite{petit21}. In our implementation, self-attention is applied to the support and query features individually before the prototype creation and segmentation process. \\
\\
\textbf{Local Self-Attention.} Local Self-Attention limits its attention to its neighbors to reduce computational complexity. It successfully combines the advantages of convolution and self-attention, namely local inductive bias and dynamic feature selection~\cite{pan23}. Local inductive bias states that nearby pixels share more similarities and are more strongly related than more distant ones. CNNs incorporate this bias in each layer throughout the whole model. Contrary to convolution, self-attention dynamically adjusts the attention weights to find the optimal feature representation~\cite{dosovitskiy21}. Local self-attention is a combination of both methods and is defined as~\cite{petit21, zhou21}:
\begin{equation}
    f_i = \sum_{j \in \Omega} \text{softmax}_j \left(\frac{q_i k_j^T}{\sqrt{d_k}}\right)v_j = \sum_{j \in \Omega} A_{i,j} v_j,
\end{equation}
where we calculate the attention-weighted features $f_i$ at pixel location $i$ by only considering the local window $\Omega$. This restricts the receptive field, so for each pixel, only a local window of neighboring pixels is considered. The remaining computation is identical to the standard self-attention as defined in Equation~\ref{eq:self_attention}. We applied local self-attention similar to the standard self-attention to the support and query features individually, to enhance their representation before the prediction head computes representative prototypes and generates predictions. \\
\\
\textbf{Cross-Attention.} Cross-attention is an adaptation of the standard self-attention, where not only the input sequence is considered, but also the desired output. The input forms the key vector $k$ and value vector $v$, whereas the query vector $q$ is defined using the output sequence~\cite{vaswani17}. Incorporating the desired output in the attention mechanism allows the model to detect relevant features and image regions more effectively and improve the subsequent matching~\cite{hou19}. In few-shot semantic segmentation, vectors $k$ and $v$ are linear projections of the support features, while $q$ is a linear projection of the query features. The mathematical definition of cross-attention is identical to standard self-attention as defined in Equation~\ref{eq:self_attention}, only the definition of vectors $q$, $k$, and $v$ varies.

\section{Experiments}\label{fss_experiments}
In this section, we report the experiments conducted with our few-shot semantic segmentation model. We first describe the training setup and implementation details. Then, we present the results for different settings and discuss their impact. Finally, we summarize the main findings and highlight the most important insights.

\subsection{Training Setup}
We trained our model and the other state-of-the-art models using an NVIDIA A100 80GB GPU with 128 CPU cores and 1,007.6 GB RAM. We used Python 3.12.2, PyTorch 2.5.1, and CUDA 12.1. Training and testing included 1,000 episodes, and we defined a batch size of 1. We used the culvert sewer defect dataset described in Section~\ref{culvert_sewer_defect_dataset}, consisting of 6,591 images of shape 128 $\times$ 128. We did not include any data augmentation techniques during training. As described in Section~\ref{fss_encoder}, we extracted the encoder part of E-FPN~\cite{alshawi24b} and implemented it as our feature extractor. Our head is inspired by PANet~\cite{wang19} and follows a prototypical approach, where prototypes are generated to represent the different semantic classes and the segmentation of an image is performed by finding the nearest prototype for each pixel. The features and prototypes are normalized using the Euclidean norm. We used the cross-entropy loss function and set the reduction to \textit{'None'}. This allows a deterministic implementation, while we compute the weighted mean manually after each loss calculation to approximate PyTorch's default implementation with reduction specified as \textit{'Mean'}. We used AdamW as our optimizer with a learning rate of 0.001 and a weight decay of 1e-5. The applied learning rate scheduler is CosineAnnealingLR with a maximum number of 50 iterations and a minimum learning rate of 1e-6. To prevent exploding gradients, we integrated gradient clipping with a max norm of 1.0 into our training loop. With the mentioned settings and a random seed of 42, we achieved the results listed in Section~\ref{fss_results}.

\subsection{Results}\label{fss_results}
 Table~\ref{tab:few_shot} shows the results for different $n$-way $k$-shot settings. Our dataset includes nine semantic classes, including background and eight defect classes. We ran experiments with $k \in \{1, 5\}$ and $n \in \{2, 4, 6, 8, 9\}$. For $k = 1$, the best performance was achieved when the support and query set included samples from eight or nine different classes. Similar results were obtained for $k = 5$, where nine distinctive classes were again the best solution. This suggests that training with more classes per episode helps the model learn more distinctive features, which improves its ability to differentiate between defect classes. Therefore, a more diverse set of classes during training is clearly beneficial. Additionally, including five samples per class in the training set produced higher overall results than using only one sample. This indicates that a single sample is not as representative of a class as a combination of multiple samples, suggesting that variation is present within a class.

\begin{table}[H]
\centering
\scriptsize
\setlength{\tabcolsep}{3.5pt}
\resizebox{1.0\textwidth}{!}{%
\begin{tabular}{cc|ccc|cc|cc|ccc}
\toprule
\multicolumn{5}{c}{\textbf{Parameter}}
 & \multicolumn{2}{c}{\textbf{F1 Score}}
 & \multicolumn{2}{c}{\textbf{mIoU}}
 & \multirow{2}{*}{\textbf{Bal. Acc.}}
 & \multirow{2}{*}{\textbf{Mean MCC}}
 & \multirow{2}{*}{\textbf{FW IoU}} \\
\cmidrule(lr){1-5}\cmidrule(lr){6-7}\cmidrule(lr){8-9}
\textbf{n} & \textbf{k} & \textbf{(M)} & \textbf{GFLOPS} & \textbf{Size (MB)} & \textbf{w/bg} & \textbf{w/o}
 & \textbf{w/bg} & \textbf{w/o}
& & & \\
\midrule
2 & 1 & 8.41 & 1.77 & 32.10 & 58.64 & 54.54 & 46.07 & 41.31 & 60.45 & 54.60 & 42.76 \\
2 & 5 & 8.41 & 1.77 & 32.10 & 74.04 & 71.44 & 62.87 & 59.46 & 76.72 & 71.60 & 65.95 \\
\midrule
4 & 1 & 8.41 & 1.77 & 32.10 &  62.70 & 58.86 & 50.72 & 46.12 & 64.11 & 59.24 & 47.79 \\
4 & 5 & 8.41 & 1.77 & 32.10 &  75.48 & 72.96 & 64.35 & 60.94 & 79.34 & 73.34 & 67.61 \\
\midrule
6 & 1 & 8.41 & 1.77 & 32.10 &  64.50 & 60.85 & 53.92 & 49.62 & 69.41 & 61.64 & 53.81 \\
6 & 5 & 8.41 & 1.77 & 32.10 &  76.88 & 74.49 & 66.01 & 62.72 & 80.57 & 75.00 & 69.22 \\
\midrule
8 & 1 & 8.41 & 1.77 & 32.10 &  \textbf{66.00} & \textbf{62.40} & 55.64 & 51.33 & 68.24 & \textbf{63.37} & 57.07 \\
8 & 5 & 8.41 & 1.77 & 32.10 &  78.10 & 75.84 & 67.20 & 64.01 & 81.82 & 76.30 & 71.10 \\
\midrule
9 & 1 & 8.41 & 1.77 & 32.10 &  65.65 & 62.04 & \textbf{55.65} & \textbf{51.39} & \textbf{68.35} & 63.01 & \textbf{57.71} \\
9 & 5 & 8.41 & 1.77 & 32.10 &  \textbf{78.99} & \textbf{76.84} & \textbf{68.07} & \textbf{64.99} & \textbf{83.67} & \textbf{77.48} & \textbf{71.42} \\
\hline
\bottomrule
\multicolumn{12}{l}{Note: M = Million, MB = Mega Byte, bg = background, Bal. Acc. = Balanced Accuracy,}\\
\multicolumn{12}{l}{FW = Frequency Weighted, MCC = Matthews Correlation Coefficient}
\end{tabular}%
}
\caption{Comparison of different training and test setup for few-shot semantic segmentation.}
\label{tab:few_shot}
\end{table}

\noindent
Table~\ref{tab:fs_bidirectional} demonstrates the advantage of including a bidirectional approach. According to Wang et al.~\cite{wang19}, an accurate segmentation of the query images allows the extracted prototypes to properly represent these classes and create an accurate segmentation of the support images. This enables data to flow in both directions, from the support set to the query set and vice versa, resulting in a richer feature representation. The bidirectional approach improves performance by approximately ten percent on all evaluation metrics.

\begin{table}[H]
\centering
\scriptsize
\setlength{\tabcolsep}{3.5pt}
\resizebox{1.0\textwidth}{!}{%
\begin{tabular}{c|cc|ccc|cc|cc|ccc}
\toprule
\multirow{2}{*}{\textbf{Method}}
 & \multicolumn{5}{c}{\textbf{Parameter}}
 & \multicolumn{2}{c}{\textbf{F1 Score}}
 & \multicolumn{2}{c}{\textbf{mIoU}}
 & \multirow{2}{*}{\textbf{Bal. Acc.}}
 & \multirow{2}{*}{\textbf{Mean MCC}}
 & \multirow{2}{*}{\textbf{FW IoU}} \\
\cmidrule(lr){2-6}\cmidrule(lr){7-8}\cmidrule(lr){9-10}
& \textbf{n} & \textbf{k} & \textbf{(M)} & \textbf{GFLOPS} & \textbf{Size (MB)}
 & \textbf{w/bg} & \textbf{w/o}
 & \textbf{w/bg} & \textbf{w/o}
& & & \\
\midrule
Unidirectional & 2 & 5 & 8.41 & 1.77 & 32.10 & 65.55 & 62.33 & 52.06 & 48.05 & 73.59 & 62.48 & 48.85 \\
Bidirectional & 2 & 5 & 8.41 & 1.77 & 32.10 & \textbf{74.04} & \textbf{71.44} & \textbf{62.87} & \textbf{59.46} & \textbf{76.72} & \textbf{71.60} & \textbf{65.95} \\

\hline
\bottomrule
\multicolumn{13}{l}{Note: M = Million, MB = Mega Byte, bg = background, Bal. Acc. = Balanced Accuracy, }\\
\multicolumn{13}{l}{FW = Frequency Weighted, MCC = Matthews Correlation Coefficient}
\end{tabular}%
}
\caption{Comparison of unidirectional and bidirectional prediction heads, showing that a bidirectional approach yields better results.}
\label{tab:fs_bidirectional}
\end{table}

\noindent
To improve the above results, we added trainable parameters to our initially non-trainable prediction head. We tested three different approaches, namely self-attention, local self-attention, and a cross-attention mechanism. We compared their results in Table~\ref{tab:proto_finetune}. Each attention block is applied before the prototypes are created, in order to enhance both the support and query features. For all three attention mechanisms, we used pretrained weights for the E-FPN encoder. The encoder was pretrained using a non-trainable prediction head and the best-performing parameter combination from Table~\ref{tab:few_shot}, which is the 9-way 5-shot setting. The encoder was trained for 1,000 episodes. After that, we trained the encoder together with the trainable prediction head for another 1,000 episodes. To ensure a fair comparison between the attention mechanisms, we also trained a model using the same pretrained weights and a non-trainable prediction head. This model is used as our baseline. As shown in Table~\ref{tab:proto_finetune}, self-attention outperforms all other attention mechanisms, including the baseline. Even though the baseline was trained for an additional 1,000 episodes, the results only increased by one percent, suggesting that 1,000 episodes are enough to train a suitable few-shot segmentation model. Furthermore, it is noticeable that the number of floating-point operations increases with the inclusion of each attention mechanism, with local self-attention causing the most dramatic increase. Thus, self-attention is not only more effective but also computationally more efficient than local self-attention.


\begin{table}[H]
\centering
\scriptsize
\setlength{\tabcolsep}{3.5pt}
\resizebox{1.0\textwidth}{!}{%
\begin{tabular}{c|cc|ccc|cc|cc|ccc}
\toprule
\multirow{2}{*}{\textbf{Method}}
 & \multicolumn{5}{c}{\textbf{Parameter}}
 & \multicolumn{2}{c}{\textbf{F1 Score}}
 & \multicolumn{2}{c}{\textbf{mIoU}}
 & \multirow{2}{*}{\textbf{Bal. Acc.}}
 & \multirow{2}{*}{\textbf{Mean MCC}}
 & \multirow{2}{*}{\textbf{FW IoU}} \\
\cmidrule(lr){2-6}\cmidrule(lr){7-8}\cmidrule(lr){9-10}
& \textbf{n} & \textbf{k} & \textbf{(M)} & \textbf{GFLOPS} & \textbf{Size (MB)}
 & \textbf{w/bg} & \textbf{w/o}
 & \textbf{w/bg} & \textbf{w/o}
& & & \\
\midrule
Baseline & 2 & 5 & 8.41 & 1.77 & 32.10 & 80.24 & 78.28 & 69.48 & 66.65 & 83.06 & 78.37 & 73.76 \\
w/SA & 2 & 5 & 8.52 & 3.38 & 32.54 & \textbf{82.81} & \textbf{81.11} & \textbf{72.38} & \textbf{69.80} & 86.03 & \textbf{81.17} & \textbf{76.52} \\ 
w/LSA & 2 & 5 & 8.54 & 24.32 & 32.60 & 81.81 & 80.07 & 70.61 & 67.95 & \textbf{89.15} & 80.37 & 72.61 \\
w/CA & 2 & 5 & 8.54 & 2.84 & 32.60 & 81.44 & 79.60 & 70.74 & 67.98 & 82.09 & 79.95 & 75.67 \\
\hline
\bottomrule
\multicolumn{13}{l}{Note: SA = Self-Attention, LSA = Local-Self-Attention, CA = Cross-Attention, M = Million, MB = Mega Byte,}\\
\multicolumn{13}{l}{bg = background, Bal. Acc. = Balanced Accuracy, FW = Frequency Weighted, MCC = Matthews Correlation Coefficient}
\end{tabular}%
}
\caption{Comparison of different attention mechanisms for few-shot semantic segmentation, where self-attention yields the best results.}
\label{tab:proto_finetune}
\end{table}

\subsection{Conclusion}
In this chapter, we investigate few-shot learning for defect segmentation. Few-shot learning is well suited for this task because collecting large datasets is costly and requires manual annotations. Our model uses a feature extractor and a prediction head. The extractor builds a suitable feature space, and the prediction head generates segmentation masks using class prototypes. In the basic version, only the encoder is trainable, while the prediction head is non-trainable. Later, we extend it with attention mechanisms that also make the prediction head trainable. We adopt a bidirectional approach. After segmenting the query images with the support set, the predicted query masks are used to segment the support images again. This forces alignment between support and query features and enables richer representations. Experiments confirm that this approach improves generalization and segmentation quality. We also tested various $n$-way $k$-shot configurations. The 9-way 5-shot setting yields the best results, suggesting that training with more classes per episode leads to better performance. Including data diversity benefits our structural defect dataset. Since defects vary significantly in their visual appearance, using multiple samples to create a representative prototype is better than using one sample. Additionally, including multiple classes allows the feature extractor to focus on distinguishing features. For our dataset, including all nine classes per episode produces the best results. Finally, we integrate attention mechanisms into the prediction head. We evaluate self-attention, local self-attention, and cross-attention. All methods improve performance compared to the non-trainable baseline. Among them, self-attention achieves the highest scores. These findings demonstrate that few-shot learning combined with bidirectional training and attention is a promising approach for defect segmentation under data-scarce conditions.

\chapter{Conclusion}\label{conclusion}

In this thesis, we studied various strategies to improve defect segmentation in situations with limited data. First, we evaluated different data preprocessing methods. Then, we designed a suitable model architecture called FORTRESS, which incorporates various enhancements to boost performance. Lastly, we investigated the applicability of few-shot semantic segmentation to our culvert sewer defect dataset. \\
\\
First, we explored different data augmentation techniques, dynamic label injection and downsampling operations to improve the training data and boost performance. Data augmentation and dynamic label injection significantly improved Intersection over Union (IoU) and F1 score, while downsampling was not effective. This indicates that preserving the natural class distribution is more beneficial than artificially balancing the dataset. A combination of data augmentation and dynamic label injection achieved the best results, leading to substantial improvements across all classes. These findings support the strategy of increasing data diversity to improve segmentation performance in defect detection tasks and indicate that a combination of multiple methods outperforms using a single one. However, the preprocessing pipeline must be designed carefully and fine-tuned multiple times to achieve the best trade-off between performance and training efficiency.\\
\\
Second, we introduced FORTRESS, a novel architecture that balances accuracy and efficiency in defect detection tasks. By combining depthwise separable convolutions, adaptive Kolmogorov–Arnold Networks (KAN), and multi-scale attention, FORTRESS reduces parameters and computational cost while maintaining high performance. It achieves state-of-the-art results on the culvert sewer defect dataset while using 63 percent fewer parameters than the second-best-performing model. It is also notable that the results remain stable even when our proposed preprocessing pipeline is removed, indicating that FORTRESS does not require data augmentation, allowing fast training. FORTRESS produces precise masks with accurate boundaries and effectively handles diverse defect appearances. The obtained results show the effectiveness of incorporating KAN in parameter-scarce implementation.\\
\\
Third, we investigated few-shot learning for defect segmentation. Few-shot methods are especially useful when only a few annotated samples are available. Our prototypical network uses a bidirectional approach. Experiments show that this improves the alignment between support and query features, resulting in richer feature representations. Experiments show that this approach improves performance by about ten percent across all evaluation metrics. Furthermore, our results highlight the benefits of incorporating multiple classes per episode as well as multiple samples per class. This allows the model to learn features that distinguish different classes and capture variations in the visual characteristics of defects. We also extended the prediction head with various attention mechanisms, where self-attention performed best. However, applying our proposed data preprocessing steps to our training data decreased the model’s performance. This is most likely because the augmented data from dynamic label injection or classic data augmentation shares strong similarities with other images of the same class, while more diverse images from the original dataset are more beneficial. These findings confirm that few-shot learning with attention and bidirectional training is effective for structural defect segmentation with very limited data.\\
\\
Each investigated approach targets the data scarcity challenge differently. Data preprocessing improves data diversity and quality by introducing new samples from already existing ones. A larger dataset is efficiently obtained without the need for expensive data collection and annotation. The applied techniques need to be selected and adjusted carefully, otherwise, they can harm the model’s performance. FORTRESS provides a strong baseline model, outperforming other state-of-the-art methods with limited data. Its efficient implementation with only 2.9 million parameters and a computational cost of 1.17 GFLOPs makes it suitable for real-time applications. In contrast, our few-shot model is computationally more expensive and requires 8.5 million parameters and 1.8 GFLOPs. However, it provides more flexibility than FORTRESS when adapting to new classes. While FORTRESS requires full retraining for new classes, our few-shot model can handle these scenarios by simply including them in the support and query sets. This makes few-shot semantic segmentation less efficient but more capable of handling new defect types. It also mitigates issues caused by class imbalance by enforcing equal samples per class during training. By using only a small amount of the available data per episode, it can handle very sparse data scenarios better than FORTRESS. However, few-shot semantic segmentation requires support and query sets during inference, slowing down the process. FORTRESS does not need these additional data, allowing efficient real-time use. Future research could focus on optimizing the few-shot model’s efficiency, especially by exploring different encoder architectures. The encoder currently contributes most of the parameters, and using a more efficient backbone could reduce model size and computational cost without sacrificing performance. Integrating FORTRESS as a few-shot encoder may also combine the strengths of both methods.\\
\\
All three methods demonstrate great potential for data-sparse applications. While data preprocessing enhances existing datasets, FORTRESS provides a powerful baseline model, and few-shot learning extends adaptability to unseen classes. Together, they are complementary solutions for efficient and accurate defect segmentation when data is limited. Preprocessing improves data, FORTRESS maximizes efficiency, and few-shot learning enables generalization when data is very limited or unavailable.




\cleardoublepage
\addcontentsline{toc}{chapter}{Bibliography}
\bibliography{references}

@misc{acsi2023,
    author={GulfSCEI},
    title = {Automated Culvert Surveying and Inspection (ACSI) Systems},
    howpublished = {https://gulfscei.cs.uno.edu/research/acsi},
    note = {Online; accessed September 05, 2024},
    year={2023}
}

@article{alshawi23a,
    author = {Alshawi, Rasha and Hoque, Md Tamjidul and Flanagin, Maik C.},
    title = {A Depth-Wise Separable U-Net Architecture with Multiscale Filters to Detect Sinkholes},
    journal = {Remote Sensing},
    volume = {15},
    year = {2023},
    number = {5}
}

@article{alshawi23b,
    title={Dual Attention U-Net with Feature Infusion: Pushing the Boundaries of Multiclass Defect Segmentation}, 
    author={Alshawi, Rasha and Hoque, Md Tamjidul and Ferdaus, Md Meftahul and Abdelguerfi, Mahdi and Niles, Kendall and Prathak, Ken and Tom, Joe and Klein, Jordan and Mousa, Murtada and Lopez, Johny Javier},
    journal = {arXiv CoRR},
    volume = {abs/2312.14053},
    year = {2023},
    url = {https://arxiv.org/abs/2312.14053}
}

@article{alshawi24a,
    title={SHARP-Net: A Refined Pyramid Network for Deficiency Segmentation in Culverts and Sewer Pipes}, 
    author={Alshawi, Rasha and Ferdaus, Md Meftahul and Hoque, Md Tamjidul and Niles, Kendall and Pathak, Ken and Sloan, Steve and Abdelguerfi, Mahdi},
    journal = {arXiv CoRR},
    volume = {abs/2408.08879},
    year = {2024},
    url = {https://arxiv.org/abs/2408.08879}
}

@article{alshawi24b,
    title={Imbalance-Aware Culvert-Sewer Defect Segmentation Using an Enhanced Feature Pyramid Network}, 
    author={Alshawi, Rasha and Ferdaus, Md Meftahul and Abdelguerfi, Mahdi and Niles, Kendall and Pathak, Ken and Sloan, Steve},
    journal = {arXiv CoRR},
    volume = {abs/2408.10181},
    year = {2024},
    url = {https://arxiv.org/abs/2408.10181}
}

@inproceedings{bali15,
  author={Bali, Akanksha and Singh, Shailendra Narayan},
  booktitle={Proceedings of the International Conference on Advanced Computing and Communication Technologies}, 
  title={A Review on the Strategies and Techniques of Image Segmentation}, 
  year={2015},
  pages={113--120}
}

@inproceedings{kim18,
    author = {Kim, Seung-Wook and Kook, Hyong-Keun and Sun, Jee-Young and Kang, Mun-Cheon and Ko, Sung-Jea},
    title = {Parallel Feature Pyramid Network for Object Detection},
    booktitle = {Proceedings of the European Conference on Computer Vision},
    year = {2018}
}

@article{su22,
  title={Research on a U-Net bridge crack identification and feature-calculation methods based on a CBAM attention mechanism},
  author={Su, Huifeng and Wang, Xiang and Han, Tao and Wang, Ziyi and Zhao, Zhongxiao and Zhang, Pengfei},
  journal={Buildings},
  volume={12},
  number={10},
  year={2022},
  publisher={MDPI}
}

@article{tong21,
  title={ASCU-Net: attention gate, spatial and channel attention u-net for skin lesion segmentation},
  author={Tong, Xiaozhong and Wei, Junyu and Sun, Bei and Su, Shaojing and Zuo, Zhen and Wu, Peng},
  journal={Diagnostics},
  volume={11},
  number={3},
  year={2021},
  publisher={MDPI}
}

@manual{nasscoPACP,
  title = {Pipeline Assessment Certification Program Reference Manual},
  author = {National Association of Sewer Service Companies (NASSCO)},
  year = {2025},
  edition = {Imperial Version 8.1}
}

@article{shorten19,
  title={A survey on Image Data Augmentation for Deep Learning},
  author={Shorten, Connor and Khoshgoftaar, Taghi M},
  journal={Journal of Big Data},
  volume={6},
  number={1},
  pages={1--60},
  year={2019},
  publisher={Springer}
}

@inproceedings{caruso24,
  title={{D}ynamic {L}abel {I}njection for {I}mbalanced {I}ndustrial {D}efect {S}egmentation},
  author={Caruso, Emanuele and Pelosin, Francesco and Simoni, Alessandro and Boschetti, Marco},
  booktitle={Proceedings of the European Conference on Computer Vision Workshops},
  year={2024}
}

@article{catalano23,
  author = {Catalano, Nico and Matteucci, Matteo},
  title  = {{F}ew {S}hot {S}emantic {S}egmentation: a review of methodologies, benchmarks, and open challenges},
  journal = {arXiv CoRR},
  volume = {abs/2304.05832},
  year = {2023},
  url = {http://arxiv.org/abs/2304.05832}
}

@inbook{goodfellow16, 
 title = {{D}eep {L}earning}, 
 chapter = {{R}egularization for {D}eep {L}earning}, 
 author = {Goodfellow, Ian  and Bengio, Yoshua  and Courville, Aaron}, 
 publisher={MIT Press},
 pages = {224--270}, 
 year = {2016} 
}

@inproceedings{simard03,
  title={{B}est {P}ractices for {C}onvolutional {N}eural {N}etworks {A}pplied to {V}isual {D}ocument {A}nalysis},
  author={Simard, Patrice Y and Steinkraus, David and Platt, John C and others},
  booktitle={Proceedings of the International Conference on Document Analysis and Recognition},
  year={2003}
}

@article{wang20b,
  author={Wang, Ke and Fang, Bin and Qian, Jiye and Yang, Su and Zhou, Xin and Zhou, Jie},
  journal={IEEE Access}, 
  title={{P}erspective {T}ransformation {D}ata {A}ugmentation for {O}bject {D}etection}, 
  volume={8},
  number={},
  pages={4935-4943},
  year={2020},
}

@article{mazhar21,
  title={{R}andom {S}hadows and {H}ighlights: A new data augmentation method for extreme lighting conditions},
  author={Mazhar, Osama and Kober, Jens},
  journal = {arXiv CoRR},
  volume = {abs/2101.05361},
  year = {2021},
  url = {http://arxiv.org/abs/2101.05361}
}

@inproceedings{ronneberger2015u,
  title={U-Net: Convolutional Networks for Biomedical Image Segmentation},
  author={Ronneberger, Olaf and Fischer, Philipp and Brox, Thomas},
  booktitle={International Conference on Medical Image Computing and Computer-Assisted Intervention},
  pages={234--241},
  year={2015},
  organization={Springer}
}

@inproceedings{lin2017feature,
  title={Feature Pyramid Networks for Object Detection},
  author={Lin, Tsung-Yi and Doll{\'a}r, Piotr and Girshick, Ross and He, Kaiming and Hariharan, Bharath and Belongie, Serge},
  booktitle={Proceedings of the IEEE Conference on Computer Vision and Pattern Recognition},
  pages={2117--2125},
  year={2017}
}

@inproceedings{oktay2018attention,
  title={Attention U-Net: Learning Where to Look for the Pancreas},
  author={Oktay, Ozan and Schlemper, Jo and Folgoc, Loic Le and Lee, Matthew and Heinrich, Mattias and Misawa, Kazunari and Mori, Kensaku and McDonagh, Steven and Hammerla, Nils Y and Kainz, Bernhard and others},
  booktitle={International Conference on Medical Imaging with Deep Learning},
  pages={1--10},
  year={2018}
}

@inproceedings{zhou2018unet++,
  title={Unet++: A nested u-net architecture for medical image segmentation},
  author={Zhou, Zongwei and Rahman Siddiquee, Md Mahfuzur and Tajbakhsh, Nima and Liang, Jianming},
  booktitle={Deep learning in medical image analysis and multimodal learning for clinical decision support: 4th international workshop, DLMIA 2018, and 8th international workshop, ML-CDS 2018, held in conjunction with MICCAI 2018, Granada, Spain, September 20, 2018, proceedings 4},
  pages={3--11},
  year={2018},
  organization={Springer}
}

@inproceedings{tan2020efficientdet,
  title={EfficientDet: Scalable and Efficient Object Detection},
  author={Tan, Mingxing and Pang, Ruoming and Le, Quoc V},
  booktitle={Proceedings of the IEEE/CVF Conference on Computer Vision and Pattern Recognition},
  pages={10781--10790},
  year={2020}
}

@inproceedings{guo2021sa,
  title={Sa-unet: Spatial attention u-net for retinal vessel segmentation},
  author={Guo, Changlu and Szemenyei, M{\'a}rton and Yi, Yugen and Wang, Wenle and Chen, Buer and Fan, Changqi},
  booktitle={2020 25th international conference on pattern recognition (ICPR)},
  pages={1236--1242},
  year={2021},
  organization={IEEE}
}

@inproceedings{huang2020unet,
  title={Unet 3+: A full-scale connected unet for medical image segmentation},
  author={Huang, Huimin and Lin, Lanfen and Tong, Ruofeng and Hu, Hongjie and Zhang, Qiaowei and Iwamoto, Yutaro and Han, Xianhua and Chen, Yen-Wei and Wu, Jian},
  booktitle={ICASSP 2020-2020 IEEE international conference on acoustics, speech and signal processing (ICASSP)},
  pages={1055--1059},
  year={2020},
  organization={IEEE}
}

@inproceedings{valanarasu2022unext,
  title={Unext: Mlp-based rapid medical image segmentation network},
  author={Valanarasu, Jeya Maria Jose and Patel, Vishal M},
  booktitle={International conference on medical image computing and computer-assisted intervention},
  pages={23--33},
  year={2022},
  organization={Springer}
}

@inproceedings{ruan2023ege,
  title={Ege-unet: an efficient group enhanced unet for skin lesion segmentation},
  author={Ruan, Jiacheng and Xie, Mingye and Gao, Jingsheng and Liu, Ting and Fu, Yuzhuo},
  booktitle={International conference on medical image computing and computer-assisted intervention},
  pages={481--490},
  year={2023},
  organization={Springer}
}

@inproceedings{liu2024rolling,
  title={Rolling-unet: Revitalizing mlp's ability to efficiently extract long-distance dependencies for medical image segmentation},
  author={Liu, Yutong and Zhu, Haijiang and Liu, Mengting and Yu, Huaiyuan and Chen, Zihan and Gao, Jie},
  booktitle={Proceedings of the AAAI Conference on Artificial Intelligence},
  volume={38},
  number={4},
  pages={3819--3827},
  year={2024}
}

@inproceedings{ghahremani2024h,
  title={H-vit: A hierarchical vision transformer for deformable image registration},
  author={Ghahremani, Morteza and Khateri, Mohammad and Jian, Bailiang and Wiestler, Benedikt and Adeli, Ehsan and Wachinger, Christian},
  booktitle={Proceedings of the IEEE/CVF Conference on Computer Vision and Pattern Recognition},
  pages={11513--11523},
  year={2024}
}

@article{cao2021swin,
  title={Swin-Unet: Unet-like Pure Transformer for Medical Image Segmentation},
  author={Cao, Hu and Wang, Yueyue and Chen, Joy and Jiang, Dongsheng and Zhang, Xiaopeng and Tian, Qi and Wang, Manning},
  journal={arXiv preprint arXiv:2105.05537},
  year={2021}
}

@article{perera2024mobileunetr,
  title={MobileUNETR: A Lightweight End-To-End Hybrid Vision Transformer For Efficient Medical Image Segmentation},
  author={Perera, Shehan and Erzurumlu, Yunus and Gulati, Deepak and Yilmaz, Alper},
  journal={arXiv preprint arXiv:2409.03062},
  year={2024}
}

@article{xie2021segformer,
  title={SegFormer: Simple and efficient design for semantic segmentation with transformers},
  author={Xie, Enze and Wang, Wenhai and Yu, Zhiding and Anandkumar, Anima and Alvarez, Jose M and Luo, Ping},
  journal={Advances in neural information processing systems},
  volume={34},
  pages={12077--12090},
  year={2021}
}

@inproceedings{hatamizadeh2024fastervit,
  title={FasterViT: Fast Vision Transformers with Hierarchical Attention},
  author={Hatamizadeh, Ali and Heinrich, Greg and Yin, Hongxu and Tao, Andrew and Alvarez, Jose M. and Kautz, Jan and Molchanov, Pavlo},
  booktitle={International Conference on Learning Representations},
  year={2024},
  url={https://arxiv.org/abs/2306.06189}
}

@article{li2024ukan,
  title={U-KAN Makes Strong Backbone for Medical Image Segmentation and Generation},
  author={Li, Chenxin and Liu, Xinyu and Li, Wuyang and Wang, Cheng and Liu, Hengyu and Yuan, Yixuan},
  journal={arXiv preprint arXiv:2406.02918},
  year={2024}
}

@inproceedings{dwibedi17,
  title={Cut, Paste and Learn: Surprisingly Easy Synthesis for Instance Detection},
  author={Dwibedi, Debidatta and Misra, Ishan and Hebert, Martial},
  booktitle={Proceedings of the IEEE International Conference on Computer Vision},
  pages={1301--1310},
  year={2017}
}

@inproceedings{dvornik18,
  title={ Modeling Visual Context is Key to Augmenting Object Detection Datasets},
  author={Dvornik, Nikita and Mairal, Julien and Schmid, Cordelia},
  booktitle={Proceedings of the European Conference on Computer Vision},
  pages={364--380},
  year={2018}
}

@inproceedings{ghiasi21,
  title={{S}imple {C}opy-{P}aste is a {S}trong {D}ata {A}ugmentation {M}ethod for {I}nstance {S}egmentation},
  author={Ghiasi, Golnaz and Cui, Yin and Srinivas, Aravind and Qian, Rui and Lin, Tsung-Yi and Cubuk, Ekin D and Le, Quoc V and Zoph, Barret},
  booktitle={Proceedings of the IEEE/CVF Conference on Computer Vision and Pattern Recognition},
  pages={2918--2928},
  year={2021}
}

@inproceedings{gupta16,
  title={Synthetic Data for Text Localisation in Natural Images},
  author={Gupta, Ankush and Vedaldi, Andrea and Zisserman, Andrew},
  booktitle={Proceedings of the IEEE Conference on Computer Vision and Pattern Recognition},
  pages={2315--2324},
  year={2016}
}

@article{karsch11,
  title={Rendering Synthetic Objects into Legacy Photograph},
  author={Karsch, Kevin and Hedau, Varsha and Forsyth, David and Hoiem, Derek},
  journal={ACM Transactions on graphics},
  volume={30},
  number={6},
  pages={1--12},
  year={2011}
}

@inproceedings{fang19,
  title={InstaBoost: Boosting Instance Segmentation via Probability Map Guided Copy-Pasting},
  author={Fang, Hao-Shu and Sun, Jianhua and Wang, Runzhong and Gou, Minghao and Li, Yong-Lu and Lu, Cewu},
  booktitle={Proceedings of the IEEE/CVF International Conference on Computer Vision},
  pages={682--691},
  year={2019}
}

@inproceedings{zhang18,
  title     = {mixup: Beyond Empirical Risk Minimization},
  author    = {Zhang, Hongyi and Cisse, Moustapha and Dauphin, Yann N. and Lopez-Paz, David},
  booktitle = {Proceedings of the International Conference on Learning Representations},
  year      = {2018}
}

@inproceedings{yun19,
  title={CutMix: Regularization Strategy to Train Strong Classifiers with Localizable Features},
  author={Yun, Sangdoo and Han, Dongyoon and Oh, Seong Joon and Chun, Sanghyuk and Choe, Junsuk and Yoo, Youngjoon},
  booktitle={Proceedings of the IEEE/CVF International Conference on Computer Vision},
  pages={6023--6032},
  year={2019}
}

@article{devries17,
  title={Improved Regularization of Convolutional Neural Networks with Cutout},
  author={DeVries, Terrance and Taylor, Graham W},
  journal = {arXiv CoRR},
  volume = {abs/1708.04552},
  year={2017},
  url = {http://arxiv.org/abs/1708.04552}
}

@article{guo18,
  title={A review of semantic segmentation using deep neural networks},
  author={Guo, Yanming and Liu, Yu and Georgiou, Theodoros and Lew, Michael S},
  journal={International Journal of Multimedia Information Retrieval},
  volume={7},
  pages={87--93},
  year={2018}
}

@book{sonka13,
  title={Image Processing, Analysis and Machine Vision},
  author={Sonka, Milan and Hlavac, Vaclav and Boyle, Roger},
  edition={4},
  year={2014},
  publisher={Cengage Learning},
  address={Boston, MA}
}

@book{butler18,
  title={Urban Drainage},
  author={Butler, David and Digman, Christopher and Makropoulos, Christos and Davies, John W},
  year={2018},
  edition={4},
  publisher={Crc Press}
}

@article{leevy18,
  title={A survey on addressing high-class imbalance in big data},
  author={Leevy, Joffrey L and Khoshgoftaar, Taghi M and Bauder, Richard A and Seliya, Naeem},
  journal={Journal of Big Data},
  volume={5},
  number={1},
  pages={1--30},
  year={2018},
  publisher={Springer}
}

@inproceedings{davis06,
  title = {{T}he {R}elationship {B}etween {P}recision-{R}ecall and {ROC} {C}urves},
  author = {Davis, Jesse and Goadrich, Mark},
  booktitle = {Proceedings of the International Conference on Machine Learning},
  year = {2006}
}

@article{vujovic21,
  title={Classification Model Evaluation Metrics},
  author={Vujovi{\'c}, {\v{Z}eljko D.}},
  journal={International Journal of Advanced Computer Science and Applications},
  volume={12},
  number={6},
  pages={599--606},
  year={2021}
}

@article{rainio24,
  title={Evaluation metrics and statistical tests for machine learning},
  author={Rainio, Oona and Teuho, Jarmo and Kl{\'e}n, Riku},
  journal={Scientific Reports},
  volume={14},
  number={1},
  pages={6086},
  year={2024}
}

@article{garcia17,
  title={A Review on Deep Learning Techniques Applied to Semantic Segmentation},
  author={Garcia-Garcia, Alberto and Orts-Escolano, Sergio and Oprea, Sergiu and Villena-Martinez, Victor and Garcia-Rodriguez, Jose},
  journal = {arXiv CoRR},
  volume = {abs/1704.06857},
  year={2017},
  url = {http://arxiv.org/abs/1704.06857}
}

@article{grandini20,
  title={Metrics for Multi-Class Classification: an Overview},
  author={Grandini, Margherita and Bagli, Enrico and Visani, Giorgio},
  journal = {arXiv CoRR},
  volume = {abs/2008.05756},
  year={2020},
  url = {http://arxiv.org/abs/2008.05756}
}

@inproceedings{girshick15,
  title={Fast R-CNN},
  author={Girshick, Ross},
  booktitle={Proceedings of the IEEE International Conference on Computer Vision},
  pages={1440--1448},
  year={2015}
}

@article{wu15,
  title={Deep Image: Scaling up Image Recognition},
  author={Wu, Ren and Yan, Shengen and Shan, Yi and Dang, Qingqing and Sun, Gang},
  journal = {arXiv CoRR},
  volume = {abs/1501.02876},
  year={2015},
  url = {http://arxiv.org/abs/1501.02876}
}

@article{krizhevsky12,
  title={ImageNet Classification with Deep Convolutional Neural Networks},
  author={Krizhevsky, Alex and Sutskever, Ilya and Hinton, Geoffrey E},
  journal={Advances in Neural Information Processing Systems},
  volume={25},
  year={2012}
}

@inproceedings{moreno18,
  title={Forward Noise Adjustment Scheme for Data Augmentation},
  author={Moreno-Barea, Francisco J. and Strazzera, Fiammetta and Jerez, Jos{\'e} M. and Urda, Daniel and Franco, Leonardo},
  booktitle={Proceedings of the IEEE Symposium Series on Computational Intelligence},
  year={2018},
}

@article{akbiyik23,
  title={Data Augmentation in Training CNNs: Injecting Noise to Images},
  author={Akbiyik, M Eren},
  journal = {arXiv CoRR},
  volume = {abs/2307.06855},
  year={2023},
  url = {http://arxiv.org/abs/2307.06855}
}

@article{lopes19,
  title={Improving Robustness Without Sacrificing Accuracy with Patch Gaussian Augmentation},
  author={Lopes, Raphael Gontijo and Yin, Dong and Poole, Ben and Gilmer, Justin and Cubuk, Ekin D},
  journal = {arXiv CoRR},
  volume = {abs/1906.02611},
  year={2019},
  url = {http://arxiv.org/abs/1906.02611}
}

@inproceedings{chatfield14,
	title = {Return of the Devil in the Details: Delving Deep into Convolutional Nets},
	author = {Chatfield, Ken and Simonyan, Karen and Vedaldi, Andrea and Zisserman, Andrew},
	year = {2014},
	booktitle = {Proceedings of the British Machine Vision Conference}
}

@inproceedings{jurio10,
  title={A Comparison Study of Different Color Spaces in Clustering Based Image Segmentation},
  author={Jurio, Aranzazu and Pagola, Miguel and Galar, Mikel and Lopez-Molina, Carlos and Paternain, Daniel},
  booktitle={Proceedings of the International Conference on Information Processing and Management of Uncertainty in Knowledge-Based Systems},
  year={2010}
}

@article{inoue18,
  title={Data Augmentation by Pairing Samples for Images Classification},
  author={Inoue, Hiroshi},
  journal = {arXiv CoRR},
  volume = {abs/1801.02929},
  year={2018},
  url = {http://arxiv.org/abs/1801.02929}
}

@article{takahashi19,
  title={Data Augmentation Using Random Image Cropping and Patching for Deep CNNs},
  author={Takahashi, Ryo and Matsubara, Takashi and Uehara, Kuniaki},
  journal={IEEE Transactions on Circuits and Systems for Video Technology},
  volume={30},
  number={9},
  pages={2917--2931},
  year={2019}
}

@inproceedings{summers19,
  title={Improved mixed-example data augmentation},
  author={Summers, Cecilia and Dinneen, Michael J},
  booktitle={Proceedings of the IEEE Winter Conference on Applications of Computer Vision},
  year={2019}
}

@inproceedings{taylor18,
  title={Improving Deep Learning using Generic Data Augmentation},
  author={Taylor, Luke and Nitschke, Geoff},
  booktitle={Proceedings of the IEEE Symposium Series on Computational Intelligence},
  year={2018}
}

@inproceedings{shaban17,
  title={One-Shot Learning for Semantic Segmentation},
  author={Shaban, Amirreza and Bansal, Shray and Liu, Zhen and Essa, Irfan and Boots, Byron},
  year={2017},
  booktitle={Proceedings of the British Machine Vision Conference},
}

@article{snell17,
  author = {Snell, Jake and Swersky, Kevin and Zemel, Richard},
  booktitle = {Advances in Neural Information Processing Systems},
  title = {Prototypical Networks for Few-shot Learning},
  volume = {30},
  year = {2017}
}

@inproceedings{dong18,
  title={Few-Shot Semantic Segmentation with Prototype Learning},
  author={Dong, Nanqing and Xing, Eric P},
  booktitle={Proceedings of the British Machine Vision Conference},
  year={2018}
}

@inproceedings{wang19,
author = {Wang, Kaixin and Liew, Jun Hao and Zou, Yingtian and Zhou, Daquan and Feng, Jiashi},
title = {PANet: Few-Shot Image Semantic Segmentation With Prototype Alignment},
booktitle = {Proceedings of the IEEE International Conference on Computer Vision},
year = {2019}
}

@article{goodfellow14,
  title={Generative Adversarial Nets},
  author={Goodfellow, Ian J and Pouget-Abadie, Jean and Mirza, Mehdi and Xu, Bing and Warde-Farley, David and Ozair, Sherjil and Courville, Aaron and Bengio, Yoshua},
  journal={Advances in Neural Information Processing Systems},
  volume={27},
  year={2014}
}

@inproceedings{karras19,
  title={A Style-Based Generator Architecture for Generative Adversarial Networks},
  author={Karras, Tero and Laine, Samuli and Aila, Timo},
  booktitle={Proceedings of the IEEE/CVF Conference on Computer Vision and Pattern Recognition},
  pages={4401--4410},
  year={2019}
}

@inproceedings{saha22,
  title={GanOrCon: Are Generative Models Useful for Few-Shot Segmentation?},
  author={Saha, Oindrila and Cheng, Zezhou and Maji, Subhransu},
  booktitle={Proceedings of the IEEE/CVF Conference on Computer Vision and Pattern Recognition},
  pages={9991--10000},
  year={2022}
}

@inproceedings{wang21,
  title={Variational Prototype Inference for Few-Shot Semantic Segmentation },
  author={Wang, Haochen and Yang, Yandan and Cao, Xianbin and Zhen, Xiantong and Snoek, Cees and Shao, Ling},
  booktitle={Proceedings of the IEEE/CVF Winter Conference on Applications of Computer Vision},
  pages={525--534},
  year={2021}
}

@inproceedings{tritrong21,
  title={Repurposing GANs for One-Shot Semantic Part Segmentation },
  author={Tritrong, Nontawat and Rewatbowornwong, Pitchaporn and Suwajanakorn, Supasorn},
  booktitle={Proceedings of the IEEE/CVF Conference on Computer Vision and Pattern Recognition},
  pages={4475--4485},
  year={2021}
}

@inproceedings{liu20part-aware,
  title={Part-Aware Prototype Network for Few-Shot Semantic Segmentation},
  author={Liu, Yongfei and Zhang, Xiangyi and Zhang, Songyang and He, Xuming},
  booktitle={Proceedings of the European Conference on Computer Vision},
  pages={142--158},
  year={2020}
}

@inproceedings{zhang19,
  title={CANet: Class-Agnostic Segmentation Networks With Iterative Refinement and Attentive Few-Shot Learning},
  author={Zhang, Chi and Lin, Guosheng and Liu, Fayao and Yao, Rui and Shen, Chunhua},
  booktitle={Proceedings of the IEEE/CVF Conference on Computer Vision and Pattern Recognition},
  pages={5217--5226},
  year={2019}
}

@article{zhang20,
  title={SG-One: Similarity Guidance Network for One-Shot Semantic Segmentation},
  author={Zhang, Xiaolin and Wei, Yunchao and Yang, Yi and Huang, Thomas S},
  journal={IEEE Transactions on Cybernetics},
  volume={50},
  number={9},
  pages={3855--3865},
  year={2020}
}

@article{liu20prototype,
  title={Prototype refinement network for few-shot segmentation},
  author={Liu, Jinlu and Qin, Yongqiang},
  journal = {arXiv CoRR},
  volume = {abs/2002.03579},
  year={2020},
  url = {http://arxiv.org/abs/2002.03579}
}

@inproceedings{dosovitskiy21,
  title     = {An Image is Worth 16x16 Words: Transformers for Image Recognition at Scale},
  author    = {Dosovitskiy, Alexey and Beyer, Lucas and Kolesnikov, Alexander and Weissenborn, Dirk and Zhai, Xiaohua and Unterthiner, Thomas and Dehghani, Mostafa and Minderer, Matthias and Heigold, Georg and Gelly, Sylvain and Uszkoreit, Jakob and Houlsby, Neil},
  booktitle = {Proceedings of the International Conference on Learning Representations},
  year      = {2021}
}

@article{vaswani17,
  title={Attention Is All You Need},
  author={Vaswani, Ashish and Shazeer, Noam and Parmar, Niki and Uszkoreit, Jakob and Jones, Llion and Gomez, Aidan N and Kaiser, {\L}ukasz and Polosukhin, Illia},
  journal={Advances in Neural Information Processing Systems},
  volume={30},
  year={2017}
}

@inproceedings{liu21swin,
  title={Swin Transformer: Hierarchical Vision Transformer Using Shifted Windows},
  author={Liu, Ze and Lin, Yutong and Cao, Yue and Hu, Han and Wei, Yixuan and Zhang, Zheng and Lin, Stephen and Guo, Baining},
  booktitle={Proceedings of the IEEE/CVF International Conference on Computer Vision},
  pages={10012--10022},
  year={2021}
}

@article{minaee22,
  title={Image Segmentation Using Deep Learning: A Survey},
  author={Minaee, Shervin and Boykov, Yuri and Porikli, Fatih and Plaza, Antonio and Kehtarnavaz, Nasser and Terzopoulos, Demetri},
  journal={IEEE Transactions on Pattern Analysis and Machine Intelligence}, 
  volume={44},
  number={7},
  pages={3523--3542},
  year={2022}
}

@article{otsu1975,
  title={A Threshold Selection Method from Gray-Level Histograms}, 
  author={Otsu, Nobuyuki},
  journal={IEEE Transactions on Systems, Man, and Cybernetics}, 
  volume={9},
  number={1},
  pages={62-66},
  year={1979}
}

@article{dhanachandra15,
  title={Image Segmentation Using K-means Clustering Algorithm and Subtractive Clustering Algorithm},
  author={Dhanachandra, Nameirakpam and Manglem, Khumanthem and Chanu, Yambem Jina},
  journal={Procedia Computer Science},
  volume={54},
  pages={764--771},
  year={2015}
}

@article{nock04,
  title={Statistical Region Merging},
  author={Nock, Richard and Nielsen, Frank},
  journal={IEEE Transactions on Pattern Analysis and Machine Intelligence},
  volume={26},
  number={11},
  pages={1452--1458},
  year={2004}
}

@article{lecun1998,
  title={Gradient-based learning applied to document recognition},
  author={LeCun, Yann and Bottou, L{\'e}on and Bengio, Yoshua and Haffner, Patrick},
  journal={Proceedings of the IEEE},
  volume={86},
  number={11},
  pages={2278--2324},
  year={1998}
}

@inbook{goodfellow16cnn, 
 title = {{D}eep {L}earning}, 
 chapter = {{C}onvolutional {N}etworks}, 
 author = {Goodfellow, Ian  and Bengio, Yoshua  and Courville, Aaron}, 
 publisher={MIT Press},
 pages = {224--270}, 
 year = {2016} 
}

@inproceedings{he16,
  title={Deep Residual Learning for Image Recognition},
  author={He, Kaiming and Zhang, Xiangyu and Ren, Shaoqing and Sun, Jian},
  booktitle={Proceedings of the IEEE Conference on Computer Vision and Pattern Recognition},
  pages={770--778},
  year={2016}
}

@inproceedings{long15,
  title={Fully Convolutional Networks for Semantic Segmentation},
  author={Long, Jonathan and Shelhamer, Evan and Darrell, Trevor},
  booktitle={Proceedings of the IEEE Conference on Computer Vision and Pattern Recognition},
  pages={3431--3440},
  year={2015}
}

@article{farabet13,
  title={Learning Hierarchical Features for Scene Labeling},
  author={Farabet, Clement and Couprie, Camille and Najman, Laurent and LeCun, Yann},
  journal={IEEE Transactions on Pattern Analysis and Machine Intelligence},
  volume={35},
  number={8},
  pages={1915--1929},
  year={2013}
}

@inproceedings{pinheiro14,
  title = {Recurrent Convolutional Neural Networks for Scene Labeling},
  author = {Pinheiro, Pedro and Collobert, Ronan},
  booktitle = {Proceedings of the International Conference on Machine Learning},
  pages = {82--90},
  year = {2014}
}

@inproceedings{hariharan14,
  author = {Hariharan, Bharath and Arbel{\'a}ez, Pablo and Girshick, Ross and Malik, Jitendra},
  title = {Simultaneous Detection and Segmentation},
  booktitle = {Proceedings of the European Conference on Computer Vision},
  year = {2014},
  pages = {297--312}
}

@InProceedings{gupta14,
author = {Gupta, Saurabh and Girshick, Ross and Arbel{\'a}ez, Pablo and Malik, Jitendra},
title = {Learning Rich Features from RGB-D Images for Object Detection and Segmentation},
booktitle = {Proceedings of the European Conference on Computer Vision},
year = {2014},
pages = {345--360}
}

@inproceedings{ganin14,
  title={$N^4$-fields: Neural network nearest neighbor fields for image transforms},
  author={Ganin, Yaroslav and Lempitsky, Victor},
  booktitle={Proceedings of the Asian Conference on Computer Vision},
  pages={536--551},
  year={2014}
}

@article{lecun89,
  title = {{B}ackpropagation {A}pplied to {H}andwritten {Z}ip {C}ode {R}ecognition}, 
  author = {LeCun, Yann and Boser, Bernhard and Denker, John S. and Henderson, Donnie and Howard, Richard E. and Hubbard, Wayne and Jackel, Lawrence D.},
  journal = {Neural Computation}, 
  volume={1},
  number={4},
  pages={541--551},
  year={1989}
}

@inproceedings{deng09,
  title={ImageNet: A large-scale hierarchical image database},
  author={Deng, Jia and Dong, Wei and Socher, Richard and Li, Li-Jia and Li, Kai and Fei-Fei, Li},
  booktitle={Proceedings of the IEEE Conference on Computer Vision and Pattern Recognition},
  pages={248--255},
  year={2009}
}

@article{gao18,
  title={An End-to-End Neural Network for Road Extraction From Remote Sensing Imagery by Multiple Feature Pyramid Network},
  author={Gao, Xun and Sun, Xian and Zhang, Yi and Yan, Menglong and Xu, Guangluan and Sun, Hao and Jiao, Jiao and Fu, Kun},
  journal={IEEE Transactions on Pattern Analysis and Machine Intelligence},
  volume={6},
  number={},
  pages={39401-39414},
  year={2018}
}

@inproceedings{zheng21,
  title={Rethinking Semantic Segmentation From a Sequence-to-Sequence Perspective With Transformers},
  author={Zheng, Sixiao and Lu, Jiachen and Zhao, Hengshuang and Zhu, Xiatian and Luo, Zekun and Wang, Yabiao and Fu, Yanwei and Feng, Jianfeng and Xiang, Tao and Torr, Philip HS and Zhang, Li},
  booktitle={Proceedings of the IEEE/CVF Conference on Computer Vision and Pattern Recognition},
  pages={6881--6890},
  year={2021}
}

@article{oktay18,
  title={Attention U-Net: Learning Where to Look for the Pancreas},
  author={Oktay, Ozan and Schlemper, Jo and Folgoc, Loic Le and Lee, Matthew and Heinrich, Mattias and Misawa, Kazunari and Mori, Kensaku and McDonagh, Steven and Hammerla, Nils Y and Kainz, Bernhard and Glocker, Ben and Rueckert, Daniel},
  journal = {arXiv CoRR},
  volume = {abs/1804.03999},
  year={2018},
  url = {http://arxiv.org/abs/1804.03999}
}

@article{gholamalinezhad20,
  title={Pooling Methods in Deep Neural Networks, a Review},
  author={Gholamalinezhad, Hossein and Khosravi, Hossein},
  journal = {arXiv CoRR},
  volume = {abs/2009.07485},
  year={2020},
  url = {http://arxiv.org/abs/2009.07485}
}

@inproceedings{szegedy17,
  title={Inception-v4, Inception-ResNet and the Impact of Residual Connections on Learning },
  author={Szegedy, Christian and Ioffe, Sergey and Vanhoucke, Vincent and Alemi, Alexander},
  booktitle={Proceedings of the AAAI Conference on Artificial Intelligence},
  volume={31},
  number={1},
  year={2017}
}

@inproceedings{hu18,
  title={Squeeze-and-Excitation Networks},
  author={Hu, Jie and Shen, Li and Sun, Gang},
  booktitle={Proceedings of the IEEE Conference on Computer Vision and Pattern Recognition},
  pages={7132--7141},
  year={2018}
}

@article{howard17,
  title={MobileNets: Efficient Convolutional Neural Networks for Mobile Vision Applications},
  author={Howard, Andrew G and Zhu, Menglong and Chen, Bo and Kalenichenko, Dmitry and Wang, Weijun and Weyand, Tobias and Andreetto, Marco and Adam, Hartwig},
  journal = {arXiv CoRR},
  volume = {abs/1704.04861},
  year={2017},
  url = {http://arxiv.org/abs/1704.04861}
}

@article{liu24,
  title = {KAN: Kolmogorov-Arnold Networks},
  author = {Liu, Ziming and Wang, Yixuan and Vaidya, Sachin and Ruehle, Fabian and Halverson, James and Soljacic, Marin and Hou, Thomas Y and Tegmark, Max},
  journal = {arXiv CoRR},
  volume = {abs/2404.19756},
  year={2024},
  url = {http://arxiv.org/abs/2404.19756}
}

@inproceedings{sabet22,
  title={An Automated Visual Defect Segmentation for Flat Steel Surface Using Deep Neural Networks},
  author={Sabet, D. N. and Zarifi, M. and Khoramdel, Javad and Borhani, Y. and Najafi, E.},
  booktitle={Proceedings of the International Conference on Computer and Knowledge Engineering},
  pages={423--427},
  year={2022}
}

@inproceedings{brateanu24,
  title={Kolmogorov-Arnold Networks in Transformer Attention for Low-Light Image Enhancement},
  author={Brateanu, A. and Balmez, R.},
  booktitle={Proceedings of the International Symposium on Electronics and Telecommunications},
  pages={1--4},
  year={2024}
}

@article{mahmoudi25,
  title={Addressing class imbalance in micro-CT image segmentation: A modified U-Net model with pixel-level class weighting},
  author={Mahmoudi, Shahin and Asghari, Omid and Boisvert, Jeff},
  journal={Computers \& Geosciences},
  year={2025},
  volume = {196},
  pages = {105853}
}

@article{oreshkin18,
  title={Tadam: Task dependent adaptive metric for improved few-shot learning},
  author={Oreshkin, Boris and Rodr{\'\i}guez L{\'o}pez, Pau and Lacoste, Alexandre},
  journal={Advances in Neural Information Processing Systems},
  volume={31},
  year={2018}
}

@article{galassi20,
  title={Attention in Natural Language Processing},
  author={Galassi, Andrea and Lippi, Marco and Torroni, Paolo},
  journal={IEEE Transactions on Neural Networks and Learning Systems},
  volume={32},
  number={10},
  pages={4291--4308},
  year={2020}
}

@article{khan22,
  title={Transformers in Vision: A Survey},
  author={Khan, Salman and Naseer, Muzammal and Hayat, Munawar and Zamir, Syed Waqas and Khan, Fahad Shahbaz and Shah, Mubarak},
  journal={ACM Computing Surveys},
  volume={54},
  number={10s},
  pages={1--41},
  year={2022}
}

@article{zhang19b,
  title={Deep Learning Based Recommender System: A Survey and New Perspectives},
  author={Zhang, Shuai and Yao, Lina and Sun, Aixin and Tay, Yi},
  journal={ACM Computing Surveys},
  volume={52},
  number={1},
  pages={1--38},
  year={2019}
}

@article{chaudhari21,
  title={An Attentive Survey of Attention Models},
  author={Chaudhari, Sneha and Mithal, Varun and Polatkan, Gungor and Ramanath, Rohan},
  journal={ACM Transactions on Intelligent Systems and Technology},
  volume={12},
  number={5},
  pages={1--32},
  year={2021}
}

@article{zhou21,
  title={ELSA: Enhanced Local Self-Attention for Vision Transformer},
  author={Zhou, Jingkai and Wang, Pichao and Wang, Fan and Liu, Qiong and Li, Hao and Jin, Rong},
  journal = {arXiv CoRR},
  volume = {abs/2112.12786},
  year={2021},
  url = {http://arxiv.org/abs/2112.12786}
}

@inproceedings{pan23,
  title={Slide-Transformer: Hierarchical Vision Transformer with Local Self-Attention},
  author={Pan, Xuran and Ye, Tianzhu and Xia, Zhuofan and Song, Shiji and Huang, Gao},
  booktitle={Proceedings of the IEEE/CVF Conference on Computer Vision and Pattern Recognition},
  pages={2082--2091},
  year={2023}
}

@inproceedings{petit21,
  title={U-Net Transformer: Self and Cross Attention for Medical Image Segmentation},
  author={Petit, Olivier and Thome, Nicolas and Rambour, Clement and Themyr, Loic and Collins, Toby and Soler, Luc},
  booktitle={Proceedings of the International Workshop on Machine Learning in Medical Imaging},
  pages={267--276},
  year={2021}
}

@article{hou19,
  title={Cross Attention Network for Few-shot Classification},
  author={Hou, Ruibing and Chang, Hong and Ma, Bingpeng and Shan, Shiguang and Chen, Xilin},
  journal={Advances in Neural Information Processing Systems},
  volume={32},
  year={2019}
}

@ARTICLE{bhattacharya22,
  author={Bhattacharya, Gaurab and Puhan, Niladri B. and Mandal, Bappaditya},
  title={Stand-Alone Composite Attention Network for Concrete Structural Defect Classification}, 
  journal={IEEE Transactions on Artificial Intelligence}, 
  year={2022},
  volume={3},
  number={2},
  pages={265-274},
}

@inproceedings{xie19,
  title={Aggregation Cross-Entropy for Sequence Recognition},
  author={Xie, Zecheng and Huang, Yaoxiong and Zhu, Yuanzhi and Jin, Lianwen and Liu, Yuliang and Xie, Lele},
  booktitle={Proceedings of the IEEE/CVF Conference on Computer Vision and Pattern Recognition},
  pages={6538--6547},
  year={2019}
}

@inproceedings{li19,
  title={Selective Kernel Networks},
  author={Li, Xiang and Wang, Wenhai and Hu, Xiaolin and Yang, Jian},
  booktitle={Proceedings of the IEEE/CVF Conference on Computer Vision and Pattern Recognition},
  pages={510--519},
  year={2019}
}

@inproceedings{ghiasi19,
  title={NAS-FPN: Learning Scalable Feature Pyramid Architecture for Object Detection},
  author={Ghiasi, Golnaz and Lin, Tsung-Yi and Le, Quoc V},
  booktitle={Proceedings of the IEEE/CVF Conference on Computer Vision and Pattern Recognition},
  pages={7036--7045},
  year={2019}
}

@ARTICLE{teymoor25,
  author={Teymoor Seydi, Seyd and Sadegh, Mojtaba and Chanussot, Jocelyn},
  title={Kolmogorov–Arnold Network for Hyperspectral Change Detection}, 
  journal={IEEE Transactions on Geoscience and Remote Sensing}, 
  year={2025},
  volume={63},
  pages={1-15},
}

@book{kolmogorov1961,
  title={On the representation of continuous functions of several variables by superpositions of continuous functions of a smaller number of variables},
  author={Kolmogorov, Andrey Nikolaevich},
  year={1961},
  publisher={American Mathematical Society}
}

@article{ss2024,
  title={Chebyshev Polynomial-Based Kolmogorov-Arnold Networks: An Efficient Architecture for Nonlinear Function Approximation},
  author={SS, Sidharth and AR, Keerthana and R, Gokul and KP, Anas},
  journal = {arXiv CoRR},
  volume = {abs/2405.07200},
  year={2024},
  url = {http://arxiv.org/abs/2405.07200}
}

@article{li24,
  title={Kolmogorov-Arnold Networks are Radial Basis Function Networks},
  author={Li, Ziyao},
  journal = {arXiv CoRR},
  volume = {abs/2405.06721},
  year={2024},
  url = {http://arxiv.org/abs/2405.06721}
}

@article{bodner24,
  title={Convolutional Kolmogorov-Arnold Networks},
  author={Bodner, Alexander Dylan and Tepsich, Antonio Santiago and Spolski, Jack Natan and Pourteau, Santiago},
  journal = {arXiv CoRR},
  volume = {abs/2406.13155},
  year={2024},
  url = {http://arxiv.org/abs/2406.13155}
}

@article{genet25,
  title={TKAN: Temporal Kolmogorov-Arnold Networks}, 
  author={Genet, Remi and Inzirillo, Hugo},
  year={2025},
  journal = {arXiv CoRR},
  volume = {abs/2405.07344},
  url = {http://arxiv.org/abs/2405.07344}
}

@article{seydi25,
  title={Kolmogorov–Arnold Network for Hyperspectral
Change Detection},
  author={Seydi, Seyd Teymoor and Sadegh, Mojtaba and Chanussot, Jocelyn},
  journal={IEEE Transactions on Geoscience and Remote Sensing},
  year={2025}
}

@article{polo24,
  title={MonoKAN: Certified Monotonic Kolmogorov-Arnold Network},
  author={Polo-Molina, Alejandro and Alfaya, David and Portela, Jose},
  year={2024},
  journal = {arXiv CoRR},
  volume = {abs/2409.11078},
  url = {http://arxiv.org/abs/2409.11078}
}

@article{moradi24,
  title={Kolmogorov-Arnold Network Autoencoders},
  author={Moradi, Mohammadamin and Panahi, Shirin and Bollt, Erik and Lai, Ying-Cheng},
  journal = {arXiv CoRR},
  volume = {abs/2410.02077},
  url = {http://arxiv.org/abs/2410.02077},
  year={2024}
}

@inproceedings{ferdaus24,
  title={KANICE: Kolmogorov-Arnold Networks with Interactive Convolutional Elements},
  author={Ferdaus, Md Meftahul and Abdelguerfi, Mahdi and Ioup, Elias and Dobson, David and Niles, Kendall N and Pathak, Ken and Sloan, Steven},
  booktitle={Proceedings of the International Conference on AI-ML Systems},
  pages={1--10},
  year={2024}
}

@article{yang24,
  title={Kolmogorov-Arnold Transformer},
  author={Yang, Xingyi and Wang, Xinchao},
  journal = {arXiv CoRR},
  volume = {abs/2409.10594},
  url = {http://arxiv.org/abs/2409.10594},
  year={2024}
}

@article{krzywda25,
  title={Kolmogorov-Arnold networks for metal surface defect classification},
  author={Krzywda, Maciej and Wermi{\'n}ski, Mariusz and {\L}ukasik, Szymon and Gandomi, Amir H},
  journal = {arXiv CoRR},
  volume = {abs/2501.06389},
  url = {http://arxiv.org/abs/2501.06389},
  year={2025}
}

@inproceedings{satorras18,
  title={Few-Shot Learning with Graph Neural Networks},
  author={Satorras, Victor Garcia and Estrach, Joan Bruna},
  booktitle={Proceedings of the International Conference on Learning Representations},
  year={2018}
}

@inproceedings{sung18,
  title={Learning to compare: Relation network for few-shot learning},
  author={Sung, Flood and Yang, Yongxin and Zhang, Li and Xiang, Tao and Torr, Philip HS and Hospedales, Timothy M},
  booktitle={Proceedings of the IEEE conference on computer vision and pattern recognition},
  pages={1199--1208},
  year={2018}
}

@article{tian20,
  title={Prior Guided Feature Enrichment Network for Few-Shot Segmentation},
  author={Tian, Zhuotao and Zhao, Hengshuang and Shu, Michelle and Yang, Zhicheng and Li, Ruiyu and Jia, Jiaya},
  journal={IEEE Transactions on Pattern Analysis and Machine Intelligence},
  volume={44},
  number={2},
  pages={1050--1065},
  year={2020}
}

@article{ren2025,
  title={MSCA: A few-shot semantic segmentation framework driven by multi-scale cross-attention and information extraction},
  author={Ren, Zhihao and Lu, Shengning and Wang, Xinhua and Liu, Yaoming and Liang, Yong},
  journal={Computer Vision and Image Understanding},
  pages={104419},
  year={2025}
}

@inproceedings{lin23,
  title={Few Shot Medical Image Segmentation with Cross Attention Transformer},
  author={Lin, Yi and Chen, Yufan and Cheng, Kwang-Ting and Chen, Hao},
  booktitle={International Conference on Medical Image Computing and Computer-Assisted Intervention},
  pages={233--243},
  year={2023}
}

@article{ding24,
  title={LERENet: Eliminating Intra-class Differences for Metal Surface Defect Few-shot Semantic Segmentation},
  author={Ding, Hanze and Wu, Zhangkai and Zhang, Jiyan and Ping, Ming and Liu, Yanfang},
  year={2024},
  journal = {arXiv CoRR},
  volume = {abs/2403.11122},
  url = {http://arxiv.org/abs/2403.11122},
}

@article{shi23,
  title={Few-shot semantic segmentation for industrial defect recognition},
  author={Shi, Xiangwen and Zhang, Shaobing and Cheng, Miao and He, Lian and Tang, Xianghong and Cui, Zhe},
  journal={Computers in Industry},
  volume={148},
  pages={103901},
  year={2023}
}

@inproceedings{jadon20,
  title={A survey of loss functions for semantic segmentation},
  author={Jadon, Shruti},
  booktitle={Proceedings of the IEEE Conference on Computational Intelligence in Bioinformatics and Computational Biology},
  pages={1--7},
  year={2020}
}

@article{terven23,
  title={Loss functions and metrics in deep learning},
  author={Terven, Juan and Cordova-Esparza, Diana M and Ramirez-Pedraza, Alfonso and Chavez-Urbiola, Edgar A and Romero-Gonzalez, Julio A},
  year={2023},
  journal = {arXiv CoRR},
  volume = {abs/2307.02694},
  url = {http://arxiv.org/abs/2307.02694},
}

@article{dice1945,
  title={Measures of the Amount of Ecologic Association Between Species},
  author={Dice, Lee R},
  journal={Ecology},
  volume={26},
  number={3},
  pages={297--302},
  year={1945}
}

@InProceedings{lin17,
author = {Lin, Tsung-Yi and Goyal, Priya and Girshick, Ross and He, Kaiming and Dollar, Piotr},
title = {Focal Loss for Dense Object Detection},
booktitle = {Proceedings of the IEEE International Conference on Computer Vision},
year = {2017}
}

@article{pachetti24,
  title={A systematic review of few-shot learning in medical imaging},
  author={Pachetti, Eva and Colantonio, Sara},
  journal={Artificial Intelligence in Medicine},
  volume={156},
  pages={102949},
  year={2024}
}

@article{kaiser17,
  title={Depthwise Separable Convolutions for Neural
Machine Translation},
  author={Kaiser, Lukasz and Gomez, Aidan N and Chollet, Francois},
  journal = {arXiv CoRR},
  volume = {abs/1706.03059},
  url = {http://arxiv.org/abs/1706.03059},
  year={2017}
}

@article{cang24,
  title={Can KAN Work? Exploring the Potential of Kolmogorov-Arnold Networks in Computer Vision},
  author={Cang, Yueyang and Liu, Yuhang and Shi, Li},
  journal = {arXiv CoRR},
  volume = {abs/2411.06727},
  url = {http://arxiv.org/abs/2411.06727},
  year={2024}
}

@article{luo24,
  title={An efficient multi-scale channel attention network for person re-identification},
  author={Luo, Qian and Shao, Jie and Dang, Wanli and Geng, Long and Zheng, Huaiyu and Liu, Chang},
  journal={The Visual Computer},
  volume={40},
  number={5},
  pages={3515--3527},
  year={2024}
}

@article{dougan23,
  title={Which pooling method is better: Max, avg, or concat (Max, Avg)},
  author={Do{\u{g}}an, Yahya},
  journal={Communications Faculty of Sciences University of Ankara Series A2-A3 Physical Sciences and Engineering},
  volume={66},
  number={1},
  pages={95--117},
  year={2023}
}

@inproceedings{woo18,
  title={CBAM: Convolutional Block Attention Module},
  author={Woo, Sanghyun and Park, Jongchan and Lee, Joon-Young and Kweon, In So},
  booktitle={Proceedings of the European Conference on Computer Vision},
  pages={3--19},
  year={2018}
}

@inproceedings{lee15,
  title={Deeply-Supervised Nets},
  author={Lee, Chen-Yu and Xie, Saining and Gallagher, Patrick and Zhang, Zhengyou and Tu, Zhuowen},
  booktitle={Artificial Intelligence and Statistics},
  pages={562--570},
  year={2015}
}

@inbook{szeliski22, 
 title = {Computer Vision: Algorithms and Applications 2nd Edition}, 
 chapter = {Semantic segmentation}, 
 author = {Szeliski, Richard}, 
 publisher={Springer},
 pages = {387--396}, 
 year = {2022} 
}

@article{ferdaus25,
  title={Few-Shot Learning in Video and 3D Object Detection: A Survey},
  author={Ferdaus, Md Meftahul and Niles, Kendall N and Tom, Joe and Abdelguerfi, Mahdi and Ioup, Elias},
  journal = {arXiv CoRR},
  volume = {abs/2507.17079},
  url = {http://arxiv.org/abs/2507.17079},
  year={2025}
}

@article{ferdaus2025karma,
  title={KARMA: Efficient Structural Defect Segmentation via Kolmogorov-Arnold Representation Learning},
  author={Ferdaus, Md Meftahul and Abdelguerfi, Mahdi and Ioup, Elias and Sloan, Steven and Niles, Kendall N and Pathak, Ken},
  journal = {arXiv CoRR},
  volume = {abs/2508.08186},
  url = {http://arxiv.org/abs/2508.08186},
  year={2025}
}

@article{kuchi21,
title = {A machine learning approach to detecting cracks in levees and floodwalls},
journal = {Remote Sensing Applications: Society and Environment},
volume = {22},
pages = {100513},
year = {2021},
author={Kuchi, Aditi and Panta, Manisha and Hoque, Md Tamjidul and Abdelguerfi, Mahdi and Flanagin, Maik C},
}

@article{thrainer25,
  title={FORTRESS: Function-composition Optimized Real-Time Resilient Structural Segmentation via Kolmogorov-Arnold Enhanced Spatial Attention Networks},
  author={Thrainer, Christina and Ferdaus, Md Meftahul and Abdelguerfi, Mahdi and Guetl, Christian and Sloan, Steven and Niles, Kendall N and Pathak, Ken},
  journal = {arXiv CoRR},
  volume = {abs/2507.12675},
  url = {http://arxiv.org/abs/2507.12675},
  year={2025}
}


\appendix
\addpart*{Appendix} 
\chapter{Appendix}\label{appdx}
The appendix includes additional results related to this thesis, providing extra evidence that supports and strengthens the findings presented in the main chapters.

\section{Results Data Preprocessing}\label{appx_data_preprocessing}

This section presents results that compare different state-of-the-art models and methods proposed in Chapter~\ref{culvert_sewer_defect_dataset}. Table~\ref{tab:appx_baseline} includes the baseline achieved by the original E-FPN model and the original dataset. Table~\ref{tab:appx_100} shows the performance of different segmentation models using all the training data and our proposed preprocessing pipeline, including data augmentation and dynamic label injection. To demonstrate stable training, Tables~\ref{tab:appx_50} and \ref{tab:appx_25} demonstrate the performance with less training data, namely 50\% and 25\% respectively.  Table~\ref{tab:appx_augmentation} presents results achieved with different data augmentation techniques, while Table~\ref{tab:appx_DLI} includes the outcomes when only dynamic label injection is performed and no traditional data augmentation.
\begin{table}[htp]
\centering
\scriptsize
\setlength{\tabcolsep}{3.5pt}
\resizebox{1.0\textwidth}{!}{%
\begin{tabular}{l|ccc|cc|cc|ccc}
\toprule
\multirow{2}{*}{\textbf{Model}}
 & \multicolumn{3}{c}{\textbf{Params}}
 & \multicolumn{2}{c}{\textbf{F1 Score}}
 & \multicolumn{2}{c}{\textbf{mIoU}}
 & \multirow{2}{*}{\textbf{Bal. Acc.}}
 & \multirow{2}{*}{\textbf{Mean MCC}}
 & \multirow{2}{*}{\textbf{FW IoU}} \\
\cmidrule(lr){2-4}\cmidrule(lr){5-6}\cmidrule(lr){7-8}
& \textbf{(M)} & \textbf{GFLOPS} & \textbf{Size (MB)}
 & \textbf{w/bg} & \textbf{w/o}
 & \textbf{w/bg} & \textbf{w/o}
& & & \\
\midrule
U-Net\cite{ronneberger2015u}   & 31.04 & 13.69 & 118.47 & 60.18 & 55.67 & 46.14 & 40.31 & 61.97 & 57.27 & 49.28 \\
FPN\cite{lin2017feature}       & 21.20 & 7.81 & 80.89 & 59.39 & 54.87 & 45.71 & 39.98 & 59.23 & 56.53 & 49.02 \\
Att. U-Net\cite{oktay2018attention} & 31.40 & 13.97 & 119.85 & 67.54 & 63.92 & 53.55 & 48.59 & 69.79 & 64.91 & 53.97 \\
UNet++\cite{zhou2018unet++}    & 4.98 & 6.46 & 19.03 & 65.77 & 61.91 & 52.83 & 47.75 & 58.55 & 64.09 & 53.52 \\
BiFPN\cite{tan2020efficientdet} & 4.46 & 17.76 & 17.07 & 69.18 & 65.73 & 55.59 & 50.82 & 64.29 & 66.73 & 57.26 \\
SA-UNet\cite{guo2021sa}   & 7.86 & 3.62 & 1.85 & 68.75 & 65.27 & 54.97 & 50.18 & \textbf{73.15} & 66.22 & 57.24 \\
UNet3+\cite{huang2020unet}     & 25.59 & 33.04 & 97.68 & 72.12 & 69.04 & 58.85 & 54.49 & 71.63 & 69.38 & 58.15 \\
UNeXt\cite{valanarasu2022unext}   & 6.29 & 1.16 & 24.02 & 71.83 & 68.72 & 58.72 & 54.35 & 69.17 & 68.95 & 57.31 \\
EGE-UNet\cite{ruan2023ege} & 3.02 & 0.31 & 11.54 & 55.91 & 50.97 & 42.44 & 36.35 & 47.97 & 52.96 & 39.44 \\
Rolling UNet-L\cite{liu2024rolling} & 28.33 & 8.22 & 108.08 & 12.06 & 01.87 & 10.66 & 01.00 & 12.03 & 04.53 & 02.61 \\
\hline
HierarchicalViT U-Net\cite{ghahremani2024h} & 14.58 & 1.31 & 55.63 & 53.73 & 48.53 & 40.23 & 33.87 & 52.16 & 50.88 & 41.69 \\
Swin-UNet\cite{cao2021swin}      & 14.50 & 0.98 & 55.42 & 58.09 & 53.41 & 44.18 & 38.27 & 54.59 & 54.20 & 43.75 \\
MobileUNETR\cite{perera2024mobileunetr}           & 12.71 & 1.07 & 48.62  & 11.87 & 01.65 & 10.55 & 00.88 & 11.92 & 03.94 & 02.37   \\
Segformer\cite{xie2021segformer}             & 13.67 & 0.78 & 52.15 & 55.31 & 50.48 & 41.10 & 35.16 & 56.02 & 50.85 & 34.78 \\
FasterVit\cite{hatamizadeh2024fastervit}             & 25.23 & 1.57 & 96.27 & 13.11 & 03.04 & 11.32 & 01.73 & 12.85 & 05.33 & 05.09 \\
\hline
U-KAN\cite{li2024ukan}          & 25.36 & 1.73 & 96.98 & 69.81 & 66.43 & 56.37 & 51.69 & 68.48 & 67.27 & 57.62 \\
\hline
E-FPN~\cite{alshawi24b}  & 8.45 & 1.83 & 32.26 & \textbf{74.95} & \textbf{72.16} & \textbf{62.27} & \textbf{58.24} & 70.49 & \textbf{73.38} & \textbf{63.92} \\
\hline
\bottomrule
\multicolumn{11}{l}{Note: M = Million, MB = Mega Byte, bg = background, Bal. Acc. = Balanced Accuracy, FW = Frequency Weighted,}\\
\multicolumn{11}{l}{MCC = Matthews Correlation Coefficient}
\end{tabular}%
}
\caption{Comparison of performance metrics across different models \\ with the original dataset, without data augmentation and dynamic label injection.}
\label{tab:appx_baseline}
\end{table}

\begin{table}[htp]
\centering
\scriptsize
\setlength{\tabcolsep}{3.5pt}
\resizebox{1.0\textwidth}{!}{%
\begin{tabular}{l|ccc|cc|cc|ccc}
\toprule
\multirow{2}{*}{\textbf{Model}}
 & \multicolumn{3}{c}{\textbf{Params}}
 & \multicolumn{2}{c}{\textbf{F1 Score}}
 & \multicolumn{2}{c}{\textbf{mIoU}}
 & \multirow{2}{*}{\textbf{Bal. Acc.}}
 & \multirow{2}{*}{\textbf{Mean MCC}}
 & \multirow{2}{*}{\textbf{FW IoU}} \\
\cmidrule(lr){2-4}\cmidrule(lr){5-6}\cmidrule(lr){7-8}
& \textbf{(M)} & \textbf{GFLOPS} & \textbf{Size (MB)}
 & \textbf{w/bg} & \textbf{w/o}
 & \textbf{w/bg} & \textbf{w/o}
& & & \\
\midrule
U-Net\cite{ronneberger2015u}   & 31.04 & 13.69 & 118.47 & 75.40 & 72.64 & 63.14 & 59.14 & 73.02 & 73.33 & 67.01 \\
FPN\cite{lin2017feature}       & 21.20 & 7.81 & 80.89 & 76.84 & 74.26 & 64.51 & 60.68 & 73.38 & 74.77 & 67.32 \\
Att. U-Net\cite{oktay2018attention} & 31.40 & 13.97 & 119.85 & 78.81 & 76.46 & 66.92 & 63.35 & 77.80 & 76.74 & 68.47 \\
UNet++\cite{zhou2018unet++}    & 4.98 & 6.46 & 19.03 & 73.85 & 70.92 & 61.27 & 57.08 & 66.58 & 72.23 & 63.64 \\
BiFPN\cite{tan2020efficientdet} & 4.46 & 17.76 & 17.07 & 77.41 & 74.89 & 65.35 & 61.62 & 73.59 & 75.38 & 67.00 \\
SA-UNet\cite{guo2021sa}   & 7.86 & 3.62 & 1.85 & 78.79 & 76.43 & 66.87 & 63.30 & \textbf{78.65} & 76.87 & 68.16 \\
UNet3+\cite{huang2020unet}     & 25.59 & 33.04 & 97.68 & 78.50 & 76.11 & 66.63 & 63.04 & 76.89 & 76.41 & 68.29 \\
UNeXt\cite{valanarasu2022unext}   & 6.29 & 1.16 & 24.02 & 76.58 & 74.00 & 64.33 & 60.52 & 74.34 & 74.26 & 64.14 \\
EGE-UNet\cite{ruan2023ege} & 3.02 & 0.31 & 11.54 & 68.90 & 65.47 & 55.08 & 50.35 & 63.46 & 66.04 & 53.40 \\
Rolling UNet-L\cite{liu2024rolling} & 28.33 & 8.22 & 108.08 & 75.24 & 72.46 & 62.46 & 58.39 & 76.75 & 73.22 & 66.87 \\
\hline
HierarchicalViT U-Net\cite{ghahremani2024h} & 14.58 & 1.31 & 55.63 & 53.97 & 48.84 & 41.64 & 35.53 & 53.70 & 50.02 & 40.79 \\
Swin-UNet\cite{cao2021swin}      & 14.50 & 0.98 & 55.42 & 70.95 & 67.75 & 57.68 & 53.22 & 71.43 & 68.07 & 58.08 \\
MobileUNETR\cite{perera2024mobileunetr}           & 12.71 & 1.07 & 48.62 & 74.69 & 71.89 & 62.05 & 58.01 & 76.11 & 72.53 & 62.78  \\
Segformer\cite{xie2021segformer}             & 13.67 & 0.78 & 52.15 & 66.63 & 62.95 & 53.14 & 48.22 & 68.05 & 63.33 & 53.63 \\
FasterVit\cite{hatamizadeh2024fastervit}             & 25.23 & 1.57 & 96.27 & 68.35 & 64.78 & 55.21 & 50.37 & 62.68 & 66.20 & 58.30 \\
\hline
U-KAN\cite{li2024ukan}          & 25.36 & 1.73 & 96.98 & 77.41 & 74.93 & 65.27 & 61.59 & 77.65 & 75.19 & 64.50 \\
\hline
E-FPN~\cite{alshawi24b} &  8.45 & 1.83 & 32.26 & \textbf{79.85} & \textbf{77.63} & \textbf{68.30} & \textbf{64.93} & 77.76 & \textbf{77.99} & \textbf{69.15} \\ 
\hline
\bottomrule
\multicolumn{11}{l}{Note: M = Million, MB = Mega Byte, bg = background, Bal. Acc. = Balanced Accuracy, FW = Frequency Weighted,}\\
\multicolumn{11}{l}{MCC = Matthews Correlation Coefficient}
\end{tabular}%
}
\caption{Comparison of performance metrics across different models \\ with  100~\% of training data, data augmentation and dynamic label injection.}
\label{tab:appx_100}
\end{table}

\begin{table}[H]
\centering
\scriptsize
\setlength{\tabcolsep}{3.5pt}
\resizebox{1.0\textwidth}{!}{%
\begin{tabular}{l|ccc|cc|cc|ccc}
\toprule
\multirow{2}{*}{\textbf{Model}}
 & \multicolumn{3}{c}{\textbf{Params}}
 & \multicolumn{2}{c}{\textbf{F1 Score}}
 & \multicolumn{2}{c}{\textbf{mIoU}}
 & \multirow{2}{*}{\textbf{Bal. Acc.}}
 & \multirow{2}{*}{\textbf{Mean MCC}}
 & \multirow{2}{*}{\textbf{FW IoU}} \\
\cmidrule(lr){2-4}\cmidrule(lr){5-6}\cmidrule(lr){7-8}
& \textbf{(M)} & \textbf{GFLOPS} & \textbf{Size (MB)}
 & \textbf{w/bg} & \textbf{w/o}
 & \textbf{w/bg} & \textbf{w/o}
& & & \\
\midrule
U-Net\cite{ronneberger2015u}   & 31.04 & 13.69 & 118.47 & 68.44 & 64.89 & 55.50 & 50.70 & 64.16 & 66.37 & 58.20 \\
FPN\cite{lin2017feature}       & 21.20 & 7.81 & 80.89 & 65.56 & 61.66 & 53.23 & 48.16 & 60.45 & 63.67 & 58.89 \\
Att. U-Net\cite{oktay2018attention} & 31.40 & 13.97 & 119.85 & 69.93 & 66.54 & 56.73 & 52.02 & 64.51 & 67.86 & 61.16 \\
UNet++\cite{zhou2018unet++}    & 4.98 & 6.46 & 19.03 & 67.16 & 63.45 & 54.16 & 49.20 & 58.33 & 65.84 & 57.34 \\
BiFPN\cite{tan2020efficientdet} & 4.46 & 17.76 & 17.07 & 69.07 & 65.57 & 55.82 & 51.01 & 63.97 & 67.14 & 60.98 \\
SA-UNet\cite{guo2021sa}   & 7.86 & 3.62 & 1.85 & \textbf{71.70} & \textbf{68.52} & \textbf{58.56} & \textbf{54.06} & \textbf{69.01} & \textbf{69.54} & 61.36 \\
UNet3+\cite{huang2020unet}     & 25.59 & 33.04 & 97.68 & 66.20 & 62.35 & 53.61 & 48.53 & 62.36 & 64.90 & 61.57 \\
UNeXt\cite{valanarasu2022unext}   & 6.29 & 1.16 & 24.02 & 66.29 & 62.47 & 53.47 & 48.42 & 59.83 & 64.88 & 58.56 \\
EGE-UNet\cite{ruan2023ege} & 3.02 & 0.31 & 11.54 & 51.68 & 46.20 & 39.15 & 32.62 & 45.85 & 49.38 & 42.00 \\
Rolling UNet-L\cite{liu2024rolling} & 28.33 & 8.22 & 108.08 & 66.53 & 62.75 & 52.86 & 47.75 & 61.47 & 64.21  & 56.69 \\
\hline
HierarchicalViT U-Net\cite{ghahremani2024h} & 14.58 & 1.31 & 55.63 & 46.49 & 40.52 & 33.42 & 26.47 & 48.24 & 42.28 & 32.28 \\
Swin-UNet\cite{cao2021swin}      & 14.50 & 0.98 & 55.42 & 59.79 & 55.28 & 45.76 & 39.97 & 57.45 & 56.21 & 47.08 \\
MobileUNETR\cite{perera2024mobileunetr}           & 12.71 & 1.07 & 48.62 & 70.32 & 67.02 & 57.03 & 52.45 & 68.10 & 67.44 & 57.49 \\
Segformer\cite{xie2021segformer}             & 13.67 & 0.78 & 52.15 & 64.30 & 60.35 & 50.59 & 45.41 & 60.03 & 60.70 & 49.19 \\
FasterVit\cite{hatamizadeh2024fastervit}             & 25.23 & 1.57 & 96.27 & 54.55 & 49.39 & 42.13 & 35.89 & 48.16 & 52.24 & 46.11 \\
\hline
U-KAN\cite{li2024ukan}          & 25.36 & 1.73 & 96.98 & 66.68 & 62.90 & 53.60 & 48.54 & 60.86 & 65.24 & 58.40 \\
\hline
E-FPN~\cite{alshawi24b} &  8.45 & 1.83 & 32.26 & 70.82 & 67.54 & 57.80 & 53.23 & 63.43 & 69.01 & \textbf{61.93} \\ 
\bottomrule
\multicolumn{11}{l}{Note: M = Million, MB = Mega Byte, bg = background, Bal. Acc. = Balanced Accuracy, FW = Frequency Weighted,}\\
\multicolumn{11}{l}{MCC = Matthews Correlation Coefficient}
\end{tabular}%
}
\caption{Comparison of performance metrics across different models with  50~\% of training data, data augmentation and dynamic label injection.}
\label{tab:appx_50}
\end{table}

\clearpage

\begin{table}[H]
\centering
\scriptsize
\setlength{\tabcolsep}{3.5pt}
\resizebox{1.0\textwidth}{!}{%
\begin{tabular}{l|ccc|cc|cc|ccc}
\toprule
\multirow{2}{*}{\textbf{Model}}
 & \multicolumn{3}{c}{\textbf{Params}}
 & \multicolumn{2}{c}{\textbf{F1 Score}}
 & \multicolumn{2}{c}{\textbf{mIoU}}
 & \multirow{2}{*}{\textbf{Bal. Acc.}}
 & \multirow{2}{*}{\textbf{Mean MCC}}
 & \multirow{2}{*}{\textbf{FW IoU}} \\
\cmidrule(lr){2-4}\cmidrule(lr){5-6}\cmidrule(lr){7-8}
& \textbf{(M)} & \textbf{GFLOPS} & \textbf{Size (MB)}
 & \textbf{w/bg} & \textbf{w/o}
 & \textbf{w/bg} & \textbf{w/o}
& & & \\
\midrule
U-Net\cite{ronneberger2015u}   & 31.04 & 13.69 & 118.47 & 62.19 & 57.96 & 48.33 & 42.82 & 59.29 & 58.70 & 49.63 \\
FPN\cite{lin2017feature}       & 21.20 & 7.81 & 80.89 & 55.35 & 50.44 & 41.85 & 35.83 & 60.20 & 53.20 & 42.06 \\
Att. U-Net\cite{oktay2018attention} & 31.40 & 13.97 & 119.85 & 66.66 & 62.92 & 53.34 & 48.34 & 63.94 & 64.24 & 55.50 \\
UNet++\cite{zhou2018unet++}    & 4.98 & 6.46 & 19.03 & 63.24 & 59.08 & 49.70 & 44.26 & 56.84 & 61.14 & 51.93 \\
BiFPN\cite{tan2020efficientdet} & 4.46 & 17.76 & 17.07 & 65.73 & 61.87 & 51.86 & 46.67 & 61.05 & 63.38 & 55.36 \\
SA-UNet\cite{guo2021sa}   & 7.86 & 3.62 & 1.85 & 67.02 & 63.34 & 53.25 & 48.25 & \textbf{70.31} & 64.54 & 54.85 \\
UNet3+\cite{huang2020unet}     & 25.59 & 33.04 & 97.68 & 65.99 & 62.18 & 52.46 & 47.38 & 64.26 & 63.11 & 54.20 \\
UNeXt\cite{valanarasu2022unext}   & 6.29 & 1.16 & 24.02 & 65.14 & 61.25 & 51.11 & 45.89 & 62.08 & 62.27 & 51.05 \\
EGE-UNet\cite{ruan2023ege} & 3.02 & 0.31 & 11.54 & 45.77 & 39.64 & 34.00 & 26.99 & 40.94 & 42.36 & 36.55 \\
Rolling UNet-L\cite{liu2024rolling} & 28.33 & 8.22 & 108.08 & 11.16 & 00.87 & 10.16 & 00.45 & 11.51 & 02.67 & 01.22 \\
\hline
HierarchicalViT U-Net\cite{ghahremani2024h} & 14.58 & 1.31 & 55.63 & 56.68 & 51.87 & 43.45 & 37.54 & 60.80 & 53.24 & 41.30 \\
Swin-UNet\cite{cao2021swin}      & 14.50 & 0.98 & 55.42 & 60.27 & 55.84 & 46.80 & 41.16 & 57.48 & 56.58 & 47.56 \\
MobileUNETR\cite{perera2024mobileunetr}           & 12.71 & 1.07 & 48.62 & 13.25 & 03.22 & 11.40 & 01.85 & 13.09 & 05.02 & 05.77 \\
Segformer\cite{xie2021segformer}             & 13.67 & 0.78 & 52.15 & 12.05 & 01.88 & 10.60 & 00.97 & 12.18 & 04.32 & 01.10 \\
FasterVit\cite{hatamizadeh2024fastervit}             & 25.23 & 1.57 & 96.27 & 12.70 & 02.60 & 11.05 & 01.45 & 12.50 & 04.45 & 04.16 \\
\hline
U-KAN\cite{li2024ukan}          & 25.36 & 1.73 & 96.98 & 66.61 & 62.87 & 53.11 & 48.08 & 63.30 & 64.12 & 53.95  \\
\hline
E-FPN~\cite{alshawi24b} & 8.45 & 1.83 & 32.26 & \textbf{67.75} & \textbf{64.14} & \textbf{54.53} & \textbf{49.65} & 61.40 & \textbf{65.87} & \textbf{57.17} \\ 
\hline
\bottomrule
\multicolumn{11}{l}{Note: M = Million, MB = Mega Byte, bg = background, Bal. Acc. = Balanced Accuracy, FW = Frequency Weighted,}\\
\multicolumn{11}{l}{MCC = Matthews Correlation Coefficient}
\end{tabular}%
}
\caption{Comparison of performance metrics across different models with  25~\% of training data, data augmentation and dynamic label injection.}
\label{tab:appx_25}
\end{table}

\begin{table}[H]
\centering
\scriptsize
\setlength{\tabcolsep}{3.5pt}
\resizebox{1.0\textwidth}{!}{%
\begin{tabular}{l|ccc|cc|cc|ccc}
\toprule
\multirow{2}{*}{\textbf{Model}}
 & \multicolumn{3}{c}{\textbf{Params}}
 & \multicolumn{2}{c}{\textbf{F1 Score}}
 & \multicolumn{2}{c}{\textbf{mIoU}}
 & \multirow{2}{*}{\textbf{Bal. Acc.}}
 & \multirow{2}{*}{\textbf{Mean MCC}}
 & \multirow{2}{*}{\textbf{FW IoU}} \\
\cmidrule(lr){2-4}\cmidrule(lr){5-6}\cmidrule(lr){7-8}
& \textbf{(M)} & \textbf{GFLOPS} & \textbf{Size (MB)}
 & \textbf{w/bg} & \textbf{w/o}
 & \textbf{w/bg} & \textbf{w/o}
& & & \\
\midrule
U-Net\cite{ronneberger2015u}   & 31.04 & 13.69 & 118.47 & 77.94 & 75.49 & 65.81 & 62.13 & 77.81 & 75.93 & 68.20 \\
FPN\cite{lin2017feature}       & 21.20 & 7.81 & 80.89 & 76.40 & 73.76 & 63.91 & 60.02 & 72.23 & 74.29 & 66.56 \\
Att. U-Net\cite{oktay2018attention} & 31.40 & 13.97 & 119.85 & 77.30 & 74.77 & 65.19 & 61.43 & 74.96 & 75.16 & 67.07 \\
UNet++\cite{zhou2018unet++}    & 4.98 & 6.46 & 19.03 & 73.45 & 70.46 & 60.85 & 56.61 & 65.56 & 72.09 & 63.33 \\
BiFPN\cite{tan2020efficientdet} & 4.46 & 17.76 & 17.07 & 77.97 & 75.53 & 66.14 & 62.53 & 76.06 & 76.00 & 66.41 \\
SA-UNet\cite{guo2021sa}   & 7.86 & 3.62 & 1.85 & 78.48 & 76.11 & 66.71 & 63.17 & \textbf{80.23} & 76.69 & 66.53 \\
UNet3+\cite{huang2020unet}     & 25.59 & 33.04 & 97.68 & 78.55 & 76.17 & 66.66 & 63.08 & 76.45 & 76.71 & \textbf{68.54} \\
UNeXt\cite{valanarasu2022unext}   & 6.29 & 1.16 & 24.02 & 75.13 & 72.37 & 62.48 & 58.46 & 72.00 & 72.85 & 64.49 \\
EGE-UNet\cite{ruan2023ege} & 3.02 & 0.31 & 11.54 & 65.52 & 61.65 & 52.19 & 47.07 & 63.10 & 57.96 & 54.42 \\
Rolling UNet-L\cite{liu2024rolling} & 28.33 & 8.22 & 108.08 & 76.91 & 74.34 & 64.39 & 60.55 & 78.12 & 74.88 & 66.51 \\
\hline
HierarchicalViT U-Net\cite{ghahremani2024h} & 14.58 & 1.31 & 55.63 & 60.55 & 56.25 & 47.12 & 41.72 & 64.38 & 56.64 & 44.60 \\
Swin-UNet\cite{cao2021swin}      & 14.50 & 0.98 & 55.42 & 70.08 & 66.77 & 56.75 & 52.16 & 68.73 & 67.06 & 57.42 \\
MobileUNETR\cite{perera2024mobileunetr}           & 12.71 & 1.07 & 48.62 & 73.28 & 70.30 & 60.61 & 56.36 & 76.46 & 71.19 & 63.71 \\
Segformer\cite{xie2021segformer}             & 13.67 & 0.78 & 52.15 & 60.38 & 55.97 & 47.04 & 41.46 & 57.52 & 56.94 & 46.50 \\
FasterVit\cite{hatamizadeh2024fastervit}             & 25.23 & 1.57 & 96.27 & 65.74 & 61.86 & 53.80 & 48.80 & 60.92 & 64.32 & 59.27 \\
\hline
U-KAN\cite{li2024ukan}          & 25.36 & 1.73 & 96.98 & 78.14 & 75.73 & 66.34 & 62.76 & 77.78 & 76.16 & 66.63 \\
\hline
E-FPN~\cite{alshawi24b}  & 8.45 & 1.83 & 32.26 & \textbf{78.71} & \textbf{76.36} & \textbf{66.86} & \textbf{63.33} & 75.08 & \textbf{76.77} & 67.71 \\
\hline
\bottomrule
\multicolumn{11}{l}{Note: M = Million, MB = Mega Byte, bg = background, Bal. Acc. = Balanced Accuracy, FW = Frequency Weighted,}\\
\multicolumn{11}{l}{MCC = Matthews Correlation Coefficient}
\end{tabular}%
}
\caption{Comparison of performance metrics across different models with 100\% of training data, with standard data augmentation techniques horizontal flip, rotation by 30 and 50 degrees, histogram equalization, perspective transform and elastic deformation and without dynamic label injection.}
\label{tab:appx_augmentation}
\end{table}

\clearpage

\begin{table}[H]
\centering
\scriptsize
\setlength{\tabcolsep}{3.5pt}
\resizebox{1.0\textwidth}{!}{%
\begin{tabular}{l|ccc|cc|cc|ccc}
\toprule
\multirow{2}{*}{\textbf{Model}}
 & \multicolumn{3}{c}{\textbf{Params}}
 & \multicolumn{2}{c}{\textbf{F1 Score}}
 & \multicolumn{2}{c}{\textbf{mIoU}}
 & \multirow{2}{*}{\textbf{Bal. Acc.}}
 & \multirow{2}{*}{\textbf{Mean MCC}}
 & \multirow{2}{*}{\textbf{FW IoU}} \\
\cmidrule(lr){2-4}\cmidrule(lr){5-6}\cmidrule(lr){7-8}
& \textbf{(M)} & \textbf{GFLOPS} & \textbf{Size (MB)}
 & \textbf{w/bg} & \textbf{w/o}
 & \textbf{w/bg} & \textbf{w/o}
& & & \\
\midrule
U-Net\cite{ronneberger2015u}   & 31.04 & 13.69 & 118.47 & 57.76 & 52.97 & 44.62 & 38.65 & 56.76 & 54.79 & 50.82 \\
FPN\cite{lin2017feature}       & 21.20 & 7.81 & 80.89 & 63.85 & 59.80 & 49.98 & 44.64 & 59.54 & 61.55 & 53.92 \\
Att. U-Net\cite{oktay2018attention} & 31.40 & 13.97 & 119.85 & 68.05 & 64.53 & 54.16 & 49.35 & 70.52 & 64.95 & 55.20 \\
UNet++\cite{zhou2018unet++}    & 4.98 & 6.46 & 19.03 & 58.75 & 54.06 & 45.68 & 39.78 & 49.43 & 57.73 & 49.21 \\
BiFPN\cite{tan2020efficientdet} & 4.46 & 17.76 & 17.07 & 67.35 & 63.69 & 53.61 & 48.63 & 63.60 & 64.90 & 55.58  \\
SA-UNet\cite{guo2021sa}   & 7.86 & 3.62 & 1.85 & 68.08 & 64.54 & 54.78 & 50.00 & 65.23 & 65.62 & 56.60 \\
UNet3+\cite{huang2020unet}     & 25.59 & 33.04 & 97.68 & 69.66 & 66.31 & 55.91 & 51.25 & 70.40 & 66.77 & 58.30 \\
UNeXt\cite{valanarasu2022unext}   & 6.29 & 1.16 & 24.02 & 64.03 & 60.00 & 50.37 & 45.06 & 65.80 & 61.60 & 54.81 \\
EGE-UNet\cite{ruan2023ege} & 3.02 & 0.31 & 11.54 & 60.16 & 55.75 & 46.29 & 40.68 & 57.22 & 56.35 & 43.99 \\
Rolling UNet-L\cite{liu2024rolling} & 28.33 & 8.22 & 108.08 & 16.01 & 06.29 & 12.89 & 03.48 & 15.05 & 09.96 & 06.68 \\
\hline
HierarchicalViT U-Net\cite{ghahremani2024h} & 14.58 & 1.31 & 55.63 & 58.52 & 53.92 & 45.06 & 39.31 & 65.43 & 55.83 & 45.31 \\
Swin-UNet\cite{cao2021swin}      & 14.50 & 0.98 & 55.42 & 55.01 & 49.96 & 41.22 & 34.96 & 51.46 & 51.08 & 41.51 \\
MobileUNETR\cite{perera2024mobileunetr}           & 12.71 & 1.07 & 48.62 & 17.40 & 07.84 & 13.69 & 04.36 & 15.24 & 12.45 & 07.01 \\
Segformer\cite{xie2021segformer}             & 13.67 & 0.78 & 52.15 & 21.65 & 12.62 & 16.11 & 07.07 & 17.98 & 17.67 & 11.59 \\
FasterVit\cite{hatamizadeh2024fastervit}             & 25.23 & 1.57 & 96.27 & 12.81 & 02.71 & 11.09 & 01.49 & 12.56 & 05.49 & 04.06 \\
\hline
U-KAN\cite{li2024ukan}          & 25.36 & 1.73 & 96.98 & 66.00 & 62.22 & 51.82 & 46.69 & 62.90 & 63.39 & 50.38 \\
\hline
E-FPN~\cite{alshawi24b}  & 8.45 & 1.83 & 32.26 & \textbf{76.35} & \textbf{73.72} & \textbf{64.01} & \textbf{60.17} & \textbf{71.09} & \textbf{74.54} & \textbf{64.14} \\
\hline
\bottomrule
\multicolumn{11}{l}{Note: M = Million, MB = Mega Byte, bg = background, Bal. Acc. = Balanced Accuracy, FW = Frequency Weighted,}\\
\multicolumn{11}{l}{MCC = Matthews Correlation Coefficient}
\end{tabular}%
}
\caption{Comparison of performance metrics across different models with 100\% of training data, with dynamic label injection and without data augmentation.}
\label{tab:appx_DLI}
\end{table}

\section{Results FORTRESS}\label{appx_fortress}

This section presents results comparing our proposed model, FORTRESS, with several state-of-the-art models. A detailed description of FORTRESS can be found in Chapter~\ref{fortress}. Table~\ref{tab:apx_fortress_100} shows the comparison using 100\% of the training data, including our preprocessing pipeline with standard data augmentation and dynamic label injection, as described in Section~\ref{data_experiments}. 

\begin{table}[H]
\centering
\scriptsize
\setlength{\tabcolsep}{3.5pt}
\resizebox{1.0\textwidth}{!}{%
\begin{tabular}{l|cc|cc|cc|ccc}
\toprule
\multirow{2}{*}{\textbf{Model}} 
 & \multicolumn{2}{c}{\textbf{Params}} 
 & \multicolumn{2}{c}{\textbf{F1 Score}} 
 & \multicolumn{2}{c}{\textbf{mIoU}} 
 & \multirow{2}{*}{\textbf{Bal. Acc.}} 
 & \multirow{2}{*}{\textbf{Mean MCC}} 
 & \multirow{2}{*}{\textbf{FW IoU}} \\
\cmidrule(lr){2-3}\cmidrule(lr){4-5}\cmidrule(lr){6-7}
 & \textbf{(M)} & \textbf{GFLOPS} 
 & \textbf{w/bg} & \textbf{w/o} 
 & \textbf{w/bg} & \textbf{w/o}
 & & & \\
\midrule
U-Net\cite{ronneberger2015u}   & 31.04 & 13.69 & 0.754 & 0.726 & 0.631 & 0.591 & 0.730 & 0.733 & 0.670 \\
FPN\cite{lin2017feature}       & 21.20 & 7.81 & 0.768 & 0.743 & 0.645 & 0.607 & 0.734 & 0.748 & 0.673 \\
Att. U-Net\cite{oktay2018attention} & 31.40 & 13.97 & 0.788 & 0.765 & 0.669 & 0.634 & 0.778 & 0.767 & 0.685 \\
UNet++\cite{zhou2018unet++}    & 4.98 & 6.46 & 0.739 & 0.709 & 0.613 & 0.571 & 0.666 & 0.722 & 0.636 \\
BiFPN\cite{tan2020efficientdet} & 4.46 & 17.76 & 0.774 & 0.749 & 0.654 & 0.616 & 0.736 & 0.754 & 0.670 \\
UNet3+\cite{huang2020unet}     & 25.59 & 33.04 & 0.785 & 0.761 & 0.666 & 0.630 & 0.769 & 0.764 & 0.683 \\
UNeXt\cite{valanarasu2022unext}   & 6.29 & 1.16 & 0.766 & 0.740 & 0.643 & 0.605 & 0.743 & 0.743 & 0.641 \\
EGE-UNet\cite{ruan2023ege} & 3.02 & 0.31 & 0.689 & 0.655 & 0.551 & 0.504 & 0.635 & 0.660 & 0.534 \\
Rolling UNet-L\cite{liu2024rolling} & 28.33 & 8.22 & 0.752 & 0.725 & 0.625 & 0.584 & 0.768 & 0.732 & 0.669 \\
\hline
HierarchicalViT U-Net\cite{ghahremani2024h} & 14.58 & 1.31 & 0.540 & 0.488 & 0.416 & 0.355 & 0.537 & 0.500 & 0.408 \\
Swin-UNet\cite{cao2021swin}      & 14.50 & 0.98 & 0.710 & 0.678 & 0.577 & 0.532 & 0.714 & 0.681 & 0.581 \\
MobileUNETR\cite{perera2024mobileunetr}           & 12.71 & 1.07 & 0.747 & 0.719 & 0.621 & 0.580 & 0.761 & 0.725 & 0.628 \\
Segformer\cite{xie2021segformer}             & 13.67 & 0.78 & 0.666 & 0.630 & 0.531 & 0.482 & 0.681 & 0.633 & 0.536 \\
FasterVit\cite{hatamizadeh2024fastervit}             & 25.23 & 1.57 & 0.684 & 0.648 & 0.552 & 0.504 & 0.627 & 0.662 & 0.583 \\
\hline
U-KAN\cite{li2024ukan}          & 25.36 & 1.73 & 0.774 & 0.749 & 0.653 & 0.616 & 0.777 & 0.752 & 0.645 \\
SA-UNet\cite{guo2021sa}       & 7.86 & 3.62 & 0.788 & 0.764 & 0.669 & 0.633 & 0.787 & 0.769 & 0.682 \\
\textbf{FORTRESS} & \textbf{2.89}  & 1.17   & \textbf{0.793} & \textbf{0.771} & \textbf{0.677} & \textbf{0.643} & \textbf{0.787} & \textbf{0.772} & \textbf{0.690} \\
\bottomrule
\multicolumn{10}{l}{\scriptsize Note: bg = background, Bal. Acc. = Balanced Accuracy, FW = Frequency Weighted,}\\
\multicolumn{10}{l}{\scriptsize MCC = Matthews Correlation Coefficient}
\end{tabular}%
}
\caption{Comparison of performance metrics across different models, including our proposed model, FORTRESS, and other state-of-the-art methods, using 100\% of training data, and a preprocessing pipeline with standard data augmentation and dynamic label injection.}
\label{tab:apx_fortress_100}
\end{table}

\noindent
Table~\ref{tab:apx_fortress_100_without_aug} presents the same comparison without the preprocessing pipeline. To demonstrate training stability, Tables~\ref{tab:apx_fortress_50} and \ref{tab:apx_fortress_25} report results using 50\% and 25\% of the training data, respectively.

\begin{table}[H]
\centering
\scriptsize
\setlength{\tabcolsep}{3.5pt}
\resizebox{1.0\textwidth}{!}{%
\begin{tabular}{l|cc|cc|cc|ccc}
\toprule
\multirow{2}{*}{\textbf{Model}} 
 & \multicolumn{2}{c}{\textbf{Params}} 
 & \multicolumn{2}{c}{\textbf{F1 Score}} 
 & \multicolumn{2}{c}{\textbf{mIoU}} 
 & \multirow{2}{*}{\textbf{Bal. Acc.}} 
 & \multirow{2}{*}{\textbf{Mean MCC}} 
 & \multirow{2}{*}{\textbf{FW IoU}} \\
\cmidrule(lr){2-3}\cmidrule(lr){4-5}\cmidrule(lr){6-7}
 & \textbf{(M)} & \textbf{GFLOPS} 
 & \textbf{w/bg} & \textbf{w/o} 
 & \textbf{w/bg} & \textbf{w/o}
 & & & \\
\midrule
U-Net\cite{ronneberger2015u}   & 31.04 & 13.69 & 0.602 & 0.557 & 0.461 & 0.403 & 0.620 & 0.573 & 0.493 \\
FPN\cite{lin2017feature}       & 21.20 & 7.81 & 0.594 & 0.549 & 0.457 & 0.400 & 0.592 & 0.565 & 0.490 \\
Att. U-Net\cite{oktay2018attention} & 31.40 & 13.97 & 0.675 & 0.639 & 0.536 & 0.486 & 0.698 & 0.649 & 0.540 \\
UNet++\cite{zhou2018unet++}    & 4.98 & 6.46 & 0.658 & 0.619 & 0.528 & 0.478 & 0.586 & 0.641 & 0.535 \\
BiFPN\cite{tan2020efficientdet} & 4.46 & 17.76 & 0.692 & 0.657 & 0.556 & 0.508 & 0.643 & 0.667 & 0.573 \\
UNet3+\cite{huang2020unet}     & 25.59 & 33.04 & 0.721 & 0.690 & 0.589 & 0.545 & 0.716 & 0.694 & 0.582 \\
UNeXt\cite{valanarasu2022unext}   & 6.29 & 1.16 & 0.718 & 0.687 & 0.587 & 0.544 & 0.692 & 0.690 & 0.573 \\
EGE-UNet\cite{ruan2023ege} & 3.02 & 0.31 & 0.559 & 0.510 & 0.424 & 0.364 & 0.480 & 0.530 & 0.394 \\
Rolling UNet-L\cite{liu2024rolling} & 28.33 & 8.22 & 0.121 & 0.019 & 0.107 & 0.010 & 0.120 & 0.045 & 0.026 \\
\hline
HierarchicalViT U-Net\cite{ghahremani2024h} & 14.58 & 1.31 & 0.537 & 0.485 & 0.402 & 0.339 & 0.522 & 0.509 & 0.417 \\
Swin-UNet\cite{cao2021swin}      & 14.50 & 0.98 & 0.581 & 0.534 & 0.442 & 0.383 & 0.546 & 0.542 & 0.438 \\
MobileUNETR\cite{perera2024mobileunetr}           & 12.71 & 1.07 & 0.119 & 0.017 & 0.106 & 0.009 & 0.119 & 0.039 & 0.024 \\
Segformer\cite{xie2021segformer}             & 13.67 & 0.78 & 0.553 & 0.505 & 0.411 & 0.352 & 0.560 & 0.509 & 0.348 \\
FasterVit\cite{hatamizadeh2024fastervit}             & 25.23 & 1.57 & 0.131 & 0.030 & 0.113 & 0.017 & 0.129 & 0.053 & 0.051 \\
\hline
U-KAN\cite{li2024ukan}          & 25.36 & 1.73 & 0.698 & 0.664 & 0.564 & 0.517 & 0.685 & 0.673 & 0.576 \\
SA-UNet\cite{guo2021sa}       & 7.86 & 3.62 & 0.688 & 0.653 & 0.550 & 0.502 & 0.732 & 0.662 & 0.572 \\
\textbf{FORTRESS} & \textbf{2.89} & \textbf{1.17} & \textbf{0.783} & \textbf{0.759} & \textbf{0.663} & \textbf{0.627} & \textbf{0.791} & \textbf{0.763} & \textbf{0.674} \\
\bottomrule
\multicolumn{10}{l}{\scriptsize Note: bg = background, Bal. Acc. = Balanced Accuracy, FW = Frequency Weighted,}\\
\multicolumn{10}{l}{\scriptsize MCC = Matthews Correlation Coefficient}
\end{tabular}%
}
\caption{Comparison of performance metrics across different models, including our proposed model, FORTRESS, and other state-of-the-art methods, without data preprocessing.}
\label{tab:apx_fortress_100_without_aug}
\end{table}

\begin{table}[H]
\centering
\scriptsize
\setlength{\tabcolsep}{3.5pt}
\resizebox{1.0\textwidth}{!}{%
\begin{tabular}{l|cc|cc|cc|ccc}
\toprule
\multirow{2}{*}{\textbf{Model}} 
 & \multicolumn{2}{c}{\textbf{Params}} 
 & \multicolumn{2}{c}{\textbf{F1 Score}} 
 & \multicolumn{2}{c}{\textbf{mIoU}} 
 & \multirow{2}{*}{\textbf{Bal. Acc.}} 
 & \multirow{2}{*}{\textbf{Mean MCC}} 
 & \multirow{2}{*}{\textbf{FW IoU}} \\
\cmidrule(lr){2-3}\cmidrule(lr){4-5}\cmidrule(lr){6-7}
 & \textbf{(M)} & \textbf{GFLOPS} 
 & \textbf{w/bg} & \textbf{w/o} 
 & \textbf{w/bg} & \textbf{w/o}
 & & & \\
\midrule
U-Net\cite{ronneberger2015u}   & 31.04 & 13.69 & 0.684 & 0.649 & 0.555 & 0.507 & 0.642 & 0.664 & 0.582 \\
FPN\cite{lin2017feature}       & 21.20 & 7.81 & 0.656 & 0.617 & 0.532 & 0.482 & 0.605 & 0.637 & 0.589 \\
Att. U-Net\cite{oktay2018attention} & 31.40 & 13.97 & 0.699 & 0.665 & 0.567 & 0.520 & 0.645 & 0.679 & 0.612 \\
UNet++\cite{zhou2018unet++}    & 4.98 & 6.46 & 0.672 & 0.635 & 0.542 & 0.492 & 0.583 & 0.658 & 0.573 \\
BiFPN\cite{tan2020efficientdet} & 4.46 & 17.76 & 0.691 & 0.656 & 0.558 & 0.510 & 0.640 & 0.671 & 0.610 \\
UNet3+\cite{huang2020unet}     & 25.59 & 33.04 & 0.662 & 0.624 & 0.536 & 0.485 & 0.624 & 0.649 & 0.616 \\
UNeXt\cite{valanarasu2022unext}   & 6.29 & 1.16 & 0.663 & 0.625 & 0.535 & 0.484 & 0.598 & 0.649 & 0.586 \\
EGE-UNet\cite{ruan2023ege} & 3.02 & 0.31 & 0.517 & 0.462 & 0.392 & 0.326 & 0.459 & 0.494 & 0.420 \\
Rolling UNet-L\cite{liu2024rolling} & 28.33 & 8.22 & 0.665 & 0.628 & 0.529 & 0.478 & 0.615 & 0.642 & 0.567 \\
\hline
HierarchicalViT U-Net\cite{ghahremani2024h} & 14.58 & 1.31 & 0.465 & 0.405 & 0.334 & 0.265 & 0.482 & 0.423 & 0.323 \\
Swin-UNet\cite{cao2021swin}      & 14.50 & 0.98 & 0.598 & 0.553 & 0.458 & 0.400 & 0.575 & 0.562 & 0.471 \\
MobileUNETR\cite{perera2024mobileunetr}           & 12.71 & 1.07 & 0.703 & 0.670 & 0.570 & 0.525 & 0.681 & 0.674 & 0.575 \\
Segformer\cite{xie2021segformer}             & 13.67 & 0.78 & 0.643 & 0.604 & 0.506 & 0.454 & 0.600 & 0.607 & 0.492 \\
FasterVit\cite{hatamizadeh2024fastervit}             & 25.23 & 1.57 & 0.546 & 0.494 & 0.421 & 0.359 & 0.482 & 0.522 & 0.461 \\
\hline
U-KAN\cite{li2024ukan}          & 25.36 & 1.73 & 0.667 & 0.629 & 0.536 & 0.485 & 0.609 & 0.652 & 0.584 \\
SA-UNet\cite{guo2021sa}       & 7.86 & 3.62 & 0.717 & 0.685 & 0.586 & 0.541 & 0.690 & 0.695 & 0.614 \\
\textbf{FORTRESS} & 2.89 & 1.17 & 0.758 & 0.732 & 0.633 & 0.594 & 0.809 & 0.737 & 0.618 \\
\bottomrule
\multicolumn{10}{l}{\scriptsize Note: bg = background, Bal. Acc. = Balanced Accuracy, FW = Frequency Weighted,}\\
\multicolumn{10}{l}{\scriptsize MCC = Matthews Correlation Coefficient}
\end{tabular}%
}
\caption{Comparison of performance metrics across different models, including our proposed model, FORTRESS, and other state-of-the-art methods with 50\% of training data, data augmentation and dynamic label injection.}
\label{tab:apx_fortress_50}
\end{table}

\clearpage

\begin{table}[H]
\centering
\scriptsize
\setlength{\tabcolsep}{3.5pt}
\resizebox{1.0\textwidth}{!}{%
\begin{tabular}{l|cc|cc|cc|ccc}
\toprule
\multirow{2}{*}{\textbf{Model}} 
 & \multicolumn{2}{c}{\textbf{Params}} 
 & \multicolumn{2}{c}{\textbf{F1 Score}} 
 & \multicolumn{2}{c}{\textbf{mIoU}} 
 & \multirow{2}{*}{\textbf{Bal. Acc.}} 
 & \multirow{2}{*}{\textbf{Mean MCC}} 
 & \multirow{2}{*}{\textbf{FW IoU}} \\
\cmidrule(lr){2-3}\cmidrule(lr){4-5}\cmidrule(lr){6-7}
 & \textbf{(M)} & \textbf{GFLOPS} 
 & \textbf{w/bg} & \textbf{w/o} 
 & \textbf{w/bg} & \textbf{w/o}
 & & & \\
\midrule
U-Net\cite{ronneberger2015u}   & 31.04 & 13.69 & 0.622 & 0.580 & 0.483 & 0.428 & 0.593 & 0.587 & 0.496 \\
FPN\cite{lin2017feature}       & 21.20 & 7.81 & 0.554 & 0.504 & 0.419 & 0.358 & 0.602 & 0.532 & 0.421 \\
Att. U-Net\cite{oktay2018attention} & 31.40 & 13.97 & 0.667 & 0.629 & 0.533 & 0.483 & 0.639 & 0.642 & 0.555 \\
UNet++\cite{zhou2018unet++}    & 4.98 & 6.46 & 0.632 & 0.591 & 0.497 & 0.443 & 0.568 & 0.611 & 0.519 \\
BiFPN\cite{tan2020efficientdet} & 4.46 & 17.76 & 0.657 & 0.619 & 0.519 & 0.467 & 0.611 & 0.634 & 0.554 \\
UNet3+\cite{huang2020unet}     & 25.59 & 33.04 & 0.660 & 0.622 & 0.525 & 0.474 & 0.643 & 0.631 & 0.542 \\
UNeXt\cite{valanarasu2022unext}   & 6.29 & 1.16 & 0.651 & 0.613 & 0.511 & 0.459 & 0.621 & 0.623 & 0.511 \\
EGE-UNet\cite{ruan2023ege} & 3.02 & 0.31 & 0.458 & 0.396 & 0.340 & 0.270 & 0.409 & 0.424 & 0.366 \\
Rolling UNet-L\cite{liu2024rolling} & 28.33 & 8.22 & 0.112 & 0.009 & 0.102 & 0.005 & 0.115 & 0.027 & 0.012 \\
\hline
HierarchicalViT U-Net\cite{ghahremani2024h} & 14.58 & 1.31 & 0.567 & 0.519 & 0.435 & 0.375 & 0.608 & 0.532 & 0.413 \\
Swin-UNet\cite{cao2021swin}      & 14.50 & 0.98 & 0.603 & 0.558 & 0.468 & 0.412 & 0.575 & 0.566 & 0.476 \\
MobileUNETR\cite{perera2024mobileunetr}           & 12.71 & 1.07 & 0.133 & 0.032 & 0.114 & 0.019 & 0.131 & 0.050 & 0.058 \\
Segformer\cite{xie2021segformer}             & 13.67 & 0.78 & 0.121 & 0.019 & 0.106 & 0.010 & 0.122 & 0.043 & 0.011 \\
FasterVit\cite{hatamizadeh2024fastervit}             & 25.23 & 1.57 & 0.127 & 0.026 & 0.111 & 0.015 & 0.125 & 0.045 & 0.042 \\
\hline
U-KAN\cite{li2024ukan}          & 25.36 & 1.73 & 0.666 & 0.629 & 0.531 & 0.481 & 0.633 & 0.641 & 0.540 \\
SA-UNet\cite{guo2021sa}       & 7.86 & 3.62 & 0.670 & 0.633 & 0.533 & 0.483 & 0.703 & 0.645 & 0.549 \\
\textbf{FORTRESS} & 2.89 & 1.17 & 0.702 & 0.669 & 0.568 & 0.523 & 0.769 & 0.676 & 0.554 \\
\bottomrule
\multicolumn{10}{l}{\scriptsize Note: bg = background, Bal. Acc. = Balanced Accuracy, FW = Frequency Weighted,}\\
\multicolumn{10}{l}{\scriptsize MCC = Matthews Correlation Coefficient}
\end{tabular}%
}
\caption{Comparison of performance metrics across different models, including our proposed model, FORTRESS, and other state-of-the-art methods with 25\% of training data, data augmentation and dynamic label injection.}
\label{tab:apx_fortress_25}
\end{table}

\end{document}